\documentclass[conference]{IEEEtran}
\IEEEoverridecommandlockouts
\usepackage{etoolbox}
\usepackage{amsmath,amssymb,amsfonts}
\usepackage{algorithm}
\usepackage{algorithmic}
\usepackage{graphicx}
\usepackage{textcomp}
\usepackage{xcolor}
\usepackage{hyperref}
\usepackage[binary-units=true, fraction=nice]{siunitx}
\usepackage{booktabs}
\usepackage{bm}
\usepackage{multirow}

\usepackage[font={sf,scriptsize},           
caption=false]                  
{subfig}
\usepackage{url}

\DeclareMathOperator*{\argmax}{arg\,max}
\DeclareMathOperator*{\R}{\mathbb{R}}
\DeclareMathOperator*{\E}{\mathbb{E}}
\DeclareMathOperator*{\KL}{\mathbb{KL}}
\DeclareMathOperator*{\ent}{\mathbb{H}}
\DeclareMathOperator{\Tr}{Tr}

\newcommand{\emb}{\mathcal{E}}
\newcommand*\diff{\mathop{}\!\mathrm{d}}

\newcommand{\lagr}{\mathcal{L}}
\newcommand{\N}{\mathcal{N}}

\usepackage{xcolor}
\definecolor{LightCyan}{rgb}{0.88,1,1}
\definecolor{cyan}{HTML}{52adcf}
\definecolor{darkcyan}{HTML}{0dadcf}
\definecolor{purple}{HTML}{9488bd}
\definecolor{darkpurple}{HTML}{9418B0}
\definecolor{green}{HTML}{5aab5a}
\definecolor{darkgreen}{HTML}{28AC08}
\definecolor{darkyellow}{HTML}{C9D501}

\graphicspath{{assets/}}

\usepackage[style=numeric,sorting=none]{biblatex}
\appto\biburlsetup{\Urlmuskip=0mu\relax}

\begin{document}

\title{GenRadar: Self-supervised Probabilistic Camera Synthesis based on Radar Frequencies\\
}

\author{\IEEEauthorblockN{Carsten Ditzel}
  \IEEEauthorblockA{\textit{Institute of Measurement, Control and Microtechnology} \\
    Ulm University, Germany\\
    carsten.ditzel@uni-ulm.de} \and \IEEEauthorblockN{ Klaus Dietmayer}
  \IEEEauthorblockA{\textit{Institute of Measurement, Control and Microtechnology} \\
    Ulm University, Germany\\
    klaus.dietmayer@uni-ulm.de} }

\maketitle

\begin{abstract}
  Autonomous systems require a continuous and dependable environment perception
  for navigation and decision-making, which is best achieved by combining
  different sensor types. Radar continues to function robustly in compromised
  circumstances in which cameras become impaired, guaranteeing a steady inflow
  of information. Yet, camera images provide a more intuitive and readily
  applicable impression of the world. This work combines the complementary
  strengths of both sensor types in a unique self-learning fusion approach for a
  probabilistic scene reconstruction in adverse surrounding conditions. After
  reducing the memory requirements of both high-dimensional measurements through
  a decoupled stochastic self-supervised compression technique, the proposed
  algorithm exploits similarities and establishes correspondences between both
  domains at different feature levels during training. Then, at inference time,
  relying exclusively on radio frequencies, the model successively predicts
  camera constituents in an autoregressive and self-contained process. These
  discrete tokens are finally transformed back into an instructive view of the
  respective surrounding, allowing to visually perceive potential dangers for
  important tasks downstream.
\end{abstract}

\section{Introduction}
\label{sec:introduction} A reliable and fail-safe capturing of the environment
builds the foundation for many modern automatic applications like self-driving
cars, robotic drones and diverse military purposes. Camera systems are
omnipresent in this field nowadays, being cost-effective, easy to operate and
their output readily interpretable in an intuitive way by either men or
machine. Yet vision sensors tend to fail or at least their capabilities degrade
in realistic circumstances featuring fog and snow or in poorly lit environments.
\cite{BijelicGruberGated2018}. It would be highly desirable to maintain a clear
impression or at least a rough idea of the system's vicinity, even within poor
conditions to allow for a correct and responsible decision-making. Research so
far has mostly focused on adding lidar sensors to improve sensing abilities
through the inclusion of explicit range information. Conventional lidars,
however, are similarly subjected to environmental influences and their
performance breaks down drastically in harsh conditions
\cite{BijelicGuberLidar2018}. Radar, due to its longer wavelengths largely
unaffected by its surroundings, offers a powerful alternative for the
enhancement of environment perceptions. This work therefore proposes a novel
two-staged multi-modal fusion approach of both vision and microwave sensor data
during neural network training.
\begin{figure}[h!] \centering \includegraphics[width=0.48\textwidth]{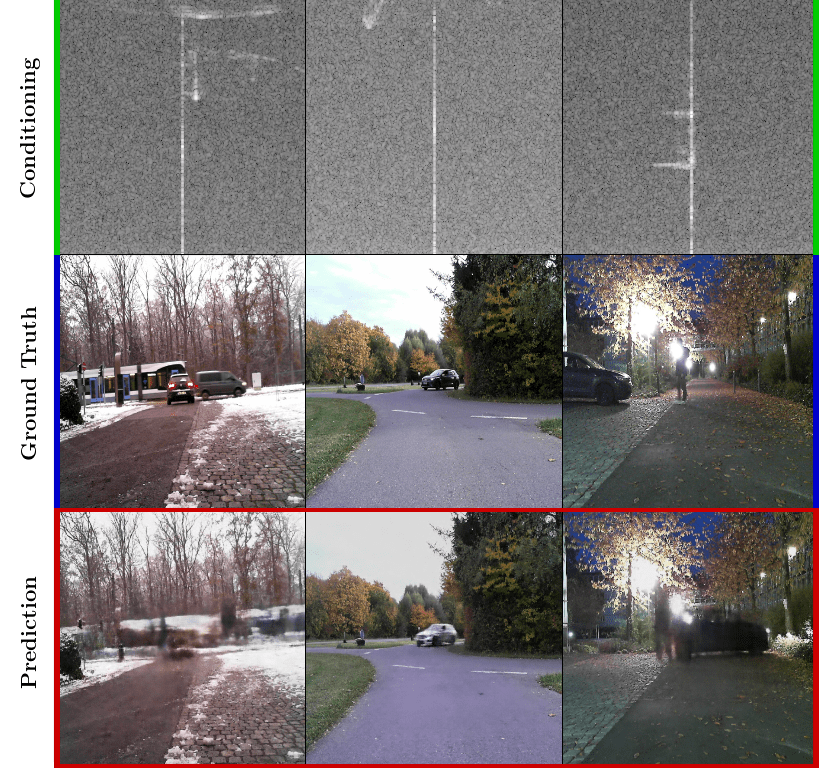}
  \caption{Camera view generation \textcolor{red}{(red)} based solely on
    radar-frequency information \textcolor{darkgreen}{(green)}. The synchronized
    camera ground-truth \textcolor{blue}{(blue)} is supplied for visual
    reference only. The model generally succeeds in inferring essential the
    characteristics and key features of the underlying real-world scenery. Less
    confidence is displayed regarding dynamic objects like the exact
    localization of pedestrians, flashing car lights or distant elements,
    particularly if present in only one of two sensor outputs, lacking
    cross-modal correspondence.}
  \label{fig:front}
\end{figure}
During inference, this allows for the probabilistic reconstruction and visual
traceability of a scene's central elements when only robust radar information is
provided or available. Figure \ref{fig:front} gives an impression of the devised
model's scene-understanding, showcasing three randomly chosen examples taken
from the test dataset, introduced in section \ref{sec:data}. To the present day,
this work is one of only a few deep learning projects, dealing with the
combination of camera and radar sensors \cite{lin2020depth},
\cite{grimm2020warping}, \cite{nobis2019deep}, \cite{lim2019radar} and the first
one demonstrating the fusion of their low-level data by means of completely
annotation-free methods.

Algorithmically, the proposed procedure can be understood as an attempt to unify
the respective strengths of two of the most ubiquitous neural architectures
these days in a self-supervised manner, namely convolutional and transformers
networks. Both models make specific but differing assumptions about the
statistical composition of provided data and have been used to great success in
various research areas in which large quantities of data are to be processed to
find patterns and concealed correlations.

Convolutional neural networks (CNNs) \cite{lecun1989backpropagation} are
predominantly used for computer vision tasks and represent the current
state-of-the-art in this field. Presuming strong regional inter-pixel
correlations, their specific architecture offers a feature called
\textit{inductive bias} allowing for the effective processing of image data,
through small learnable filters which convolve across the input. In combination
with weight-sharing among its many kernels and sub-sampling this makes
convolutional layers approximately equivariant
\cite{zhang2019making}. Colloquially speaking, more importance is placed on the
mere existence of certain information in the input rather than its precise
location within. Their constrained receptive fields, however, allow CNNs to
acquire higher-level features and identifying distant relations in the data only
after the repeated stacking of numerous filters. This hierarchical structure is
often necessary and sufficient for object detection tasks or semantic
segmentation purposes, but may also hinder signal flow through the network if
taken to an extreme. In this work, CNNs will be used in the first stage to
derive effective modal-agnostic compression models. These learn to quantize the
continuous high-dimensional measurements of both sensors into a stochastic and
discrete number of constituents, summarizing their essential content.

Transformers \cite{vaswani2017attention} in contrast, are specifically designed
to model long-range interactions and find far-reaching concepts along the entire
signal processing chain right from the start. The key ingredient to this ability
lies within transformers self-attention mechanism, allowing them to capture
global as well as local interaction between its sequential input alike. By stringing
together multiple of these attention elements these architectures continue to
achieve state-of-the-art results in many diverse fields of application. Their
computational complexity, though, grows quadratically with sequence length,
making them generally inapplicable for high-dimensional input. Operating on data
streams of arbitrary length and with no prior knowledge about datas' inherent
structuring included in their design, they rely on auxiliary positional
information to help them encode notions of localization. This work relies on
transformer architectures to establish cross-modal links and to compose
environments in the stochastic latent space by learning distributions over
discrete constituents. Decompressing the modeled sequences back into the
continuous domain restores views of the sensors surroundings and allows to
visually examine the implications of combining two complementary sensor types.

The main contributions of this submission are as follows:
\begin{enumerate}
\item Presentation of a versatile and comprehensive multi-modal dataset featuring
  synchronized camera and low-level radar data capturing various real-world
  situations and traffic scenarios (section \ref{sec:data}).
\item Introduction of a novel method for compressing continuous sensor output
  into a stochastic sequence of integers via a consistent probabilistic
  self-supervised deep learning procedure (section \ref{sec:dvae}).
\item Derivation of a plausible and fully-probabilistic environment prediction
  through generation of RGB camera views based solely on streams of reliable and
  robust radio-frequency (RF) information (section \ref{sec:trafo}).
\end{enumerate}

\subsection{Sensor specifics and data representations}
\label{sec:radar}
Radar while being a powerful and versatile sensor for many applications has not
received quite as much attention as camera and lidar \cite{pfeuffer2019robust},
\cite{pfeuffer2018optimal} when it comes to the field of Deep Learning. This is
partially due to its intricate parameterization and inherently complex data
output which traditionally has to undergo numerous elaborate processing steps
before being accessible to further case-related tasks downstream. Yet, radars
key advantages cannot be excluded for much longer when it comes to automatic
navigation and other areas in which a safety-critical environment perception is
indispensable. Not only are radar sensors robust towards compromised lighting
conditions or adverse weather influences (e.g. rain, fog or snow) they also
allow for the direct and parallel measurement of radial distance and velocity of
objects in its field of view. As such, they present a valuable asset to the
familiar camera-based setup, bringing unique properties to the table that pure
vision sensors are lacking. The most common representation of radar measurement
results, employed within the field of autonomous driving, is in the form of
discrete data points representing the positions of potential obstacles
exemplified in red in Figure \ref{fig:rd_cam}. Each so-called target stands for
one specific reflection center of the sensors transmitted electromagnetic wave
and generally has further physical quantities inscribed:
\begin{figure}[h!]
  \centering \includegraphics[width=0.48\textwidth]{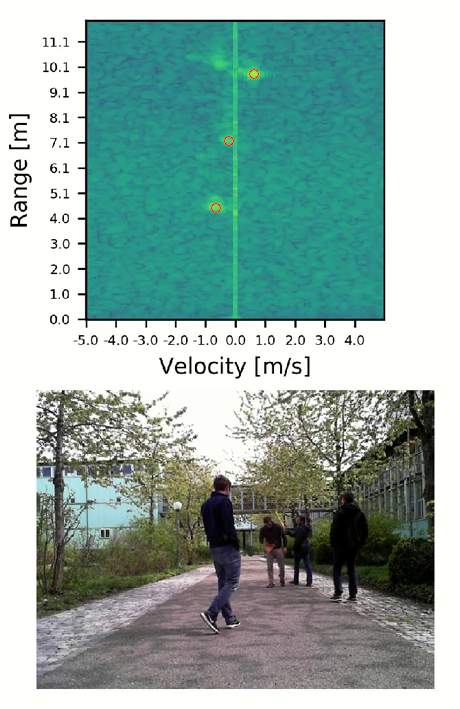}
  \caption{Dynamic scene depiction in range-Doppler and camera modality with
    superimposed radar targets in red. The latter are the result of laborious
    calculations depending on various estimates and thus typically prone to
    rejecting vital details of the measurements. As a case in point, the
    reflections of the fourth person in the far right back, present in the
    original recording, were discarded somewhere along the calculations and
    are absent from the high-level radar point cloud representation. Relying on
    the lower-level frequency plots instead retains most of the crucial
    information concealed within the data, including the fourth
    reflector. Colorization and axis labels of radar image given for
    illustrative purposes only.}
  \label{fig:rd_cam}
\end{figure}
The radar cross-section (RCS) as a measure of the objects specific reflectivity
characteristics and the associated relative radial velocity effectively
resulting in a four-dimensional vector per point. This description is practical
both in terms of intuition and data volume. Yet, it ignores a large amount of
vital information which has been discarded along the signal processing chain by
various assumptions, heuristic parameters and the algorithmic treatment
altogether. This outflow of potentially crucial information is termed
equivocation in information theory and commonly undesired yet in reality often
times unavoidable, given constraints in both memory and compute. This work aims
for reducing equivocation by drawing the data at an earlier stage\footnote{The
  term \emph{raw data} is intentionally omitted due to its inconsistent use in
  both community and scientific literature}, postulating that the preserved
information may be of value for subsequent algorithmic tasks and procedures. In
fact, the only processing the sensors raw voltage signals are subjected to are
multidimensional Fourier transforms (cf. section \ref{sec:radpostprocessing}),
mapping the time domain data into frequency plots as shown in Figure
\ref{fig:rd_cam}. The intensity distributions within these 1-channels images not
only indicate locations and objects reflectivity. They also provide an immediate
impression of approaching and removing targets by exploiting the Doppler
effect. This allows to separate equidistant objects based on the slightest
differences in movement illustrated by the two persons at a distance of \num{10}
meters in the displayed measurement. The vertical center line at zero velocity
includes static objects at all distances, whereas the left and right image
halves correspond to negative and positive radial velocities with respect to the
sensor. This technique thus offers discriminative features, camera stills and
their two-dimensional projections of the real world are missing. For although
camera shots offer favorable lateral and vertical resolutions, they usually lack
any depth information and computers still struggle with their
extraction. Then again, these so-called range-Doppler (rD) maps really only carry
meaning to humans either with associated axis labels detailing physical units or
in combination with another synchronized sensor as in Figure
\ref{fig:rd_cam}. Apart from differential Doppler reflections of extended
objects, no angular information is included in these specific plots, making them
complex to read at first glance. Resorting to this rather pure form of radar
data representation is justified though by the presumed existence of decisive
information concealed within which are not to be missed due to premature noise
assumptions and biased data cleansing. And although the memory demands necessary
to store these data formats is magnitudes higher than those for sparse point
clouds representations, the importance of the former cannot be
overstated. Indeed, its significance is not to be assessed manually but should
instead be evaluated by neural algorithms for which even the slightest preserved
signal constituent might prove essential.

\subsection{Foregoing explicit data annotations}
\label{sec:nolabels}
Evidently, radar data at this level are both regularly structured and highly
instructive and yet have not been given that much of attention within the
machine learning community thus far. This can largely be explained by the
difficulties touched on above but also because of the need for explicit
annotations required for supervised learning --- last decades dominant subfield
within deep learning. Annotating radar signatures at any level is complicated
due to its counterintuitive distribution patterns, especially if compared to
camera and lidar sensor output. Additionally, multi-reflections and multi-path
scattering as well as signal interference by electromagnetic radiation are just
some of the effects frequently observed in spectral representations further
contributing to their complexity. Annotation of radar point clouds most often
relies on semi-automated labeling pipelines in which certain radar targets
spatially coinciding with previously marked lidar points or camera frames are
assigned the corresponding label. A human then attempts to correct wrongly
classified points that inevitably result from imprecise calibrations and
erroneous time-synchronizations. As an undesirable side effect, this introduces
subjective judgment and individual bias into the data preparation, corrupting
the purity of the data. This is further aggravated by the fact that reference
camera stills commonly show occlusions or compromised views in adverse weather
which leaves precise annotations up to guess-work and rough estimates. Manual
intervention should therefore be abandoned or at least kept to a bare
minimum. Labeling radar frequency plots, though, is a futile endeavor both in
terms of time expenditure and data assessment. Making sense of abstract raw
intensity maps is difficult and relating the many local maxima to traffic scenes
based on associated camera footage appears to be virtually infeasible for even
the simplest of situations. Still, recent years saw occasional publications of
datasets, including annotated radar information in an attempt to close the gap
to the other sensors for which multiple benchmarks have already existed for
years. However, to simplify the described problems, those collections often
feature only scarce amounts of radar data and labels given on discrete
target-level \cite{caesar2019nuscenes}, \cite{meyer2019automotive}. This is turn
necessitates particular methods like PointNets \cite{qi2017pointnet} for further
processing. Some sets expose low sample rates and imbalanced object class
distributions whereas other records do not include Doppler information
\cite{sheeny2020radiate}, \cite{kim2020mulran}, depriving themselves of radars
unique feature altogether. Only recently, first radar datasets featuring
annotations on the frequency level were made publicly available
\cite{quaknine2021carrada}, \cite{zhang2021raddnet}, \cite{wang2021rodnet} but
it remains to be seen if these are going to have a similar impact on the
community and will incite comparable research efforts as the famous KITTI
\cite{geiger2013vision} and Cityscapes \cite{cordts2016cityscapes} benchmarks
did for vision-based scene-understanding. Both aspects, the tedious annotation
of radar data followed by an inevitable inflow of misinformation and the
preferable elimination of equivocation call for completely different approaches,
establishing the subfield of self-supervised learning. This domain dispenses
with the traditional dependence on explicit labels and instead focuses on
finding patterns and inherent structures within data by using parts of it as
stimulus to supervise the remaining portion. It is suspected that
self-supervised methods can exploit vastly more signals than conventional
supervised learning approaches since the neural networks are no longer told what
they are to infer. Rather, they have to find meaning in the input by themselves
and construct high-dimensional feature vectors based only on the original
data. Also, self-supervision encourages the algorithms to look for
cross-correlations in multi-modal data in a more natural way, aligning nicely
with the measurement setup used in this work. In fact, the feedback provided by
low-level sensor output should be tremendous and surpass that of processed data
records and explicit labels by large margins. As an additional appeal, unmarked
data already exists or can be generated almost at will and with little cost. As
such, self-supervised learning seems to be a natural fit for the regime of radar
data in which labels are rare and hard to come by. Consequently, this
publication delves into this fields mathematical and algorithmic intricacies in
quite some detail, trying to take advantage of its full potential.

\subsection{Related Work}
In recent years, quite a few publications dealt with deep learning methods
applied to radar data at various levels, aiming to expand the knowledge about
the systems environment in one way or another. A first broad classification
distinguishes between approaches on synthetic aperture radar (SAR) data whose
image-like format is readily suitable to all sorts of pattern recognition models
and continuous wave (CW) radar output, which is usually not immediately
applicable for machine learning purposes. All projects can further be
partitioned into supervised approaches which rely on explicit ground truths of
some sort as described above and annotation-free methods which are most relevant
to the present work. Among the latter are self-supervised techniques using
little but the original radar data distributions to denoise micro-Doppler
signatures, as shown in \cite{abdulatif2019towards} or enhance frequency plot
resolutions demonstrated in \cite{armanious2019adversarial}. The authors
construct a complex parameter fitting procedure that involves a Unet-like
generator and patchGAN-reminiscent discriminator to tell apart clean and noisy
spectra. To this end, they combine perceptual loss functions with a
reconstruction criterion and perform adversarial training, learning to retain
high-frequency details and remove unwanted artifacts. Ultimately, the authors
advocate for resorting to neural algorithms for the outlined tasks as those
arguably yield superior results compared to traditional thresholding or
intensity correction schemes. A similar approach was followed by the authors in
\cite{rock2019complex} and \cite{rock2020deep} where convolutional architectures
are introduced for the purpose of reducing noise and interference in automotive
radar spectra. CNN-based models are applied to the complex coefficients rather
than their modulus at different stages along the signal processing chain. This
allows to obtain a clear gain in information amount at the cost of added
complexity and an increase in data volume. The devised neural approaches are
input a combination of both real and simulated data for ease of retaining clean
spectra for output comparison. Finally, those are benchmarked against
conventional noise suppression techniques in an effort to demonstrate their
benefits in terms of peak preservation and improvements in signal-to-noise
ratio. Perhaps the work most related to this project has been described in
\cite{zhao2018through} by inferring human poses occluded by obstacles merely
through Wi-Fi signals. In contrast to the above mentioned publications which
perform intra-modal supervision, the authors present an approach which exploits
time-synchronized camera data for supervision at train time in a teacher-student
setup. This incites the network to acquire cross-modal correspondence features
which, at test time, effectively allow for the sensing of human dynamics through
walls and other barriers. This certainly raises some ethical questions, but also
highlights the powerful nature and promising future of self-supervised systems
without manual interference or exposure to human influence. All things
considered, the previously addressed publications present strong arguments for
self-supervised methods and underline the above stated claim of \textit{leaving
  the data alone} and letting the neural structures decide for themselves which
information to attend to the most.

\section{Data Collection and Sensor Settings}
\label{sec:data}
\begin{figure}[h!]
  \centering
  \includegraphics[width=0.48\textwidth]{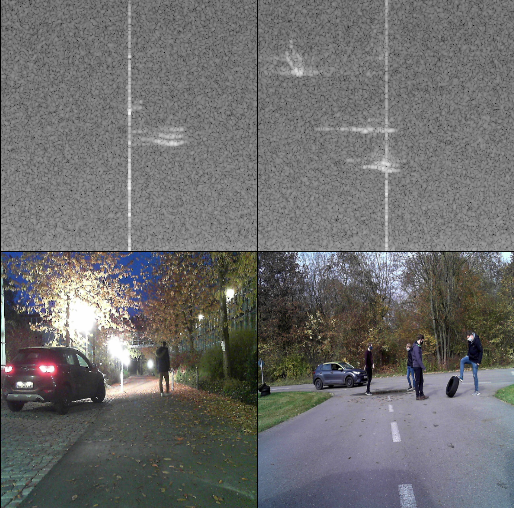}
  \caption{Synchronized radar and camera images of different outdoor scenes used
    as input for the neural algorithms, which will be described later. In case
    of extended objects, multiple reflections are easily distinguishable and a
    trademark of highly resolved range-Doppler maps. Moving parts like spinning
    wheels and human limbs show clearly through differences in relative
    velocities compared to their main compound objects. The plots underline the
    benefits of resorting to low-level sensor information for the multi-modal
    fusion process.}
  \label{fig:repr}
\end{figure}

The presented experiments were conducted on a custom dataset comprising roughly
\num{50000} samples of time-synchronized radar and camera images. The collection
captures diverse real-world scenery around Ulm --- Germany, varying in terms of
both weather and lighting conditions. It features all kinds of realistic traffic
scenarios ranging from clusters of pedestrians over lost-cargo situations and
oncoming vehicles to the passing of trams and buses. The recordings were taken
across the change of seasons during the course of one year to render them more
versatile and expressive and to include a maximum of environmental diversity. A
typical excerpt of the gathered data is shown in Figure \ref{fig:repr} with more
samples of both domains available at
\href{https://cditzel.github.io/GenRadar/}{cditzel.github.io/GenRadar} in
detail and temporal succession.
Both types of sensors were operated with a frame rate of \SI{100}{\milli\second}
which was found to present a decent compromise between covering the intrinsic
dynamics of outside scenes and limiting the amount of data to be stored. The
entire multi-modal dataset was collected with an experimental stationary sensor
setup to avoid having to compensate for ego-motions, which is no trivial task at
the chosen data representation level. Also, no extrinsic calibration was
performed prior to any recording, but both sensors were spatially correlated by
accurately aligning their horizontal field of views. This, however, does not
prevent instances from being captured by one modality only due to different
measurement ranges and aperture angles of both sensors. In fact, camera shots
occasionally show distant objects beyond the chosen maximum range of the radar,
whereas objects in the lateral fringe areas turn up via frequencies but not in
the camera. This renders the entire dataset quite realistic but also extremely
complicated in terms of mutual correspondence learning. The assembly streams
about \SI{70}{\mega\bit\per\second} of RGB data and
\SI{40}{\mega\bit\per\second} of sampled RF data for a single enabled receive
antenna so that about \SI{14}{\mega\byte} were written to disk every second,
emphasizing the extraordinary memory demands when working with low-level sensor
data. Typical sequence lengths last from \SIrange{10}{60}{\second} capturing
both short snippets and longer periods of random outdoor interactions.

\subsection{Details on Radar Parameterization}
\label{sec:radparameterization}
\begin{figure}[h!]
  \centering
  \includegraphics[width=0.48\textwidth]{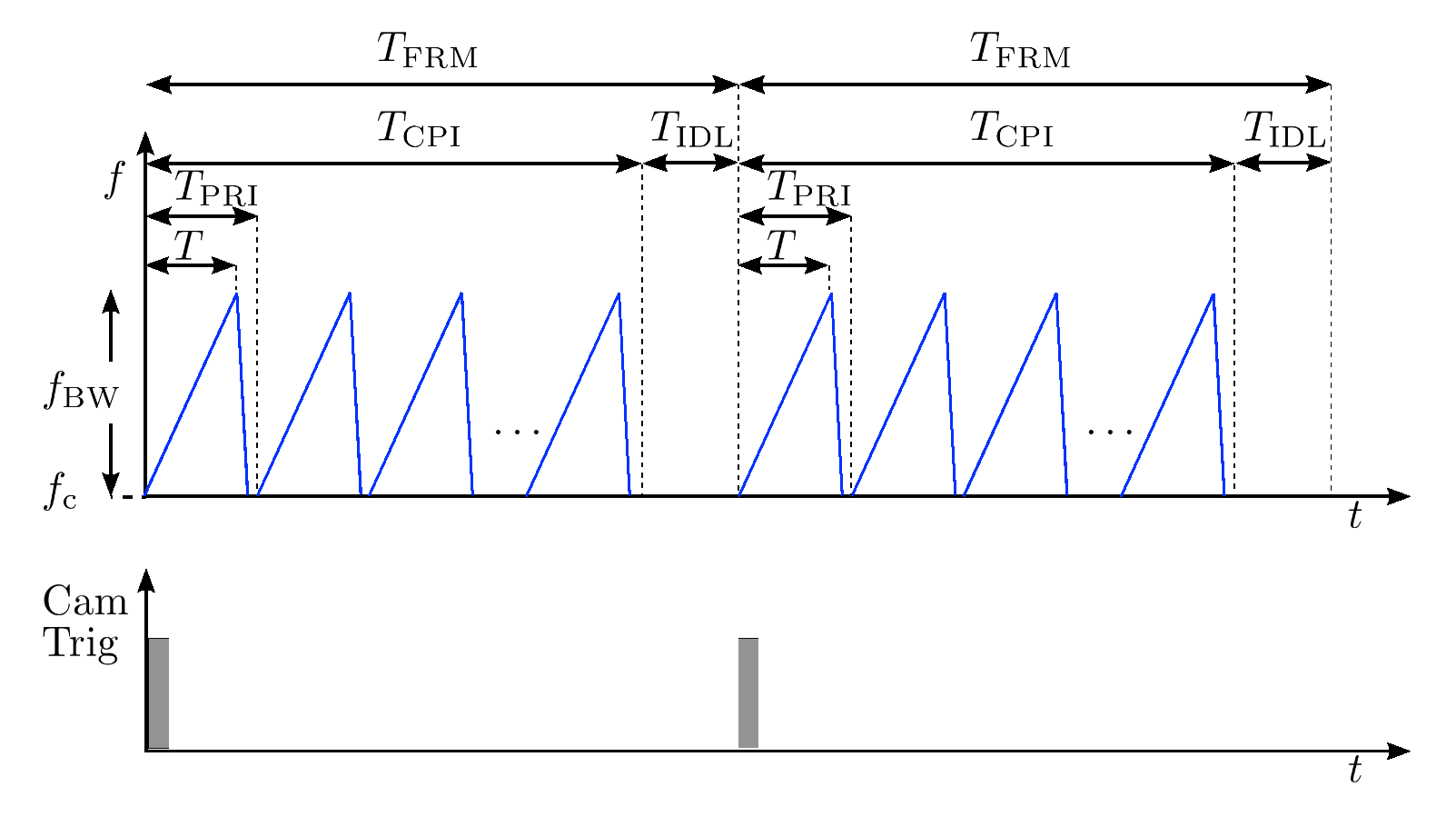}
  \caption{Ramp-synchronous sampling of radar and camera data across two
    consecutive chirp sequences resulting in two camera-aligned rD images
    $T_{\text{FRM}}$ seconds apart.}
  \label{fig:radparam}
\end{figure}

The parameterization of chirp-sequence radars is notoriously complicated and
involves the careful adjustment of several conflicting parameters which
determine the maximum of both range and velocity as well as corresponding
resolutions. Changing one parameter to improve e.g. the maximum detectable range
often causes the acquisition of another measurand to deteriorate so that special
care has to be taken to find some middle ground suitable for the task at
hand. The MIMO-FMCW radar\footnote{RadarLog@www.inras.at} used to obtain the
data for the present submission features a \SI{77}{\GHz} front-end and offers a
multitude of configuration possibilities in terms of both signal modulation and
data processing. The specific parameters selected for the gathering of this
dataset are outlined in Table \ref{tab:radparam} for reference. Here only a
brief abstract of the central sensor parameters, related physical quantities and
measurement characteristics are given, crucial for the understanding of the data
representations used in the following. For further information on how to derive
the associated mathematical relations and working principles of a chirp-sequence
radar as well as their interplay see \cite{brooker2005understanding} and
\cite{winkler2007range}. The concrete values for the numerous properties are of
exemplary nature and represent a sensor configuration that was found to work
reasonably well for the recording of the dataset without pushing the sensor to
its limits.
\begin{table}[t]
  \caption{Parameterization for chirp sequences of electromagnetic radiation.}
  \centering
  \begin{tabular}{lcS[table-number-alignment = center]s}
    \toprule
    {Physical quantity} & {Acronym} & {Value} & {Unit} \\
    \midrule
    Transmit power &$P_{\text{TX}}$ & 60 & mW \\
    Carrier frequency &$f_c$ & 76 & \giga\hertz \\
    Wavelength &$\lambda$ & \si{\sim}4 & \milli\meter \\
    Bandwidth & $f_{\text{BW}}$ & 1.5 & \giga\hertz \\
    Sampling period & $T$ & 51.2 & \micro\second \\
    Number of samples & $N$ & 512 & [-] \\
    Sampling frequency & $f_s$ & 10 & \mega\hertz \\
    Frequency resolution & $\Delta f$ & 19.53 & \kilo\hertz \\
    Sampling interval & $\Delta t$ & 100 & \nano\second \\
    Pulse repetition interval & $T_{\text{PRI}}$ &90& \micro\second \\
    Pulse repetition frequency & $f_{p}$ &11.11& \kilo\hertz \\
    Coherent processing interval & $T_{\text{CPI}}$ & 46.08 & \milli\second \\
    Idle time & $T_{\text{IDL}}$ & 53.92 & \milli\second \\
    Frame time & $T_{\text{FRM}}$ &100& \milli\second \\
    Sensor frequency & $f_{\text{MEAS}}$ &10& \hertz \\
    Center frequency & $f_{0}$ &76.75& \giga\hertz \\
    Max. Doppler frequency & $f_{\text{D}}$ &5.55& \kilo\hertz \\
    Number of chirps & $K$ &512& [-] \\
    \bottomrule
  \end{tabular}
  \label{tab:radparam}
\end{table}
Starting from a carrier frequency $f_c=\SI{76}{GHz}$ a directed electromagnetic
wave is continuously emitted via a transmitting antenna, with its frequency
being linearly modulated across a large bandwidth of
$f_{\text{BW}}=\SI{1.5}{GHz}$ during the sampling period
$T=\SI{51.2}{\micro\second}$. Fractions of this radiation are scattered by
objects in the wave's path and reflected back into the direction of the receive
antennas, which collect the energy with certain delays due to round-trip times
before converting it into a fluctuating voltages. Mixing both transmitted and
received voltage signals (homodyne processing) followed by low-pass filtering
results in a narrow-band Intermediate Frequency (IF) signal with a bandwidth
orders of magnitudes lower than the swept frequency range.
\begin{figure}[h!]
  \centering
  \includegraphics[width=0.45\textwidth]{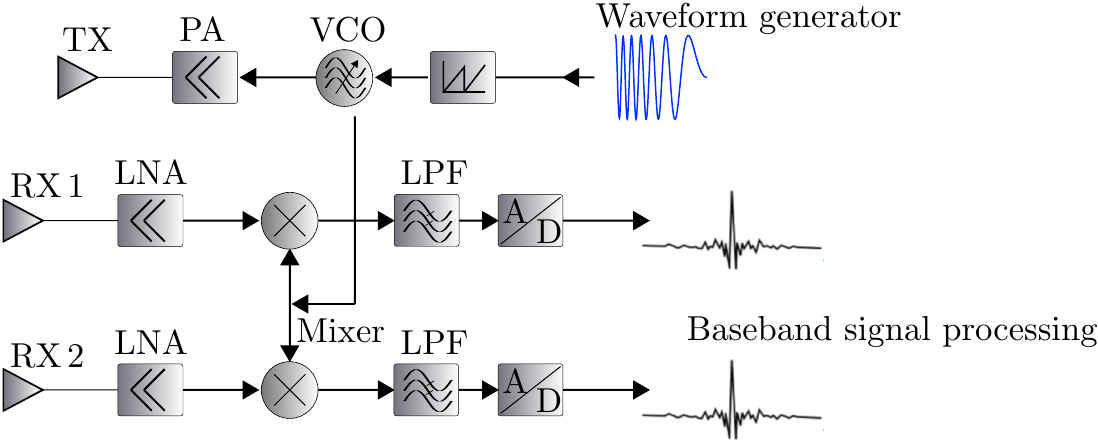}
  \caption{MIMO homodyne radar measuring principle. The frequency modulated
    voltage signal is amplified and sent into space through the transmit antenna
    (TX). Reflected portions of the signal are collected by the receive antennas
    (RX), enhanced and correlated with the transmitted signal to deduce range
    and velocity information.}
  \label{fig:components}
\end{figure}
The sampling rate of the Analog-to-Digital Converter (ADC) can thus be chosen to
a moderate $f_s=\SI{10}{MHz}$ capturing this time-series equidistantly with
$N=512$ samples during every up-chirp period. The measuring process is
illustrated in Figure \ref{fig:components} for an exemplary setup of two receive
antennas. Due to the modulation, the individual frequency components that
superimpose within this approximately sinusoidal IF-signal relate directly to
objects distances in the sensors field of view and the amplitudes correspond to
their respective reflection strengths. The frequency excursion is then reversed
for a short period of time with no further sampling taking place before ramping
the frequency up again. This procedure repeats every pulse repetition interval
$T_{\text{PRI}}=\SI{90}{\micro\second}$. A total of $K=512$ precisely timed
chirps are transmitted in this manner within the so-called coherent processing
interval $T_{\text{CPI}}= \SI{46.08}{\ms}$ before entering into a short idle
time $T_{\text{IDL}}$. This so-called ramp-synchronous sampling process,
depicted in Figure \ref{fig:radparam}, is of utmost importance as it allows to
identify infinitesimal deviations within consecutive IF-signals.
More precisely, sequences of multiple chirps taken in close succession exhibit
small gradual phase shifts in case of relative movement between the
electromagnetic source and objects in the wave's path. Doppler frequencies
quantify the rate of those phase changes and correspond linearly to respective
relative velocities. The formerly detailed sampling process takes place within a
frame time of $T_{\text{FRM}}=\SI{100}{\milli\second}$ which constitutes the
sensor measuring frequency of $f_{\text{MEAS}}=\SI{10}{Hz}$. The selected
parameterization given in Table \ref{tab:radparam} defines the following
measurement relations for the range dimension
\begin{align}
  \label{eq:rangereso}
  \Delta r &= \frac{c}{2f_{\text{BW}}} = \SI{0.1}{\meter} \\
  r_{\text{max}} &= \frac{cT}{4f_{\text{BW}}} f_s = \frac{N\Delta r}{2} = \SI{25.6}{\meter}
                   \label{eq:maxrange}
\end{align}
and the velocity dimension in mutual interdependence
\begin{align}
  \label{eq:velreso}
  \Delta v &= \frac{\lambda}{2T_{\text{CPI}}} = \pm \SI{0.042}{\meter\per\second} \\
  v_{\text{max}}  &= \frac{c}{4f_0}f_{p}= \frac{K \Delta v}{2} = \pm \SI{10.8}{\meter\per\second}.
                    \label{eq:maxvel}
\end{align}
These quantities span a virtual uniform measurement grid with an extent of
$r_{\text{max}}$ in longitudinal and $\pm v_{\text{max}}$ in lateral direction
limited by the sampling frequency $f_s$ of the ADC and the pulse repetition
frequency $f_{p}=1/T_{\text{PRI}}$ respectively. The coverage area is subjected
to the Nyquist-Shannon bound so that targets beyond cause aliasing of the signal
spectrum as well as ambiguous estimates for range, velocity or both. The range
resolution $\Delta r$, the distance at which two targets are still marginally
discernible, is dictated by the swept bandwidth of the frequency
modulation. This entity effectively discretizes the range dimension into
equidistant range gates so that multiple targets located within the same gate
segment will be indistinguishable. The velocity resolution $\Delta v$ is the minimum
relative difference in speed between two equidistant targets that can be
identified and defines the cell width in lateral direction. It can be enhanced
by prolonging the continuous transmission time of the periodic signal
$T_{\text{CPI}}$, and thereby the entire duration of energy integration either
through a larger number of chirps or a longer duration thereof. As a result, the
velocity resolution depends largely on the modulation and can thus be increased
almost arbitrarily. However, reducing the cell size for better resolution in
either range or velocity shrinks the grid extent in the corresponding dimension
and hence decreases the maximum range or velocity detection capability. The
parameters in equation \eqref{eq:rangereso} to equation \eqref{eq:maxvel}
therefore constitute a compromise between the accurate acquisition of
discriminative Doppler and IF signals and a sufficiently large field of view to
also capture distant real-world events. The total information accumulated by
above design choices is written to file in the form of $N$ discrete samples of
the IF-signal for every consecutive chirp $K$ yielding a \SI{16}{\bit} data
matrix $\bm x_{\text{IF}}\in \mathbb{N}^{K\times N}$ for every chirp sequence within
one recording.

\subsection{Details on Camera Parameterization}
The monocular camera footage was captured in unsigned \SI{8}{\bit} $3$-channel
RGB format with a resolution of $\bm x \in \mathbb{N}^{480 \times 640\times 3}$ pixels and no further
processing was applied prior to serialization. In particular, no steps were
taken to reduce glaring reflections, blooming effects or to remove any imaging
artifacts.

\subsection{Details on Radar Preprocessing}
\label{sec:radpostprocessing}
\begin{figure}[h!]
  \centering
  \includegraphics[width=0.5\textwidth]{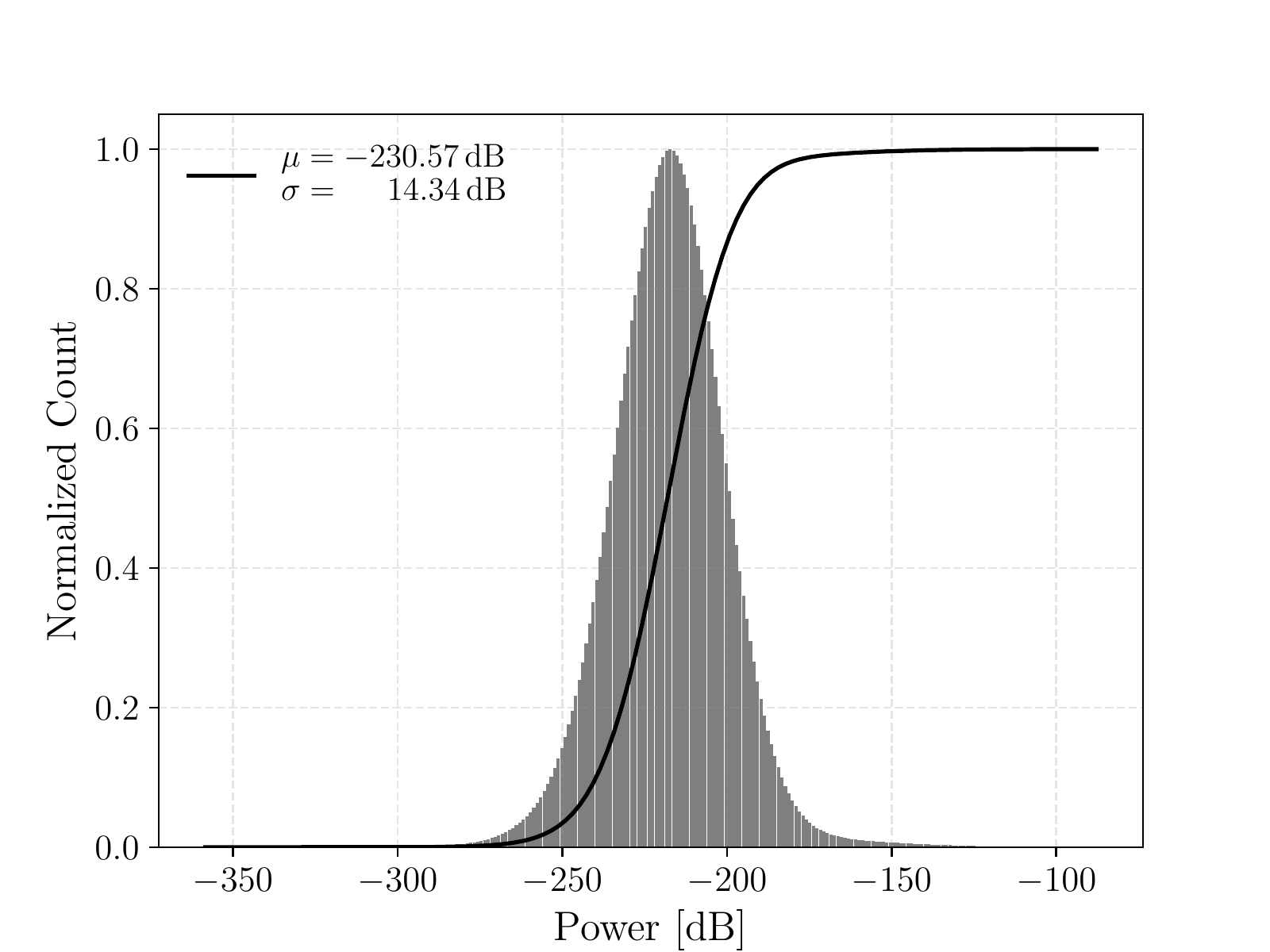}
  \caption{Intensity distributions and first two moments for all range-Doppler
    maps contained within the training dataset.}
  \label{fig:rad_histogram_train_pure}
\end{figure}

\begin{figure}[h!]
  \centering
  \includegraphics[width=0.5\textwidth]{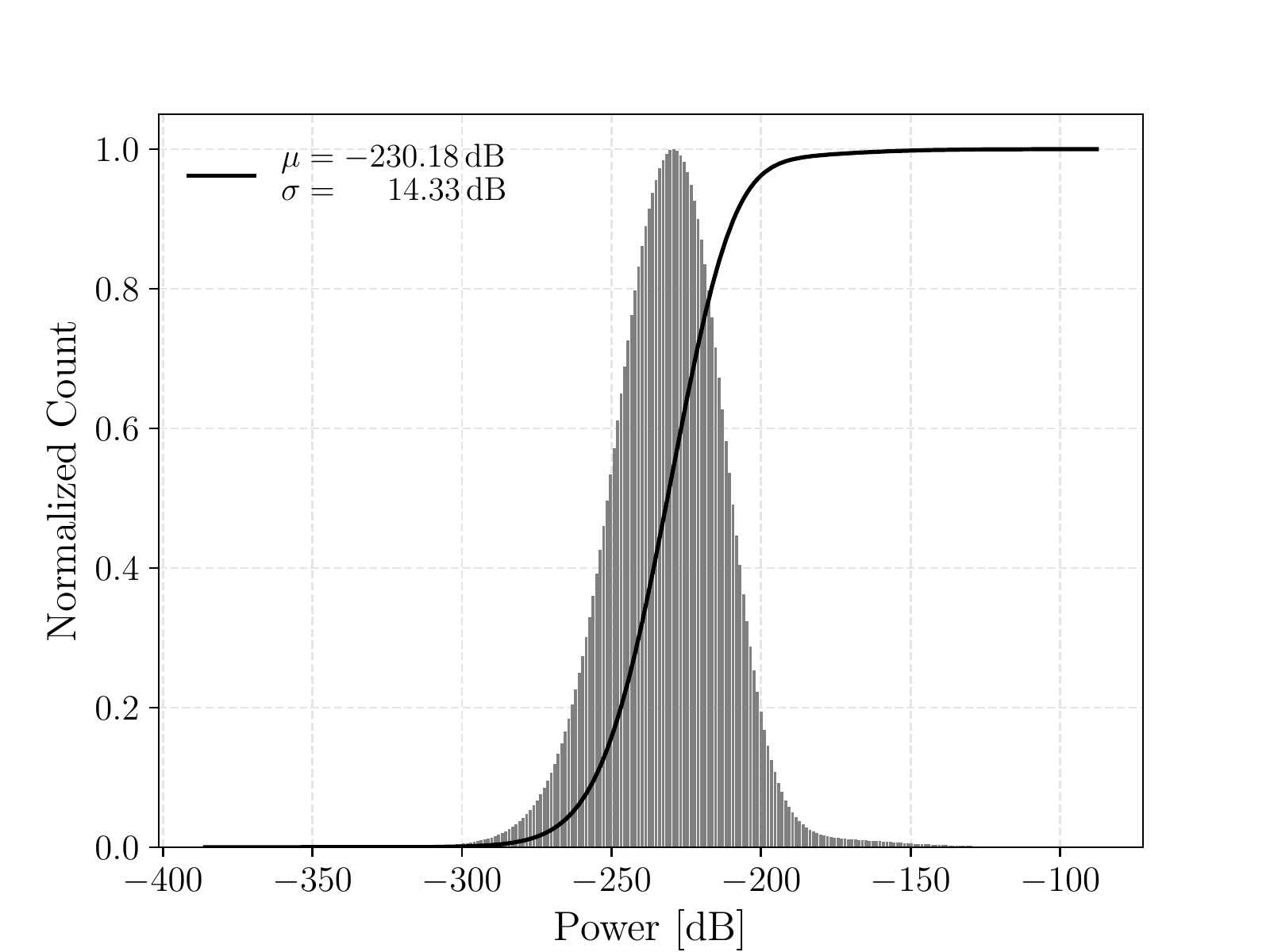}
  \caption{Intensity distributions and first two moments for all range-Doppler maps
    contained within the validation dataset.}
  \label{fig:rad_histogram_valid_pure}
\end{figure}

The notion of \emph{imaging radar} stems from the discretization described in
section \ref{sec:radparameterization} and the structural resemblance of the data
to natural images despite its vastly more abstract and counterintuitive
content. To extract the range and velocity information defined in equations
\eqref{eq:rangereso} to \eqref{eq:maxvel} onto a regular grid, the de-serialized
time-series undergo a number of frequency decomposition steps which are applied
to each collated data matrix $\bm x_{\text{IF}}\in \mathbb{N}^{K\times N}$ of every
frame under consideration. A single row of this matrix
$\bm x^k_{\text{IF}}\in \mathbb{N}^{N}$ contains the entire intermediate frequency
signal captured during one frequency burst. Successive columns
$\bm x^n_{\text{IF}}\in \mathbb{N}^{K}$ represent individual samples taken at
coinciding points in time within each sampling period across consecutive
chirps. In line with the arguments given in Section \ref{sec:radar}, processing
of these matrices is kept to a minimum and merely involves finding the maxima of
2D spectra of the 1D time-series, representing the unknown differential and
Doppler frequencies. According to mild requirements, formulated in
\cite{wojtkiewicz1997two}, the optimal estimator of unknown frequencies within
harmonic signals, observed during a finite time interval, are the maxima
locations of the complex modulus applied to multidimensional Fourier
spectra. The following steps are therefore performed in order and applied in
parallel to the low-level data matrices.
\begin{enumerate}
\item A two-dimensional Hanning window is applied as trade-off between a decent
  side-lobe suppression and sufficient amplitude preservation in the frequency
  domain which comes at the expense of slightly broadened main lobes, see
  \cite{scholl2016exact} for further details. This also reduces spectral leakage
  at the boundaries by mitigating the implicit assumption of
  infinitely-continuing signals included in the Fourier transform. The
  diminished signal peaks are later compensated for by appropriate window gain
  corrections to recover the signals true amplitudes.
\item An appropriate choice of zero-padding in both dimensions is made, which
  not only improves the granularity of the following frequency estimations, but
  also offers an opportunity to alter the resulting image extents as this
  modifies the frequency bin sizes which correspond to individual radar image
  pixels.
\item A Fast Fourier Transform \cite{cooley1965algorithm} over the discrete
  samples of each chirp yields a matrix of complex coefficients
  $\bm x_{\text{R}}\in \mathbb{C}^{K\times (N/2+1)}$ with half the number of columns
  due to redundant spectral points mirrored along the Nyquist frequency $f_s/2$
  for real-valued transforms. The elements in every row
  $\bm x^k_{\text{R}} \in \mathbb{C}^{N/2+1}$ now correspond to range bins (except
  for every first ones being the DC offset) and are multiplied by compensation
  factors to make up for window and transform-specific signal reductions.
  Taking the complex modulus, these values would immediately disclose radial
  distances to reflective obstacles reached by the electromagnetic wave within
  every chirp.
\item Additional Fourier transforms over the phasors along columns are
  tantamount to time derivatives of the aforementioned phase shifts across
  chirps. The matrix $\bm x_{\text{RD}}\in \mathbb{C}^{K\times (N/2+1)}$ then reveals
  unique Doppler frequencies for every range bin. These translate to signed
  radial velocities for possibly multiple equidistant reflectors with varying
  speeds relative to the sensor. To restore the original signal peaks, these
  quantities are again compensated for due to the previous windowing. A
  subsequent shifting locates the zero-Doppler bin at the center of the
  frequency diagram. The graphical summary of this two-dimensional Fourier
  transform processing is illustrated in Figure \ref{fig:2dfft} in the appendix.
\item What follows is the extraction of desired radial distances within a range
  of \SIrange[range-units = single]{1.3}{25.6}{\meter} and Doppler information
  within the interval $\pm\SI{2.72}{\meter\per\second}$ via a submatrix
  $\bm x_{\text{RD}}\in \mathbb{C}^{256\times 256}$. This choice respects the limits
  given by equations \eqref{eq:maxrange} and equation \eqref{eq:maxvel} and
  reduces the compute cycles of the upcoming neural operations considerably.
\item The computation of the power spectrum by squared modulus calculation of
  the complex coefficients ultimately results in a single-channel scalar-valued
  range-Doppler representation for every chirp sequence. A final log computation
  compresses the spectrum as it generally contains both very large and
  comparatively small entries and brings it to the more meaningful decibel scale
  $\bm x_{\text{rad}} = 20\cdot\log_{10}\lvert \bm x_{\text{RD}}\rvert^2\in \R^{256\times
    256}\quad\forall \bm x_{\text{rad}}\in \mathcal{X}_{\text{rad}}$.
\end{enumerate}
The obtained intensity distribution provides insight into objects radial
distances, velocities and reflectivity characteristics, with the latter
depending on various factors like aspect angle, polarization, geometric extent
and material properties. The proper selection of modulation parameters and the
diligent discretization of the measurement grid in both dimensions even enables
the detection of multiple reflection centers originating from one and the same
object. This becomes evident with regard to Figure \ref{fig:repr} in which
several maxima in the rD maps are allocated to the wheels of the
vehicle or the extremities of the pedestrians. This observation serves as yet
another justification for resorting to low-level radar data. Particularly the
ability to discriminate structures via their Doppler spectrum proves powerful
and informative for the following approaches. A limiting, but rather theoretical
aspect in this regard is the implicit requirement that the radial velocity of
all targets in the radar's field of view is sufficiently low so as not to having
them transition from one range bin to another during the course of one chirp
sequence. Violation of this bound, called range-cell migration, impairs the
accuracy of range and velocity estimates, most noticeable by spectral
intensities being blurred-out across a number of rD cells. Figure
\ref{fig:rad_histogram_train_pure} and Figure \ref{fig:rad_histogram_valid_pure}
report the radar intensity distributions of both train and validation set after
the transformations. Prior to neural data processing, the frequency plots are
transformed to the interval $\bm x_{\text{rad}} \in [-1,1]$ to enhance training
efficiency and diminish the impact of outliers in the datasets.

\subsection{Details on Camera Preprocessing}
\label{sec:campostprocessing}
\begin{figure}[h!]
  \centering
  \includegraphics[width=0.5\textwidth]{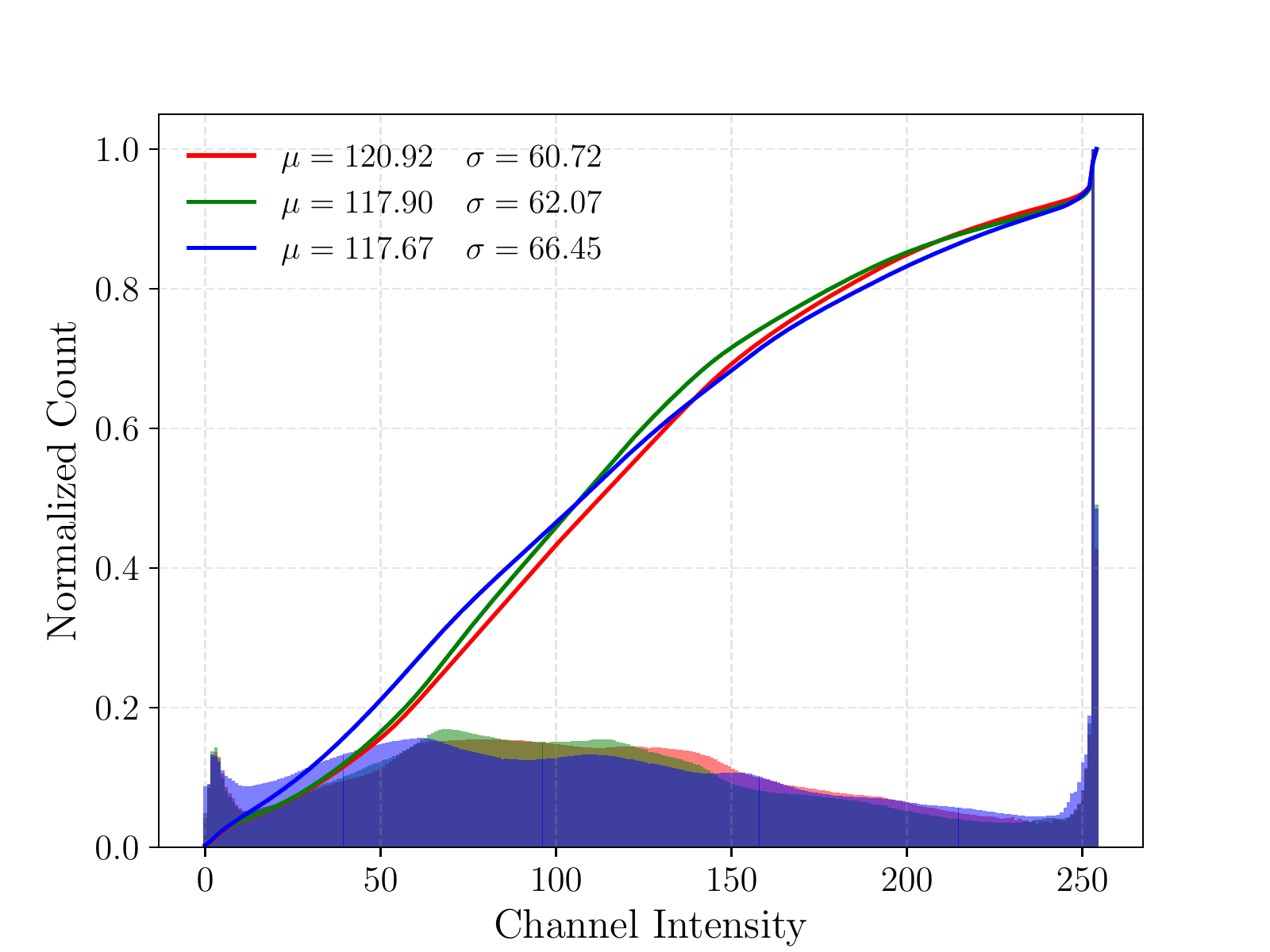}
  \caption{Pixel distributions and first two moments for RGB camera images
    contained within the training dataset.}
  \label{fig:cam_histogram_train_pure}
\end{figure}

\begin{figure}[h!]
  \centering
  \includegraphics[width=0.5\textwidth]{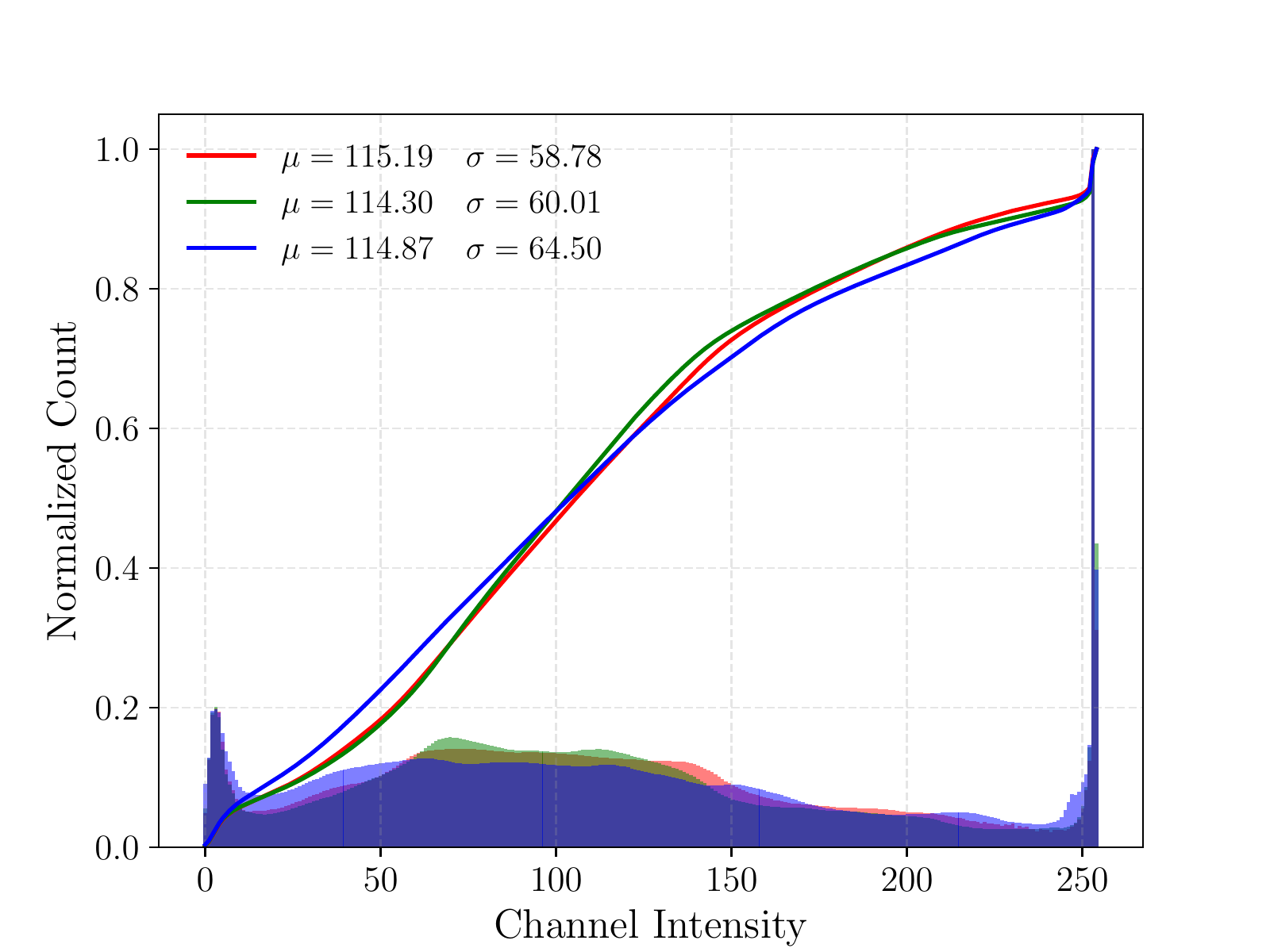}
  \caption{Pixel distributions and first two moments for RGB camera images
    contained within the validation dataset.}
  \label{fig:cam_histogram_valid_pure}
\end{figure}

The devised data processing pipeline loads batches of synchronized multi-modal
records from disk in parallel, using them as input for the neural network
training, described in subsequent chapters. Upon deserialization, the camera
images are channel-wise re-scaled from $\bm x_{\text{cam}} \in [0,255]$ to the
interval $\bm x_{\text{cam}} \in [-1, 1]$ for this centering gives better
conditioning of the numerical objective later on. As a next step, all data are
resized to
$\bm x_{\text{cam}} \in \R^{256\times 256\times3}\quad \forall \bm x_{\text{cam}}\in
\mathcal{X}_{\text{cam}}$ reducing their dimension and making it consistent with
the extend of the rD maps in the radar subset $\mathcal{X}_{\text{rad}}$. The
channel-wise pixel histograms of the entire collection are given in Figure
\ref{fig:cam_histogram_train_pure} and Figure \ref{fig:cam_histogram_valid_pure}
alongside their cumulative distributions separated by train and validation split
respectively.
It is worth noting that no form of data cleansing is employed in any stage of
signal preparation. In particular, no precautions are taken to guarantee the
existence of central objects or salient features within every single multi-modal
sample. Scenes may occasionally contain no dynamics at all, while other snippets
are likely to show large groupings of moving individuals. This is to prevent the
attention of the neural processing from becoming biased towards certain image
regions and placing focus only on the most prominent parts within the input.
Periodically, objects are clearly resolved in the camera image, while they are
being lumped together in the corresponding rD map due to identical distances and
similar velocities to the sensor. At other times, objects are clearly
discernible through their reflections in the frequency plot but occluded in the
camera image. Consequently, this dataset may be considered rather realistic,
especially compared to the many polished benchmarks that exist nowadays. At the
same time, this makes the presented algorithmic approach all the more
challenging and hard to compare to existing methods. The entire dataset
consisting of \num{50000} dual-domain samples was split into \num{43000} train
samples, leaving recordings of \num{7000} samples for validation. Importantly,
the division was done inter-sequentially, assigning only complete sequences to
either set and not breaking recordings apart. It is expected, however, that this
decision renders the entire task defined in section \ref{sec:introduction} much
more difficult.

\section{Methodology}
\label{sec:method}
The overall objective of explaining a captured radio-frequency spectrum through
the probabilistic synthesis of an associated RGB camera image is ambitious due
to the high-dimensional data of both domains. The most successful related
approaches of deep conditional generations can thus far be found in the field of
natural language processing. Here, byte-pair encoded words or fractions thereof
are used, usually contained within finite-sized vocabularies. As sentences
generally comprise only a small number of words, this drastically reduces the
dimensionality of the problem, as predictions can only take on so many discrete
possible outcomes. The situation is vastly different for natural images or
spectral plots in which data points can take on a continuous, theoretically
infinite range of values within certain intervals. Additionally, those forms of
input notoriously exhibit a lot of noise and redundancies making it more
difficult to extract relevant information. Consequently, previous image
generation methods had to rely on massive amounts of computation power and
restrict themselves to relatively small image sizes, as was done by the authors
in \cite{chen2020generative}. Coping with the above challenges in a unique and
seminal way was recently demonstrated by the work described in
\cite{ramesh2021zero}. The authors successfully combined descriptive captions
with associated compressed image components during autoregressive trainings in
an attempt to render expressive, high-quality visuals based on content
description in human language. Not only did they collect an unprecedented amount
of image-text pairs, they also employed an unheard-of amount of compute to reach
this goal. Moreover, they took advantage of the many tokenization schemes and
predefined embeddings that exist for natural language processing and so had to
discretize only one of two modalities, which they carried out in a
all-encompassing probabilistic manner. More recently, a different group of
researchers proved in \cite{esser2020taming} token-based image generation
successful also for large-scale, high-resolution images if conditioned on a wide
variety of accompanying visuals like depth images, semantic maps or salient
objects within corresponding images. The authors had to come up with a way to
jointly discretize two continuous high-dimensional modalities and decided on a
deterministic vector quantization method \cite{oord2017neural} followed by a
patchGAN-style discriminator downstream for further improvements in image
construction quality. Another remarkable aspect of their report was the
execution of their approach on a relatively modest hardware setup, especially if
compared to the work of \cite{ramesh2021zero}.

The present work follows along the steps of those projects but employs two
physical measuring principles naturally related through the propagation of
electromagnetic waves for information retrieval. The entire proposed algorithm
comprises two disparate stages which are both trained end-to-end in a
self-supervised fashion without the need for expensive and time-consuming
annotations. In the first stage, RGB and RF data are probabilistically
decomposed into separate discrete constituents in a holistic and
mathematically-consistent way by means of an effective convolutional
compression-decompression scheme described in section \ref{sec:dvae}. Both
highly-compressed multi-modal data streams are then used in the second phase, to
autoregressively train an attention-based model and stochastically predict
camera-associated tokens conditioned on radio-frequency information as explained
in section \ref{sec:trafo}. To this end, the holistic nature of the previous
encoding process is abandoned and only the compression part is retained for
discretization purposes. Upon successful sequence construction, the decompresser
decodes the modeled integer series into a coherent and expressive camera view,
relevant for several important use cases. Only after the training of both stages
has been completed are predictions performed whose quality is assessed in
section \ref{sec:synthesis} in detail.

\subsection{Probabilistic Discretization of the Latents}
\label{sec:dvae}
Variational autoencoders, short for VAEs, are a fundamental concept in the field
of modern self-supervised learning. Originally introduced in
\cite{kingma2014auto} to perform variational inference and scalable expectation
maximization \cite{bishop2006pattern} on large datasets, their convolutional
counterparts have found their ways into numerous practical and scientific
applications like representation learning and density estimation. As
representatives of so-called \textit{directed generative models}, they allow
unconditional sampling of new data from a parametric approximation
$p_{\bm \theta}(\bm x)$ to the true underlying distribution $p^*(\bm x)$ which
observations $\bm x$ are taken to originate from. Data-driven iterative maximum
likelihood estimation (MLE) of the model parameters $\bm \theta$ causes the marginal
approximation to successively converge to the unknown density so that
\begin{align}
  p_{\bm \theta}(\bm x) \approx p^*(\bm x) \quad\forall \bm x\in \mathcal{X}.
\end{align}
Here, VAEs primary application will be the transformation of high-dimensional
physical measurements into discrete low-dimensional equivalents. Also termed
\textit{deep latent variable models}, they are effectively used to perform
variational inference over latent variables $\bm z$ assumed to be generating the
observed data. The introduction of these latents increases model complexity and
expressiveness so that the resulting joint distribution is capable of
approximating even most complicated data collections
\begin{align}
  \label{eq:marginal}
  \E_{{\bm x}\sim p(\bm x)}\left[\log p_{\bm \theta}(\bm x)\right]=\E_{{\bm x}\sim p(\bm x)}\left[\int_{\mathcal{Z}}\log p_{\bm \theta}(\bm x, \bm z)\diff {\bm z}\right].
\end{align}
Maximizing this marginal log-likelihood is computationally intractable though
which is why, using a parametric posterior distribution, the so-called
variational lower bound (VLB) is optimized instead to approximate the model
evidence above
\begin{align}
  \label{eq:marginal}
  \argmax_{\bm \phi,\bm\theta}\E_{\substack{{\bm x}\sim p(\bm x)\\{\bm z}\sim q_{\bm \phi}(\bm z\mid \bm x)}
  }\left[\log \frac{p_{\bm \theta}(\bm x, \bm z)}{q_{\bm \phi}(\bm z\mid \bm x)}\right]\leqslant\E_{{\bm x}\sim p(\bm x)}\left[\log p_{\bm \theta}(\bm x)\right].
\end{align}
Traditionally, Monte Carlo sampling and averaging replaces the calculation of
the expectations. Within the context of deep learning, this bound is usually
decomposed into a reconstruction error and a regularization term forming the
overall objective whose detailed derivation can be found in
\cite{kingma2014semi}
\begin{align}
  \begin{split}
    \label{eq:elbo}
    \lagr_{\bm \phi,\bm \theta}(\bm x) = \E_{{\bm z}\sim q_{\bm \phi}(\bm z\mid \bm
      x)}&\left[\log p_{\bm \theta}(\bm x\mid \bm z)\right] \\
    - \KL&\left[q_{\bm \phi}(\bm z\mid \bm x)\;\|\;p(\bm z)\right]\leqslant
    \log p_{\bm \theta}(\bm x).
  \end{split}
\end{align}
Above densities are parameterized by individual neural networks each with its
own set of variational parameters ${\bm \phi}$ and ${\bm \theta}$ for joint optimization
of this surrogate objective. Maximizing per sample-estimates of the VLB with
stochastic gradient-based procedures then facilitates effective maximization of
the evidence in end-to-end training. In the following section, the task-specific
choices of the conditionals are established and assumptions on the distributions
are justified which pave the way for an effective signal processing and the
required quantization of the continuous measurement domains later on.

\subsubsection{Theory About Categorical Variational Autoencoders}
\label{sec:cvae}
The proposed compression approach necessitates the discretization of the latent
space, preferably in a stochastic manner while forcing the latents at the same
time to take on only a predefined range of values. Categorical variational
autoencoders \cite{rolfe2016discrete} are a special case of variational
inference models described in the former section, most often used when a
discrete probabilistic selection of features is desired, as in the present
case. On an abstract level this architecture consists of an encoder and decoder part
with a discrete stochastic bottleneck in between. The encoder network comprises
a series of spatial downsampling convolutions with learnable filters defined by
weights and biases $\bm \phi$. Transforming the input $\bm x \in \mathcal{X}$ into a
discrete stochastic latent representation $\bm c \in \mathcal{C}$ of lower
dimension forces it to learn an efficient compression by uncovering hidden
concepts within the data. This Bayesian network is also called inference model
\cite{kingma2019introduction} and is used to approximate the variational
posterior in the VLB. In this context, it infers probability masses
$\bm \pi\in\R^{N\times K}$ as nonlinear functions of the data where
$N=h \times w$ is the downsampled spatial extent of the input
\begin{align}
  \bm \pi = \text{encoder}_{\bm \phi}(\bm x) \qquad \bm x \stackrel{\text{iid}}{\sim} p(\bm x).
\end{align}
The encoder thus effectively parameterizes factorized $K$-categorical
distributions over the $N$ discrete latents collectively denoted as $\bm c$
henceforth
\begin{align}
  \label{eq:enc_dist}
  q_{\bm \phi}(\bm c\mid \bm x) = \prod^N_{i=1}q_{\bm \phi}(c_i\mid \bm x)=
  \prod^N_{i=1}\text{Cat}(c_i;\bm \pi_i(\bm x)).
\end{align}
Relying on a technique called \textit{amortized inference}, all categoricals
efficiently share the same set of variational parameters $\bm \phi$, yet, varying
input induces different posterior conditionals. Every latent variable follows
exactly one categorical which restricts the possible values, it can take on to a
finite, potentially large number of discrete latent codes $K
=|\mathcal{C}|$. This aligns neatly with the proposed intention of
deconstructing continuous data into related discrete representations located in
a finite space $\mathcal{C}$. In fact, various applications exist in which
categorical latents are more suitable than their continuous counterpart, one of
them being the subdivision of images into concrete constituents that this work
aims for. Following \cite{thickstun2020discrete}, the decision for a restricted
number of possible outcomes of the sampling process allows to quantify the
compression of the network from an information-theoretical point of view:
According to Shannon's source coding theorem \cite{shannon1948mathematical} this
choice defines an explicit upper bound $\log K$ on the number of bits of
information $\mathcal C$ can represent, an aspect which will be picked up again
in the next paragraph. Discrete samples from the latent distributions in
equation \eqref{eq:enc_dist}, representing square patches of the input image,
are passed through the decoder, a sequence of learnable convolutional upsampling
blocks governed by weights and biases $\bm \theta$ in an attempt to restore the
original data as accurately as possible. This so-called generator approximates
the likelihood in the VLB by mapping i.i.d. latents to corresponding mean
vectors $\bm \mu\in\R^{M}$ with $M=H\times W\times C$ of the input so that
\begin{align}
  \bm \mu = \text{decoder}_{\bm \theta}(\bm
  c) \qquad \bm c \stackrel{\text{iid}}{\sim} q_{\bm \phi}(\bm c\mid\bm x).
\end{align}
The investigation of the data distributions detailed in section
\ref{sec:data} allows for the simplifying assumption of pixel-wise
independence with channel-wise fixed variance. This Bayesian model then induces
spherical Gaussians in image space
\begin{align}
  \begin{split}
    \log p_{\bm \theta}(\bm x \mid \bm c) &= \log \prod^M_{j=1}p_{\bm \theta}(x_j\mid \bm c)\\
    &= \sum^M_{j=1}\log\N(x_j; \mu_j(\bm c),\sigma^2)
    \label{eq:gendist}
  \end{split}
\end{align}
quantifying the information lost through transmission. Aside from concrete
density choices and fuzzy latents, VAEs employ yet another regularization by
introducing prior knowledge into their latent space, influencing the shape of
the categorical posteriors and forcing samples to attain certain
properties. Illustratively, these regularity constraints cause similar data
points to end up close together in feature space, whereas the distance for
dissimilar ones is increased. To this end, the divergence in equation
\eqref{eq:elbo} rewards proximity of the posterior
$q_{\bm \phi}(\bm c\mid\bm x)$ to a task-specific existing belief $p(\bm
c)$. Likewise, posterior distributions deviating too much from the specified
priors, cause substantial perturbances of the overall objective. Seeking for an
optimal exploitation of the entire code space $\mathcal{C}$, fixed priors in the
form of diffuse uniforms are chosen here in agreement with the relevant
literature \cite{lavda2019improving}. Following the notation in
\cite{bishop2006pattern}, these represent the maximum entropy configuration for
the discrete case
\begin{align}
  \label{eq:cat}
  p(\bm c)= \prod^N_{i=1}\text{Cat}(c_i; \bm \pi_{i}) =
  \prod^N_{i=1}\prod^K_{k=1}\pi^{c_k}_{ik},~~~ \pi^{c_k}_{ik} = 1/K
\end{align}
and recommend equal chance for every category to be selected. This imposes
rather strict requirements on the latent space, but also helps in acquiring
diverse and semantically meaningful features. These should eventually be able to
efficiently decompose a large and versatile range of input into a set of
well-defined constituents. On a more technical note, an appropriate prior choice
also mitigates the risk of posterior collapse, a phenomenon in which the model
contents with relying on only a few fixed latents for data restoration. In this
pathological case, described in \cite{goyal2017z} the categorical VAE would
mimic a standard autoencoder with little stochastic characteristics. Although
this might not affect its reconstruction abilities noticeably, it compromises
the models expressiveness and generative potential profoundly. In view of the
specific distributions defined above and with joint uniform priors in place, the
VLB can be revised, offering further insights from an information-theoretical
perspective. In particular, the divergence in expression \eqref{eq:elbo} now
disintegrates into Shannon's entropy and an upper bound on the encoders
compression capabilities as explained before
\begin{align}
  \begin{split}
    \label{eq:kl}
    \KL\left[q_{\bm \phi}(\bm c\mid \bm x)\;\|\;p(\bm c)\right] &= \E_{q_{\bm \phi}(\bm c \mid
      \bm x)}\left[\log q_{\bm \phi}(\bm c \mid \bm x) - \log K^{-1}\right] \\
    &=-\ent\left[q_{\bm \phi}(\bm c\mid\bm x)\right] + \log K.
  \end{split}
\end{align}
Both terms yield an estimate on the expected amount of information the model
transmits about the data via its latents. Optimization of the variational bound
thus involves maximizing both the likelihood and the conditional entropy of the
categorical posterior bounded from above by a hyperparameter constant to be
chosen a priori
\begin{align}
  \begin{split}
    \label{eq:beta}
    \lagr_{\bm \phi,\bm \theta}(\bm x)= \E_{\bm c\sim q_{\bm \phi}(\bm c\mid \bm x)}&\left[\log
      p_{\bm \theta}(\bm x\mid \bm c)\right] \\
    + \beta &\left(\ent \left[q_{\bm \phi}(\bm c\mid \bm x)\right] - \log K\right).
  \end{split}
\end{align}
An alternative interpretation is given in \cite{doersch2016tutorial} as the
divergence term being the expected amount of information necessary to convert an
uninformed sample of $p(\bm c)$ into one from the approximated posterior
$q_{\bm \phi}(\bm c\mid\bm x)$. For a posterior in equilibrium, i.e. exhibiting
maximum entropy the divergence vanishes completely but the average number of
bits required to communicate the state of its latent space reaches a maximum.
Pertaining to the categorical model, a gain in entropy by striving for a uniform
latent space utilization obstructs information flow through the network,
essential for the overall reconstruction goal \cite{bishop2006pattern}. This
contradiction of a more disentangled latent representation on one side and an
optimal data restoration on the other is controlled by the introduction of an
application-specific weighting parameter $\beta>0$ in expression
\eqref{eq:beta}. Originally proposed in \cite{higgins2017beta} this
hyperparameter allows to subtly balance the relative tradeoff between both
opposing terms in the VLB. Strictly speaking, the bound only holds for $\beta=1$
recovering the original formulation. Yet, in practice, differing values have
proven helpful to consolidate numerical convergence as demonstrated in later
sections. In summary, the model now constitutes a mixture-of-Gaussians
approximation to the marginal likelihood in equation \eqref{eq:marginal}. It
features a categorical encoder mapping continuous input to parameters of
discrete distributions that are drawn from once per latent variable during each
forward pass of a single data sample. The decoder thus receives varying input in
every iteration even for identical data points before transforming the latents
into parameterized normal distributions over the continuous image space,
rendering the entire training process probabilistic.

\subsubsection{Declining Relaxation of Categorical Feature Selections}
\label{sec:gumbel}
One of the strengths of modern neural networks is the effective approximation of
high-dimensional functions by adapting millions or even billions of parameters
through repetitive iterations over large datasets. After running a randomly taken
subset of data through the directed graph, deviations from the respective
objective are propagated back in reverse direction to adjust the networks
weights via gradient updates for a lower error value in the next
pass. Stochastic gradient descent algorithms are usually employed to fit the
networks complexity to the respective data via batch-wise estimates of the
negative likelihood. With regard to the model presented in the previous section,
data-based optimization of the VLB should follow this same efficient
procedure. However, changes in entropy require passing gradients of sampled
entities back to the inference network. This poses a problem since it is not
possible to back-propagate through stochastic nodes in computational graphs let
alone through points at which discrete sampling took place. Various solutions to
this issue have been proposed over the years, described at length in the
literature \cite{bengio2013estimating}, \cite{williams1992simple} alongside
their respective characteristics. The method employed in this work relies on a
smooth relaxation of the categoricals during the course of the data-fitting
process by replacing the discrete samples with continuous
approximations from so-called concrete distributions. Instead of drawing from
true categoricals parameterized by the approximate posterior
\begin{align}
  \bm c \sim  \text{Cat}(\bm c; \bm \pi_{\bm \phi}(\bm x)) = q_{\bm \phi}(\bm c\mid \bm x),\qquad
  \bm c \in \mathbb{N}^N
  \label{eq:categorical}
\end{align}
a vector-valued proxy sample is taken from a Gumbel-Softmax distribution for
every of the $N$ latent variables
\begin{align}
  \label{eq:gumbelsampling}
  \tilde{\bm c}_i = \text{GS}(\tilde{\bm c_i};\bm \pi_{\bm \phi_i}(\bm x),\bm \gamma_i, \tau) \qquad \bm \gamma_i \stackrel{\text{iid}}{\sim} \text{Gumbel}(0,1)
\end{align}
with every component of the vector $\tilde{\bm c}_i \in\R^K$ calculated by
\begin{align}
  \label{eq:decline}
  \tilde{c}_{il}=\frac{\exp\left(\left(\log
  \pi_{\bm \phi_{il}}+\gamma_{il}\right)/\tau\right)}{\sum_{k}\exp\left(\left(\log \pi_{\bm \phi_{ik}}+\gamma_{ik}\right)/\tau\right)},\quad
  l,k = 1,\dots,K.
\end{align}
Lower indices $l$ and $k$ denote categories now and $\tau>0$ designates a
temperature hyperparameter whereas $\log \pi_{\bm \phi_{ik}}$ are the encoder logits
which are interpreted as unnormalized log-probability of each category. This
transformation provides well-defined gradients of the concrete distribution with
respect to the parameters of the encoders final layer. In fact, it can be
considered a variant of the reparameterization trick proposed in
\cite{kingma2014auto} which turns sampling of the latents into a deterministic
function of the encoders logits and some independent additive noise from a
predetermined distribution. For the discrete case, these perturbations
$\bm \gamma_i$ in equation \eqref{eq:gumbelsampling} follow a standard Gumbel
distribution \cite{gumbel1958statistics} allowing their efficient calculation
via a standard uniform
\begin{align}
  \bm \gamma_i = -\log(-\log(\bm u_i)) \qquad \bm u_i \stackrel{\text{iid}}{\sim} \text{Unif}(0,1)
\end{align}
\begin{figure}[h!!]
  \centering
  \includegraphics[width=0.4\textwidth]{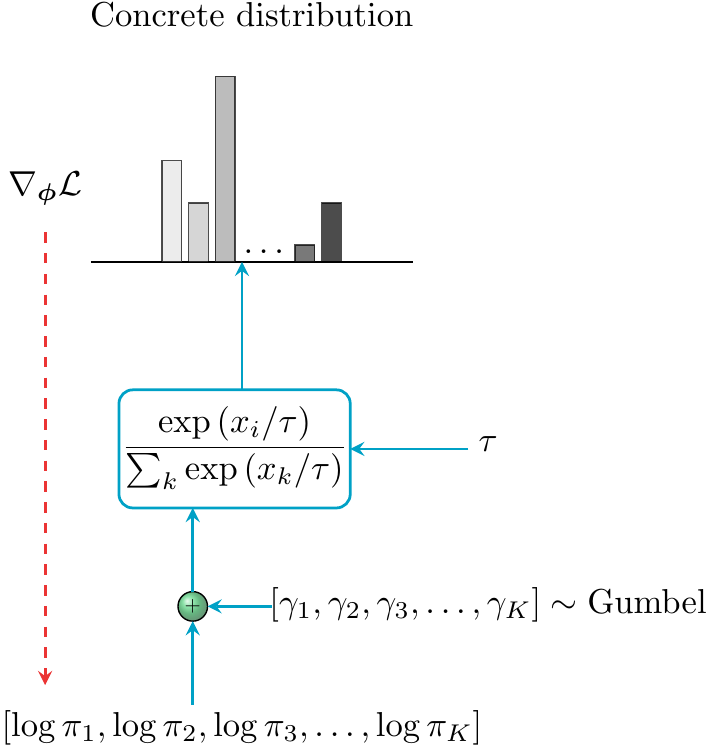}
  \caption{Gradient flow through the sampling process after
    Gumbel-Softmax reparameterization. The introduction of a temperature
    parameter and the addition of perturbation noise leads to pseudo-categorical
    latent variables with well-defined gradients. Errors can therefore be
    back-propagated to the encoder model, allowing efficient end-to-end training
    of the corresponding neural network.}
  \label{fig:gumbel_path}
\end{figure}
inserting stochastic properties into the learning process without having to
differentiate through the sampling itself. Gradients only flow along the
re-parameterized deterministic nodes of the feature selection, see Figure
\ref{fig:gumbel_path}, so that the VLB can be optimized w.r.t the inference
networks parameters by
\begin{align}
  \nabla_{\bm \phi}\lagr_{\bm \phi,\bm \theta}(\bm x)  \simeq \nabla_{\bm \phi}\log p_{\bm \theta}(\bm x\mid\tilde{\bm c})-\nabla_{\bm \phi} \log q_{\bm \phi}(\tilde{\bm c}\mid \bm x).
\end{align}
This kind of expressions are referred to as \textit{Stochastic Gradient
  Variational Bayes} (SGVB) estimators in the literature
\cite{kingma2019introduction} and in this case produce biased gradients for the
parameter adaption for reasons explained in the following. Gumbel-Softmax
distributions were introduced simultaneously in \cite{jang2016categorical} and
\cite{maddison2016concrete} as an extension to the Gumbel-Max trick proposed in
\cite{maddison2014astarsamp} to consistently sample from discrete distributions
during end-to-end training of neural networks via backpropagation. The
differentiable softmax function replaces the argmax operation used in
\cite{maddison2014astarsamp} while maintaining the parameterizations relative
ordering. Not only does its application normalize the raw logits, it also
provides a soft approximation to the true categorical distribution without ever
yielding one-hot samples exactly. Consequently, for temperatures
$\tau \in \R^+$, the expectation of the concrete distribution never exactly matches
that of the categorical actually sought after. The deviation between both and
thereby the approximation quality is controlled by adjusting the temperature
accordingly. Technically, this allows balancing the bias-variance tradeoff
between the gradient estimates of taken samples. Larger temperatures emphasize
the samples' bias while reducing their variance due to higher entropies, causing
the distributions to become increasingly uniform. And although this complies
with the prior choice and has been reported to improve training robustness, it
also hinders the encoder from moving towards outputting the desired categorical
distributions. For lower temperature values close to zero, the one-hot encoding
nature of the distributions is more accentuated as they converge towards true
categoricals. The gradient estimates become more unbiased, but their variance
increases, which generally disturbs the updating steps of the associated
parameters and thereby the overall learning process. Choosing an appropriate
$\tau$ is thus a complex task and the authors in \cite{jang2016categorical}
recommend to either learn its value or anneal a prefixed one according to a
predefined schedule which is what has been done in the current work as described
in section \ref{sec:training}. The gradients of equation \eqref{eq:beta}
w.r.t. the generators variational parameters are calculated by Monte Carlo
estimates of the batch-wise conditional likelihood with latent samples taken
from the pseudo-categorical approximate posterior in the center of the graph
during each forward pass.
\begin{align}
  \nabla_{\bm \theta}\lagr_{\bm \phi,\bm \theta}(\bm x) \simeq \nabla_{\bm \theta} \log p_{\bm \theta}(\bm x\mid \tilde{\bm c})\qquad \tilde{\bm c} \sim q_{\bm \phi}(\tilde{\bm c}\mid \bm x).
\end{align}

\begin{figure*}[h!]
  \centering
  \includegraphics[width=\textwidth]{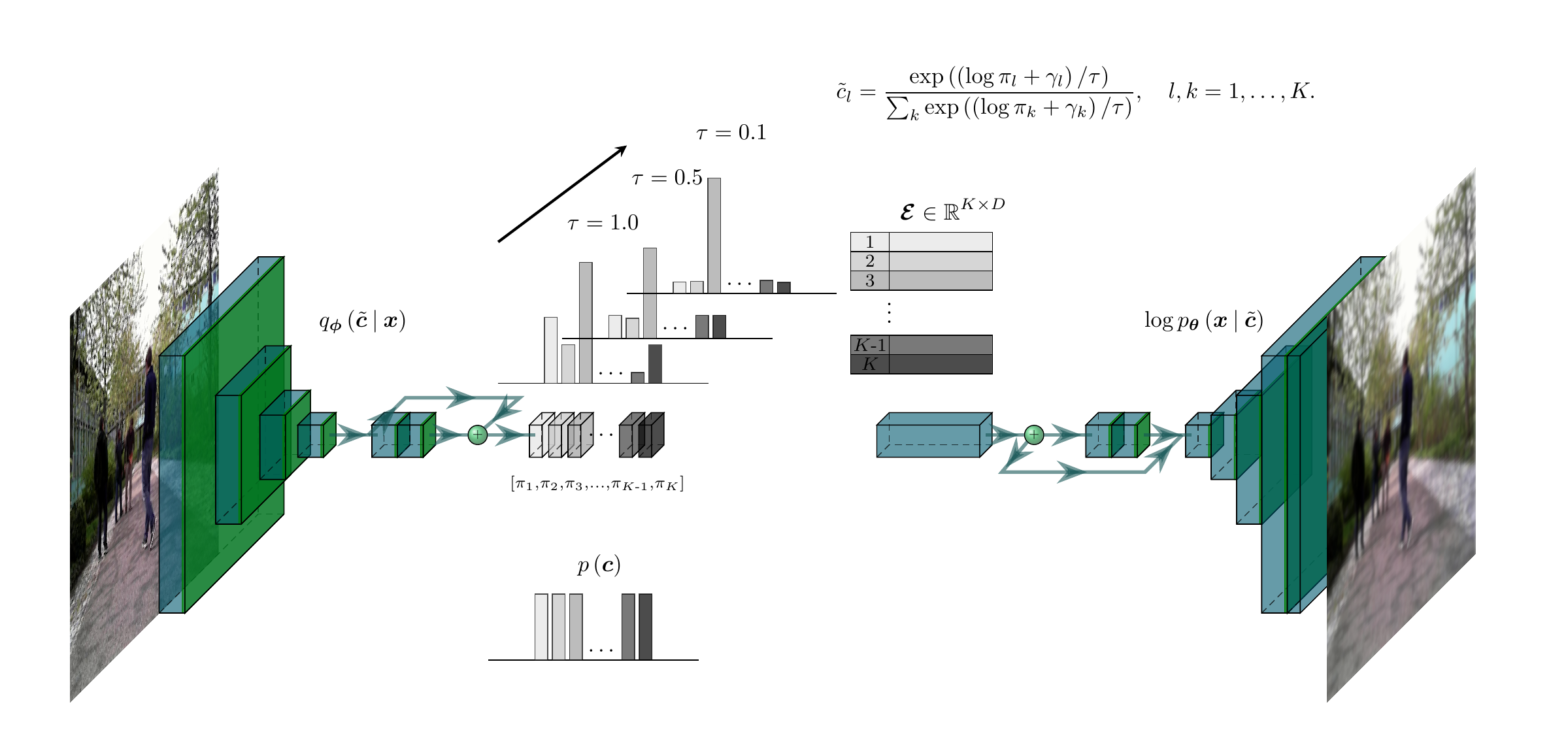}
  \caption{Categorical variational autoencoder architecture used to compress
    each modality input separately into discrete integer sequences. The
    stochastic bottleneck samples from discrete distributions by means of a
    continuous relaxation scheme to determine which of the learnable semantic
    vocabulary entries represents best which patch of the input image. A uniform
    prior rewards equal vocabulary use and a decline in temperature causes the
    sampling procedure to become increasingly confident. Finally, the selected
    feature vectors are translated back into a preferably accurate replica of
    the respective input. For clarity, this process is illustrated for only one
    latent variable. Illustration based on the TikZ script \cite{iqbal2018plot}
    for plotting artificial neural network architectures.}
  \label{fig:dvae}
\end{figure*}

\subsubsection{Implementation Details of the Categorical Autoencoders}
\label{sec:architecture}
Even though a wide range of autoencoder architectures exist both in theory and
code and despite the fact that weights of numerous well-known networks are
readily available for download and deployment in frameworks like
\cite{paszke2019pytorch}, the specific data used in this project necessitate
custom training. Most backbones are typically pre-trained on purified and
cleansed benchmarks rather than realistic, application-oriented datasets which
renders them unsuitable for the current project as their output lacks
reconstruction quality or shows severe distortions if supplied with input
described in section \ref{sec:data}. Moreover, the categorical bottleneck
\ref{sec:gumbel} and design choices tailored to the presented use case and given
in the remainder of this section require training vanilla categorical VAEs for
each modality from scratch. Figure \ref{fig:dvae} illustrates the concrete
autoencoder architecture employed in the first stage of this work. Both
inference and generator networks are fully-convolutional \cite{lecun1999object}
seeking to exploit the immanent advantages of those models discussed in the
beginning. The nonlinearities are ReLU activation functions
\cite{maas2013rectifier} following every of the four down- and upsampling
operations, respectively. In view of the data-specific distributions the decoder
is to model, biases of all convolutional layers were preemptively disabled so as
to prevent the model from solely learning a constant mapping from input to
respective means as is warned against in \cite{ruff2018deep}. The stochastic
network maintains a constant number of \num{256} kernels per filter throughout,
with a kernel size of four and deliberately foregoes pooling layers in the
encoder in favor of double-strided convolutions with same padding. Pooling
operations are notorious for preserving predominantly low frequency information
while neglecting fine-graded details \cite{rippel2015spectral}. Retaining
high-frequency content was argued to be essential though for reasons outlined on
numerous occasions in this paper so that there exists genuine interest to not
dispose of them by careless design decisions from the outset. Batch
normalization \cite{ioffe2015batch}, although still ubiquitous in modern deep
learning, was dispensed with too, as was any other normalization method
ordinarily included in most architectures. Though often shown to improve
training stability and accelerating convergence rates during training, batch
normalization is prone to introduce peculiar dependencies of results on chosen
batch sizes and data distributions. Every so often, this leads to unfavorable
mismatches between training and validation runs due to inconsistent feature
scaling in both phases. As a positive side effect, doing away with norm layers
limits the total number of parameters to be optimized. Instead, normalization of
the data is performed prior to network fitting by rescaling and recentering the
data points according to their underlying modalities as described in sections
\ref{sec:radpostprocessing} and \ref{sec:campostprocessing}
respectively. Additional pre-activation residual blocks with pure skip
connections \cite{he2016deep} are inserted immediately before and after the
variational bottleneck to efficiently further information transmission to the
posterior parameterization. These are succeeded by scaling $1\times1$ convolutions to
bring the latent features in agreement with the desired number of categories
$K$. The final logits $\bm \pi$ of the inference network parameterize
$N=h\times w =16 \times 16=256$ categorical distributions per data sample over a single
common learnable dictionary $\bm \emb\in\R^{K\times D}$ prepended to the
generator. Latent samples then look up specific entries of this data-adaptable
embedding, comparing them for semantic similarity during the learning
process. Subsequent convolutions with a receptive field of $1\times1$ spatially
aligns the vocabularies feature dimension $D$, with the decoder blocks in a
computationally efficient manner. Since recent investigation suggests that
generators too powerful have the potential of approaching the data distribution
by ignoring the majority of latents and the prior conditioning
\cite{rubenstein2019variational}, the decoder mirrors the encoder architecture
as closely as possible and with modest complexity. In terms of reconstruction
quality, no substantial difference was found between using either transposed
convolutions \cite{dumoulin2016guide} or upsampling followed by regular
convolutions. The infamous checkerboard artifacts \cite {odena2016deconvolution}
frequently observed when using transposed convolutions were of little concern
here due to quadratic kernel sizes of four being an integer multiple of the
stride. A final convolutional layer transforms the last block of feature maps
into \num{3}- and \num{1}-channel output of original spatial extent, modeling
the pixel means for camera and radar input, respectively. The Gaussian
assumption on those pixel distributions of both domains with unitary covariance
as defined by equation \eqref{eq:gendist}, makes the reconstruction term in the
VLB proportional to the expected mean squared error under the decoder model
$\log_{\bm \theta} p(\bm x\mid\tilde{\bm c}) \propto\frac{1}{2\sigma^2}\Vert \bm x - \bm \mu_{\bm
  \theta}(\tilde{\bm c})\Vert^2$ \cite{goodfellow2016deep}. A different choice for the
distribution of the observation model in form of yet another categorical over
the discrete pixel space was suggested in \cite{snderby2017Continuous}. This was
decided against here for two reasons: First, calculation of the corresponding
cross-entropy loss requires an output channel number equal to the number of
possible pixel values and thus significantly increases both model size and
parameter space. Second, while the range of quantized integers for the unscaled
camera data is known, radar spectra can assume largely varying values in a
highly dynamic range depending on various factors, as stated in
\ref{sec:radpostprocessing}. Even if the value range could be specified a
priori, subsequently casting continuous frequency points into integers would
discard valuable information and hence stands in stark contrast to the arguments
presented in chapter \ref{sec:data}.

\subsubsection{Acquisition of Modal-Specific Probabilistic Dictionaries}
\label{sec:training}
The primary reason for compressing both modalities into integer sequences is to
reduce the memory footprint and compute requirements of the transformer model
detailed in section \ref{sec:trafo}. For this purpose, the original images
$\bm x_{\text{cam}} \in \R^{H\times W\times3}$ and
$\bm x_{\text{rad}} \in \R^{H\times W}$ are downsampled by a factor of 16 while passing
through the encoder part of the network, yielding compression ratios of
\num{768} and \num{256} for the camera and radar domain respectively upon
discretization. This might seem to contradict the initially stated goal of
preserving a maximum of information but rather substantiates the argument of
giving the network the freedom to decide which information to keep and which to
discard by adjusting its kernel weights. As such, this drastic compression
approach is a tradeoff between information preservation and tenable computation
effort with regard to the attention mechanism to follow. At the same time, this
underlines the relevance of having to acquire an expressive dictionary, able to
translate quantized information into semantically meaningful learned vectors to
be eventually decoded by the generator. Given the general design of the overall
network structure, outlined in the former chapter, this procedure leaves two
main adjustment options, namely the size of the vocabulary $K$ and the feature
length $D$ of each associated independent embedding vector $\bm e \in\R^D$. One of
the experiments, detailed in the following, ablates the number of selectable
image constituents $K$ as this integral quantity should have major influence on
information conservation. Concretely, the dataset's variability, apparent also
in the spatially confined $16\times16$ partially overlapping patches of the input, is
compressed into a single integer per such section. It is believed that too small
of a number $K$ should give too little choice to represent image content,
forcing the dictionary to forfeit discriminative power in its feature
dimension. A larger number of categories, on the contrary, gives it more freedom
to develop patch-specific feature characteristics and is therefore likely to
offer sufficient flexibility for the retention of fine details. Concerning the
length of the feature vectors the vocabulary encompasses, $D=512$ is picked
after carefully balancing out requirements in feature expressiveness and memory
demands. The biased gradient estimator, described in section \ref{sec:gumbel},
is used, enabling error propagation through the probabilistic nodes of the
computation graph at train time. During this phase, for every input data point,
\num{256} concrete sample vectors $\tilde{\bm c}_i \in\R^{K}$ are linearly but
separately combined with the dictionary $\bm \emb\in\R^{K\times D}$ along the
categorical dimension $K$. Owing to the continuous relaxation strategy, this
contraction is weighted according to the current output of the inference model
and further influenced by noise and the current temperature. This in turn
encourages the involvement of all latent categories, especially in early
training, facilitates signal flow through the network and provides valuable
feedback to the encoder. An initial temperature around $\tau=1$ in combination with
randomly initialized encoder parameters stimulates a more uniform exploitation
of the dictionary. In this context, the posteriors probability mass functions
(PMF), given by the encoder logits are an indicator for the attraction from each
image patch to every vocabulary entry. By measuring proximity in feature space
between logits and learnable dictionary vectors, the inference model serves as a
classifier of sorts, distributing scores across the entire embedding. The
softmax normalizes the logits promoting numerical robustness while avoiding
having image constituents settling too early for certain vocabulary
entries. Sampling Gumbel noise at the bottleneck as given by equation
\eqref{eq:gumbelsampling} has the added benefit of reducing the risk becoming
stuck in local minima during the data-fitting process. Attempting to close the
variational gap between VLB and true likelihood, requires $\tau\rightarrow 0 $ monotonically
so that every image patch in the dataset should eventually be assigned one
unique category. Decreasing the temperature causes the encoder to become
increasingly self-confident about the category each image section belongs to by
shifting probability masses accordingly. Consequently, the temperature is slowly
but steadily annealed from $\tau = 1$ in every training step $t$ according to
\begin{align}
  \tau =\max\left(0.0625, \exp(-0.000015\,t)\right)
\end{align}
for the first \num{10} and \num{50} epochs for radar and camera input
respectively. This schedule decreases the parameter exponentially until reaching
a minimum value of $\tau=0.0625$ gently converging concrete samples towards true
categoricals at the expense of a larger variance in gradient estimation. Having
reached the lowest temperature, it is assumed that the model has adjusted its
weight sufficiently to not be thrown off guard by coarse gradient values,
turning unstable in subsequent iterations. During the validation phase, taking
place after every training epoch, the temperature is frozen at the current
value. To assess the models reconstruction quality, both qualitatively and
quantitatively, it is then passed test samples from the validation dataset. The
encoders response $\tilde{\bm c} \in \R^{h\times w\times K}$ in the form of concrete samples
with nonzero values for all elements is then contracted along the dictionary
entries as is done within training steps. Additionally, as no differentiation is
performed at test time, the mode of the encoder-parameterized distribution for
every of the $N$ latent variables is selected
\begin{align}
  c_i = \argmax_{k\in [1,K]} \pi_{\bm \phi_{ik}}(\bm x),\quad i = 1,\dots,N=h\times w
  \label{eq:hardsampled}
\end{align}
indicating where the bulk of the probability mass is currently located. To
include genuine discrete sampling in the validation probing, the encoders output
is also used to define $N$ true categoricals as a third alternative. Sampling
these adds slight regularization and some degree of fuzziness to the selection
of the probabilistic latents. Moreover, excluding the temperature influence
allows to examine the impact of an increasingly confident encoder on the
discretization and restoration capabilities of the model. The sample spaces for
the last two index collections consist of finite integer sequences
$\bm s\in \mathbb{N}^N$ with $N=256$ representing the compressed input image of
the respective modality in raster order. Extracting these series of tokens lays
the foundation for the sequence modeling in the second stage. Using them here
for the discrete selection of reconstruction features gives an outlook on the
achievable image quality and serves as a visual upper bound to the
autoregressive generation performed later on. Except for the concrete case, only
individual embeddings $\bm e\in \R^D$ are selected now by using the sampled
integers to index into the corresponding domain-specific dictionary
$\bm \emb\in\R^{K\times D}$. This allows to retrieve associated feature vectors which
are hoped to have acquired some notion of context-awareness during the course of
previous training. In doing so, the entire token sequence
$\bm s\in \mathbb{N}^N$ is transformed into a continuous latent equivalent
$\bm z\in \R^{N\times D}$. After reshaping into
$\bm z\in \R^{h\times w\times D}$ the decoder reconstruction begins to reduce the feature
dimension while concurrently upsampling spatial dimensions. In theory, at
convergence of the data-fitting process, only subtle differences between the
three different token sets and their visual restorations should remain: Upon
training termination, the encoder should have gained the competence to
unambiguously classify subimages into one of many categories, while the low
temperature effectively turns sampling from the Gumbel-Softmax distribution into
nearly discrete operations with only minor deviations due to the influence of
Gumbel noise. Optimizing the models objective in equation \eqref{eq:beta}
involves the opposing goals of increasing the entropy with respect to $\bm \phi$
while raising the lower bound in terms of $\bm \theta$ to maximize the data
likelihood. To mitigate this tradeoff fixed values
$\tilde{\beta}_{\text{cam}}=\num{5e-4}$ and
$\tilde{\beta}_{\text{rad}}=\num{5e-5}$ were found for the data
decompositions. Their magnitudes reflect to a certain degree the estimated
per-sample difference in content relevance between both modalities and are
combined with the inverse compression ratios $\beta = \tilde{\beta}N/M$ with $N$ as the
size of the latent space and $M$ as the number of input/output dimensions as
before. It was found that these values allow to uphold the delicate balance
between the contrasting objectives and provide for sufficiently structured
latent spaces at the cost of slightly distorted reconstructions. Resorting to a
dynamic schedule for this parameter, reported successful in
\cite{bowman2015generating}, did not show any improvements for the present
analysis. In fact, gradually increasing the divergence weight corrupted
convergence from the very beginning, presumably due to conflicts with the
temperature annealing. Minimizing the equivalent negative lower bound was
performed by Adam \cite{kingma2015adam} as the numerical solver of choice due to
its reputation of being rather forgiving to suboptimal hyperparameter choices
\cite{karpathy2019recipe}. This also includes rough initial guesses on the
learning rate, which was set to \num{3e-4} but reduced by half after every
\num{10} validation epochs in which no overall loss reduction could be
observed. Kaiming initialization of both sets of variational parameters
$\bm \phi$ and $\bm \theta$ in the convolutional layers was done following
\cite{he2015delving} while the weights of the vocabulary were drawn from a
uniform distribution. Separate models were trained for both modalities and all
computations were run on single GeForce RTX 2080 TI units with
\SI{\sim12}{\giga\byte} of RAM, using a maximum batch size of
\num{24}. Calculations were terminated after the validation loss stopped
increasing for more than 20 epochs. The outlined probabilistic compression is a
complex task as is, so no further data augmentation was used on either
domain. Range-Doppler maps carry inherent meaning of various physical quantities
which cropping or random resizing operations would destroy or at least severely
impair. Specific intensity distributions immediately establish useful
correspondences to the time-associated content of the camera frame and must not
be tinkered with. For similar reasons, the frequency plots were not subjected to
flipping of any kind, as this would reverse the velocity or range estimates
included in the RF data. Likewise, stochastically mapping the camera footage to
an accurate replica is difficult due to the diverse content and VAEs infamous
tendency of imposing certain amounts of blurriness onto their output
\cite{kingma2019introduction}. So no further processing was applied to the RGB
data other than that described in section \ref{sec:data}.

\subsubsection{Results and Discussion of Probabilistic Decompositions}
\label{sec:vae_res}
Evaluation of the implemented models is complicated because of the custom
datasets and the novel intention-driven stochastic decomposition method of
continuous domains. Traditional benchmarks do not apply to the specific case of
radar frequency data, which is why ablation studies among different model
configurations are performed to obtain an impression of their abilities. Also,
no universally agreed-upon performance metrics for self-supervised learning in
general exist as of today. This is particularly true for generative models and
the quality of their synthesizing capabilities, whose evaluation is still an
open research question. Over the years, many measures have been proposed without
one of them ever exclusively taking prevalence over the others. Yet, certain
approaches have proven helpful to examine the abilities of trained networks, at
least to some extent. Here two methods are used, relying on similarities between
original data and reconstructions in pixel and feature space, respectively. The
first one assesses the quality with which models are mapping discrete stochastic
latents back to the continuous domain by application of two metrics, the
Frobenius norm and the peak signal-to-noise ratio between input and means output
by the generators. Table \ref{tab:frob_cam} to table \ref{tab:psnr_rad} show the
results for both modalities, different vocabulary sizes and sampling
methods. Additionally, the number of feature maps constant along the network
architecture is ablated as this gives clear evidence on the necessity of
sufficient complexity within the models for restoration quality. Since
the Frobenius norm is only defined for matrices, it is calculated and averaged
channel-wise for the camera images. The results are then averaged over the
entire validation dataset and across three consecutive runs to reduce sampling
noise. Also, the tables report the epoch number after which the iterations were
deemed to having converged, i.e. after which no improvement in validation error
occurred for more than ten epochs.
\begin{table}[h!]
  \caption{Frobenius norm of probabilistic camera reconstructions for different
    sampling methods and varying dictionary sizes, averaged over three consecutive runs on
    the validation set}
  \renewrobustcmd{\bfseries}{\fontseries{b}\selectfont}
  \sisetup{detect-weight,mode=text,group-minimum-digits = 4}
  \centering
  \begin{tabular}{l
    S[table-format = 2.2]
    S[table-format = 2.2(1), separate-uncertainty]
    S[table-format = 2.2(1), separate-uncertainty]
    |c
    |c}
    \toprule
    $K$ & {Mode} & {Categorical}  & {Concrete}  & Depth & Epoch\\
    \midrule
    \multirow{3}{*}{64}
        &19.40 & 19.41(11) & 19.33(2) & 64 & 122\\
        &16.54 & 16.59(13) & 16.45(5) & 128 & 141\\
        &13.59 & 13.64(44) &13.57(8) & 256 & 68\\
    \midrule
    \multirow{3}{*}{256}
        &18.88 & 18.87(4) &18.84(2) & 64 & 156\\
        &15.77 & 15.77(2) &15.75(4) & 128 & 146\\
        &13.43 & 13.44(18) &13.37(15) & 256 & 59\\
    \midrule
    \multirow{3}{*}{1024}
        &17.84 & 17.87(22) &17.72(6) & 64 & 156\\
        &15.46 & 15.51(20) &15.42(5) & 128 & 142\\
        &\bfseries 13.16 & \bfseries 13.23(14) &\bfseries 13.17(2) & 256 & 75\\
    \bottomrule
  \end{tabular}
  \label{tab:frob_cam}
\end{table}
\begin{table}[h!]
  \caption{Frobenius norm of probabilistic radar reconstructions for different
    sampling methods and varying dictionary sizes, averaged over three
    consecutive runs on the validation set.}
  \renewrobustcmd{\bfseries}{\fontseries{b}\selectfont}
  \sisetup{detect-weight,mode=text,group-minimum-digits = 4}
  \centering
  \begin{tabular}{l
    S[table-format = 2.2]
    S[table-format = 2.2(1), separate-uncertainty]
    S[table-format = 2.2(1), separate-uncertainty]
    |c
    |c}
    \toprule
    $K$ & {Mode} & {Categorical}  & {Concrete}  & Depth & Epoch\\
    \midrule
    \multirow{3}{*}{64}
        &23.39 & 23.42(3) &23.36(5) & 64 & 49\\
        &23.30 & 23.33(5) &23.26(4) & 128& 36\\
        &23.41 & 23.42(11) &23.37(3) & 256 & 24\\
    \midrule
    \multirow{3}{*}{256}
        &25.90 & 24.25(112) &23.79(3) & 64 & 15\\
        &29.78& 25.35(97) &23.76(29) & 128 & 6\\
        &23.22 & 23.25(12) &23.16(3) & 256 & 51\\
    \midrule
    \multirow{3}{*}{1024}
        &29.35 & 24.27(3) &23.69(4) & 64 & 9\\
        &30.60 & 23.68(5) &25.03(1) & 128 & 4\\
        &\bfseries 22.99 & \bfseries 23.06(6) &\bfseries 22.95(3) & 256 & 71\\
    \bottomrule
  \end{tabular}
  \label{tab:frob_rad}
\end{table}

\begin{figure}[h!]
  \centering
  \includegraphics[width=0.48\textwidth]{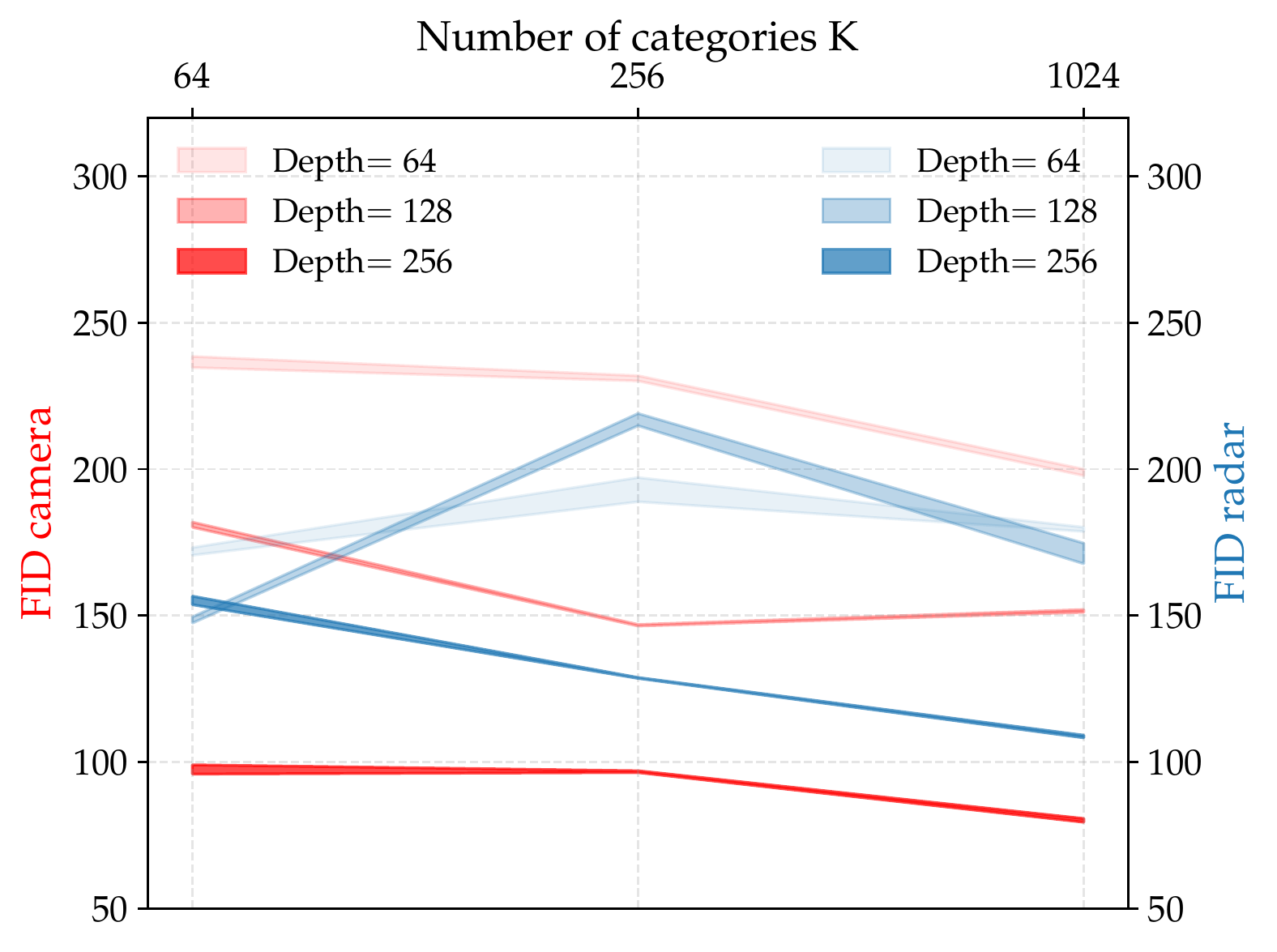}
  \caption{Fréchet inception distance for radar and camera reconstructions.}
  \label{fig:fid}
\end{figure}

\begin{figure}[h!]
  \centering
  \subfloat[Semantic camera concepts for increasing topological token order]{\includegraphics[width=0.48\columnwidth]{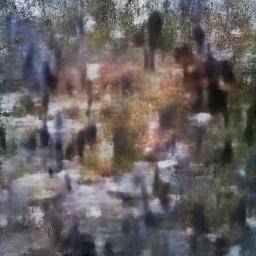}}
  \hfil
  \hspace{0.25em}
  \subfloat[Semantic radar concepts for increasing topological token order]{\includegraphics[width=0.48\columnwidth]{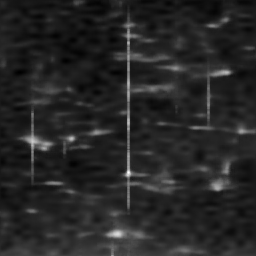}}
  \caption{Visually probing the compression models with $K=256$ by using all
    dictionary entries once in increasing order. For the camera model, roads
    segments, vegetation and pedestrian-resembling shades can be clearly
    identified underlining the adopted semantic concepts. Their visual appearance
    is also influenced by the tokens topological order in the latent space as well
    as the combination of associated embedding vectors and the decoders transposed
    upsampling convolutions.}
  \label{fig:semantic_concepts}
\end{figure}

\begin{figure}[h!]
  \centering
  \subfloat[Structure of camera latent space]{\includegraphics[width=0.48\columnwidth]{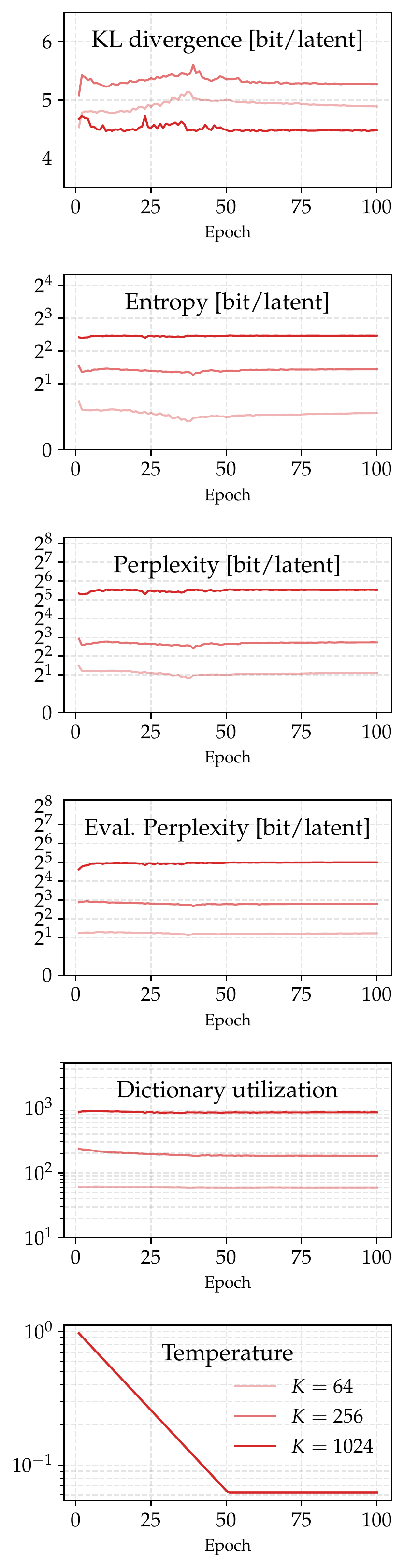}}
  \hfil
  \hspace{0.25em}
  \subfloat[Structure of radar latent space]{\includegraphics[width=0.48\columnwidth]{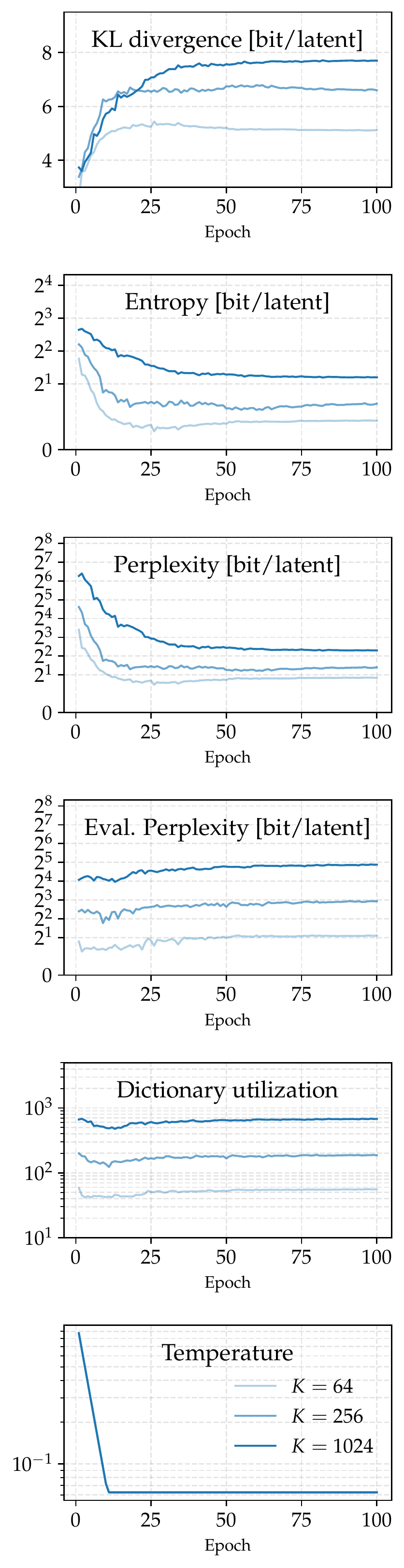}}
  \caption{Evolvement of latent spaces evaluated on the validation dataset.}
  \label{fig:perplx}
\end{figure}

\begin{figure}[h!]
  \centering \subfloat[Reconstructed pixel distributions for different
  vocabularies]{\includegraphics[width=0.48\columnwidth]{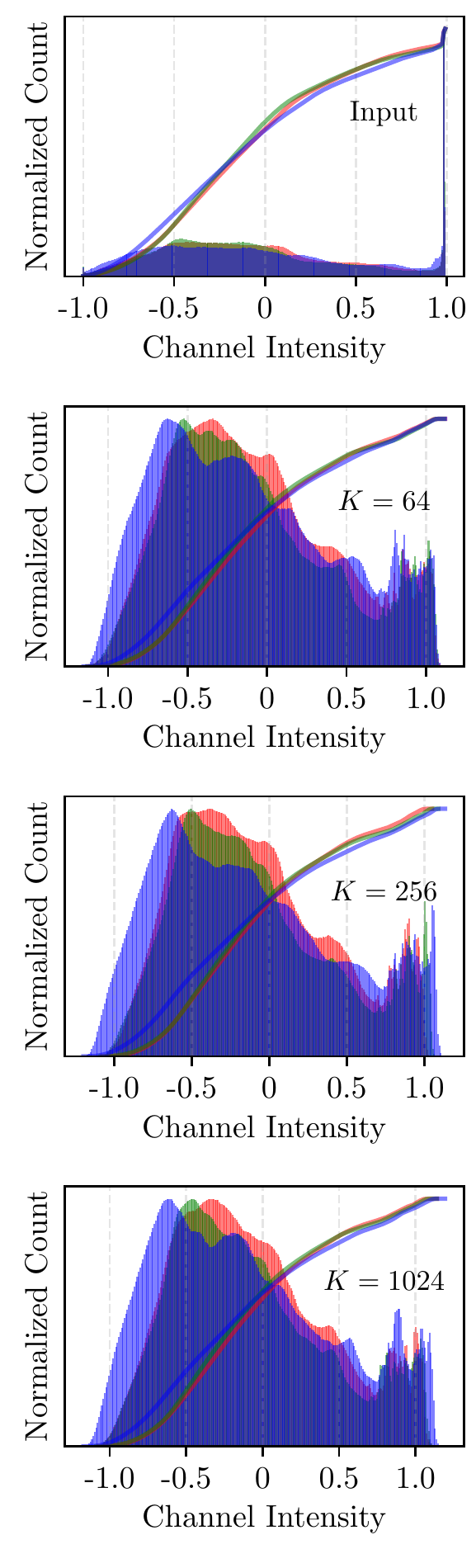}}
  \hfil \hspace{0.25em} \subfloat[Reconstructed intensity distributions for
  different
  vocabularies]{\includegraphics[width=0.48\columnwidth]{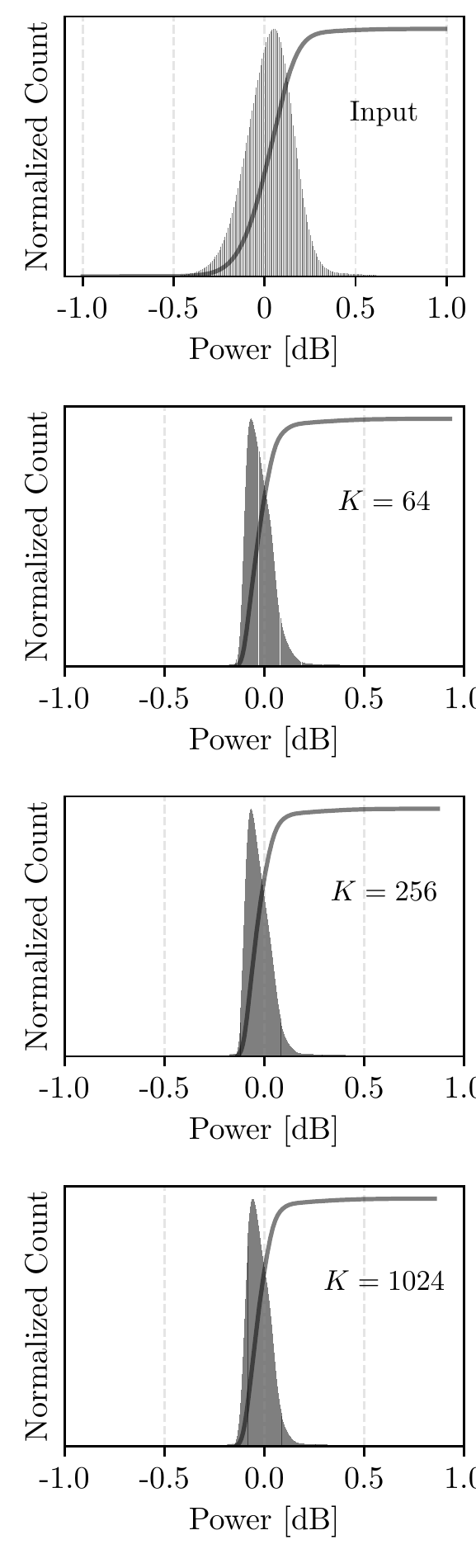}}
  \caption{Distributions of the probabilistically restored validation dataset.}
  \label{fig:recon_dist}
\end{figure}

\begin{figure*}[h]
  \centering
  \includegraphics[width=0.98\textwidth]{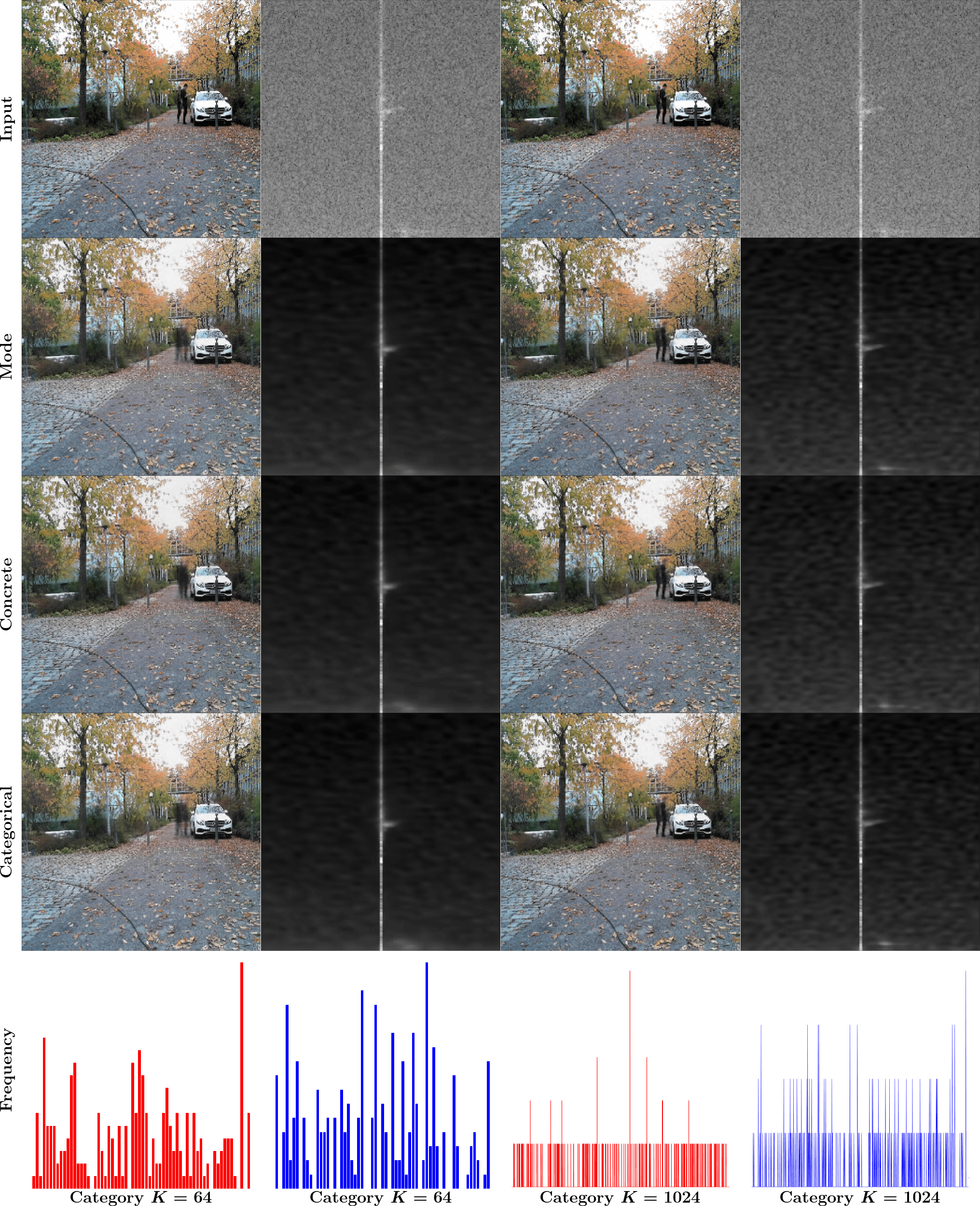}
  \caption{Camera and radar input (tow row) and probabilistic reconstructions
    for dictionary sizes of $K=64$ (left) and $K=1024$ (right). With a larger
    vocabulary size, the model is able to retain more details by assigning
    separate categories to infrequent image content. This becomes particularly
    apparent by looking at the pedestrians rendering. The reconstruction quality
    based on categorical latent samples is comparable to the restoration
    achieved by the other two sampling methods.}
  \label{fig:autumn}
\end{figure*}

\begin{figure*}[h]
  \centering
  \includegraphics[width=0.98\textwidth]{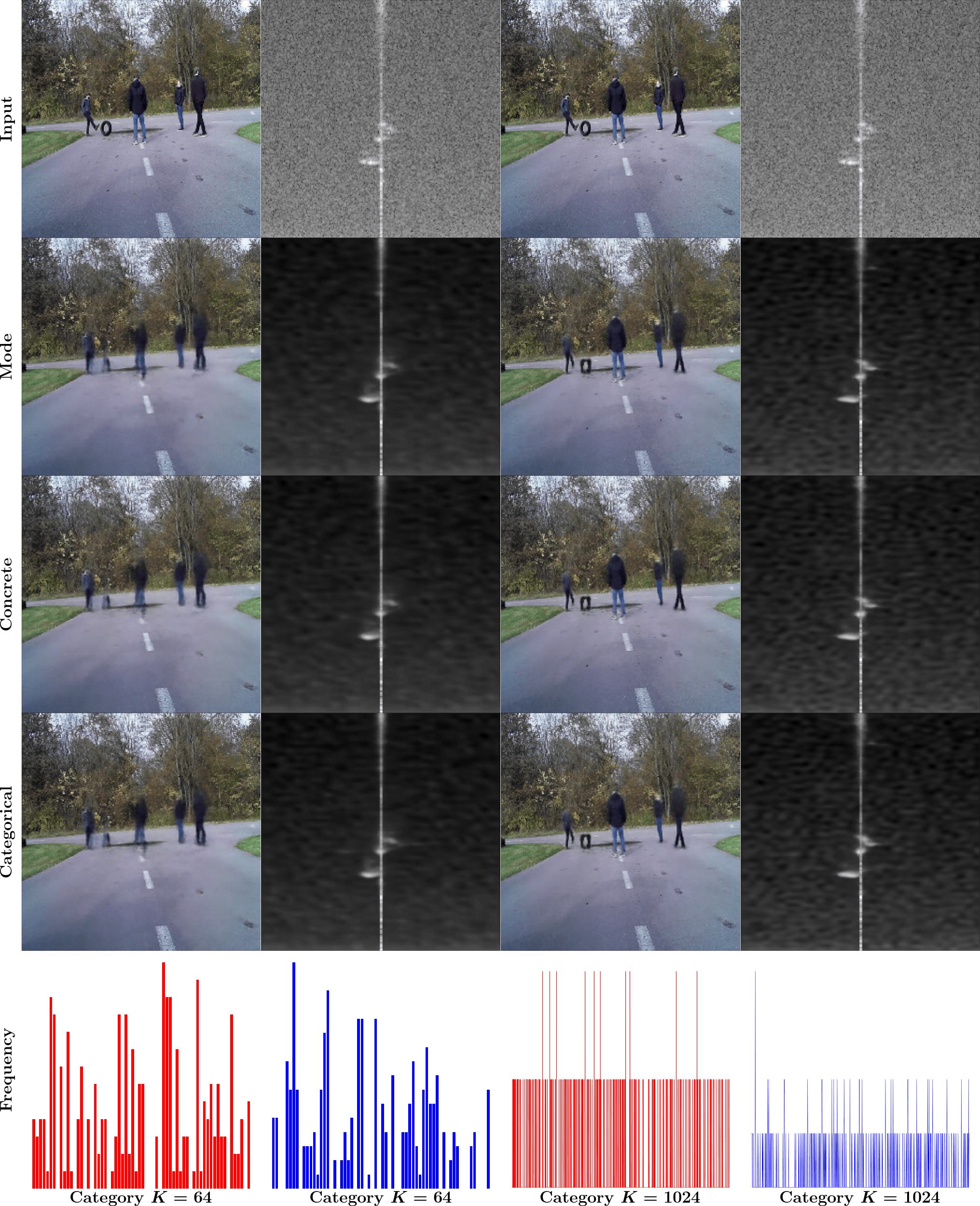}
  \caption{Camera and radar input (tow row) and probabilistic reconstructions
    for dictionary sizes of $K=64$ (left) and $K=1024$ (right). With a larger
    vocabulary size, the model is able to retain more details by assigning
    separate categories to infrequent image content. As a case in point, the
    lost tyre is barely recovered by the model with $K=64$ but is clearly
    retained by the model with $K=1024$. Radar models with larger dictionaries
    are able to resolve the noise pattern within rD plots with larger accuracy.}
  \label{fig:crossing}
\end{figure*}

\begin{figure*}[h]
  \centering
  \includegraphics[width=0.98\textwidth]{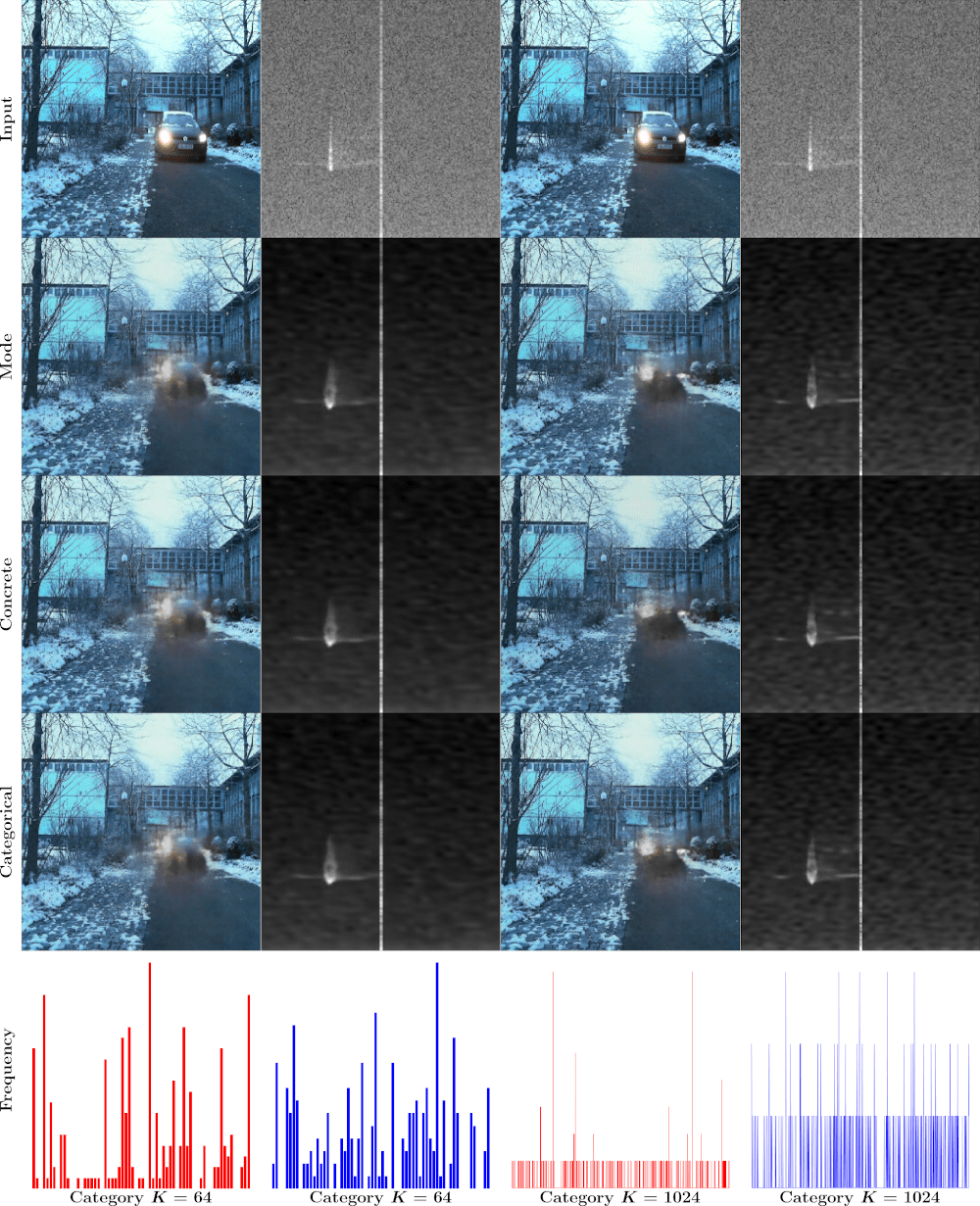}
  \caption{In this particular example, it is difficult to recognize the
    superiority of the camera model having a vocabulary size of $K=1024$ over
    the reconstruction capabilities of the model with $K=64$. The training data
    probably does not feature enough samples akin to the above to allow the
    network to reconstruct the scene in sufficient quality. The histogram plot
    speaks for a rather uniform dictionary utilization in both cases so that the
    model assigned separate categories to similar but distinct image patches.}
  \label{fig:winter}
\end{figure*}

\begin{figure*}[h!]
  \centering \subfloat[All modes for every latent variable of every PMF induced
  by the camera compression model over the validation dataset. The model seems
  to repeatedly prefer certain categories for a number of latents, indicated in
  darker gray
  color.]{\includegraphics[width=0.9\columnwidth]{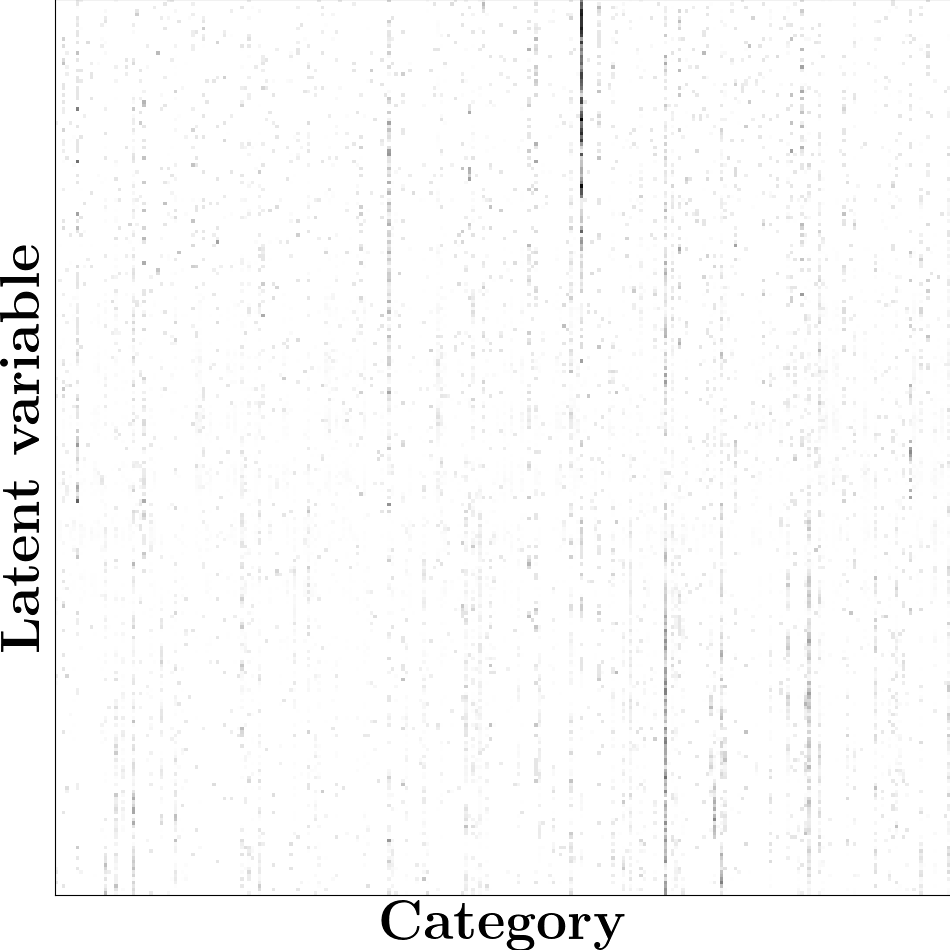}} \hfil
  \subfloat[All modes for every latent variable of every PMF induced by the
  radar compression model over the validation dataset. Presumably due to the
  large noise areas within rD maps, the model nearly assigns each
  category to every latent variable at least
  once.]{\includegraphics[width=0.9\columnwidth]{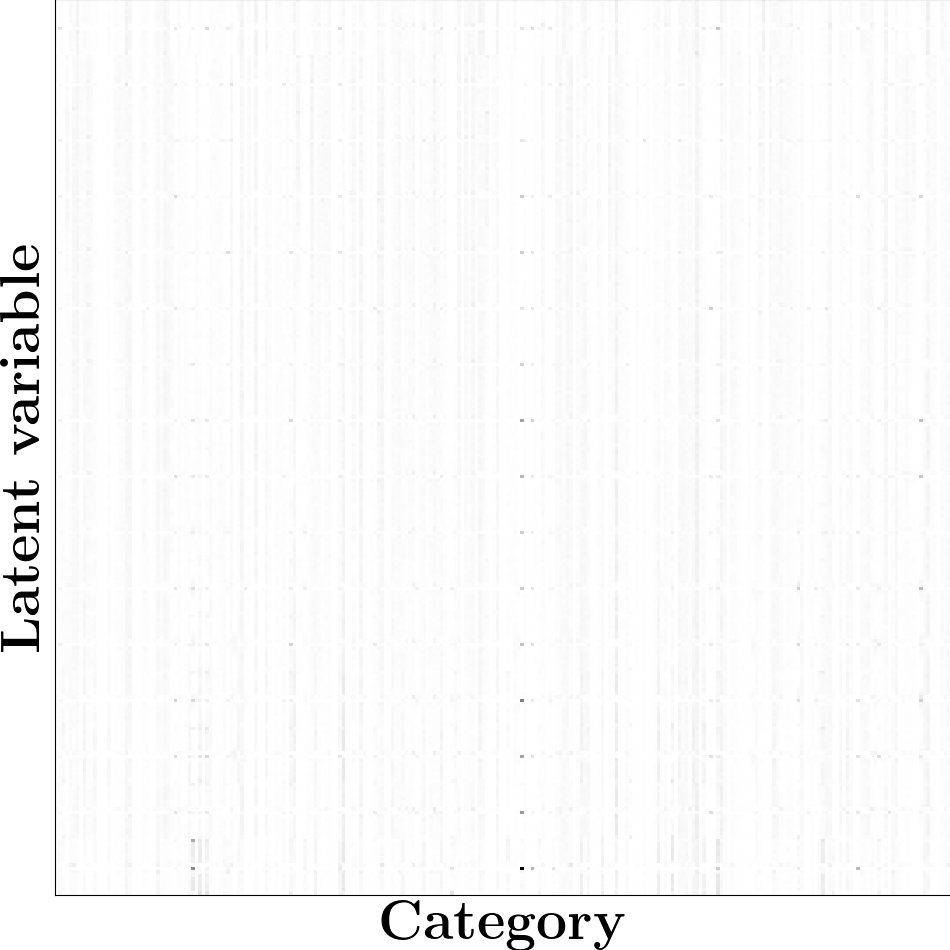}}
  \caption{Illustration of the latent code space utilization over the entire
    validation dataset. For the illustration, the modes of the $256$
    encoder-induced probability mass functions are recorded and accumulated
    across all input samples. This allows for yet another visual impression
    regarding the efficacy of the probabilistic discretization procedure and the
    desired uniform exploitation of the designed latent space. Random
    distribution patterns speak for a versatile and expressive use of the
    provided vocabulary. A smaller number of pronounced vertical lines on the
    contrary means a regular selection of that particular category by a large
    number of latent variables.}
  \label{fig:mod_util}
\end{figure*}

The depth of convolution blocks has vastly more impact on the metrics than the
dictionary size. This is not surprising given this parameter's linear influence
on the number of weights the networks have at their disposal, determining their
flexibility. As expected, the larger the number of possible latent categories,
the better the general approximation quality. However, this parameter seems to
entail only minor differences for both metrics across the entire validation
set. A possible reason is that the collection features a lot of static
backgrounds for the camera data and vast regions of noise for the radar input
which generally vary little from frame to frame and thus do not exert much
influence on the pixel-wise comparisons. It is therefore imperative to evaluate
the models' validity with a different kind of measure which mimics human visual
responses more realistically and might prove helpful in substantiating the
tendencies apparent in the tables.
\begin{table}[ht]
  \caption{Peak signal-to-noise ratio of probabilistic camera reconstructions
    for different sampling methods and varying dictionary sizes, averaged over
    three consecutive runs on the validation set.}
  \renewrobustcmd{\bfseries}{\fontseries{b}\selectfont}
  \sisetup{detect-weight,mode=text,group-minimum-digits = 4}
  \centering
  \begin{tabular}{l
    S[table-format = 2.2]
    S[table-format = 2.2(1), separate-uncertainty]
    S[table-format = 2.2(1), separate-uncertainty]
    |c
    |c}
    \toprule
    $K$ & {Mode} & {Categorical}  & {Concrete}  & Depth & Epoch\\
    \midrule
    \multirow{3}{*}{64}
        &16.86 & 16.85(1) & 16.89(1) & 64 & 122\\
        &18.47 & 18.45(6) & 18.53(3) & 128 & 141\\
        &20.58 & 20.54(4) &20.58(8) & 256 & 68\\
    \midrule
    \multirow{3}{*}{256}
        &17.17 & 17.17(5) &17.17(8) & 64 & 156\\
        &18.98 & 18.98(1) &18.99(7) & 128 & 146\\
        &20.64 & 20.66(2) &20.73(3) & 256 & 59\\
    \midrule
    \multirow{3}{*}{1024}
        &17.66 & 17.64(5) &17.72(1) & 64  & 156\\
        &19.17 & 19.13(2) &19.18(1) & 128 & 142\\
        &\bfseries 20.94 & \bfseries 20.87(7) &\bfseries 20.86(3) & 256 & 75\\
    \bottomrule
  \end{tabular}
  \label{tab:psnr_cam}
\end{table}
\begin{table}[ht]
  \caption{Peak signal-to-noise ratio of probabilistic radar reconstructions for
    different sampling methods and varying dictionary sizes, averaged over three
    consecutive runs on the validation set.}
  \renewrobustcmd{\bfseries}{\fontseries{b}\selectfont}
  \sisetup{detect-weight,mode=text,group-minimum-digits = 4}
  \centering
  \begin{tabular}{
    l
    S[detect-weight,group-minimum-digits = 4,table-format = 2.2]
    S[table-format = 2.2(1), separate-uncertainty]
    S[table-format = 2.2(1), separate-uncertainty]
    |c
    |c}
    \toprule
    K & {Mode} & {Categorical}  & {Concrete}  & Depth & Epoch\\
    \midrule
    \multirow{3}{*}{64}
      &20.81 & 20.80(5) &20.82(3) & 64 & 49\\
      &20.84 & 20.83(56) &20.86(11) & 128& 36\\
      &20.80 & 20.79(12) &20.81(4) & 256 & 24\\
    \midrule
    \multirow{3}{*}{256}
      &19.92 & 20.49(7) &20.66(3) & 64 & 15\\
      &18.75 & 20.10(3) &20.67(1) & 128 & 6\\
      &20.87 & 20.86(1) &20.89(2) & 256 & 51\\
    \midrule
    \multirow{3}{*}{1024}
      &18.88 & 20.48(3) &20.69(9) & 64 & 9\\
      &18.52 & 20.21(2) &20.70(1) & 128 &4\\
      &\bfseries 20.96 & \bfseries 20.93(8) &\bfseries 20.97(1) & 256 &71\\
    \bottomrule
  \end{tabular}
  \label{tab:psnr_rad}
\end{table}
This other method for measuring the faithfulness of generative models has been
proposed by \cite{heusel2017gans} in form of the so-called \textit{Fréchet
  inception distance} (FID) as an improvement over the Inception score (IS)
\cite{salimans2015improved} widely used before. It evaluates the quality of
generated images by determining their distance to the originals through
higher-level features rather than pixel-wise. The rationale here is that feature
embedding vectors are more likely to correspond to real-world objects and image
constituents akin to how a human would perceive similarity and judge visual
quality. To this end, both original and generated data are independently fed
into the \textit{Inception-v3} model \cite{szegedy2016rethinking} pretrained on
the ImageNet \cite{deng2009imagenet} dataset. Optimized to predict the contained
\num{1000} object classes to high accuracy, this model serves as a feature
extractor, effectively transforming high-dimensional images into a
lower-dimensional latent space in which similar input should have a certain
proximity. Tapping into its architecture after the last pooling layer allows to
summarize its \num{2048} activations as multivariate Gaussians by fitting mean
and covariance to the respective data distribution under consideration. The FID
then is the Wasserstein distance between the corresponding first two statistical
moments calculated over the entire validation dataset and three averaging runs
on its probabilistic reconstructions
\begin{align}
  \text{FID} =\lVert  \bm \mu_{d}-\bm \mu_{g}\rVert^2_2+\Tr
  \left[\bm \Sigma_{d}+\bm \Sigma_{g}-2\left(\bm \Sigma_{d}\bm \Sigma_{g}\right)^{\frac{1}{2}}\right].
  \label{eq:fid}
\end{align}
The smaller this distance, the more related two datasets are and the finer the
reconstruction quality is believed to be. Consequently this value vanishes for
identical distributions. FID is far from an optimal measure as it reveals a
strong bias towards the size of the datasets. It does, however, provide a more
elaborate insight into the models inner workings and recognizes qualitative
tendencies better than any other metric introduced so far as it represents the
current state-of-the-art for the evaluation of generative models as of 2021. The
FID for both modalities is depicted in Figure \ref{fig:fid} as a function of
dictionary size ablated once again over the channel depth of the networks. To
bring the radar input into agreement with the first layers of the inception
model, the rD maps were replicated thrice before feeding them into the network.
Surprisingly, the plot shows non-monotonous graphs for some of the smaller
network depths, but overall confirms the observations made before about the
larger dictionaries being most potent regarding the maximization of the VLB
objective. The metrics significance about the frequency plots remains
questionable to say the least, but nonetheless offers an opportunity for
intra-modal comparison. From here on out, only models with a kernel number of
$256$ will be used further as those undoubtedly produce the most favorable
results. To obtain a comprehensive notion of the models' versatility, their
latent space utilization for a size of $K =|\mathcal{C}|=256$ is recorded
separately for every latent variable over the validation dataset. To yield a
reproducible result, the modes of the data-induced PMF, given by equation
\eqref{eq:hardsampled} are accumulated for every input sample and displayed in
Figure \ref{fig:mod_util} for both domains. Pronounced vertical lines indicate
frequent use for that particular category across many latent variables whereas
less regular and interrupted structures speak for a more evenly utilized code
space. The learned semantic concepts and acquired knowledge of the
discretization models with $K =|\mathcal{C}|=256$ can also be explored by
querying each vocabulary entry exactly once and in increasing order before
decoding the resultant continuous feature space, shown in Figure
\ref{fig:semantic_concepts}. To give a visual impression of the models' variety,
both generators were also probed for their acquired knowledge by successively
selecting indices from $\bm c=1$ to $\bm c=256$ one at a time for all latents.
Figure \ref{fig:mod_patches} in the appendix shows the results of successively
decoding these identical categories into individual images. The more distinct
these images, the more expressive the latent codes and the more effective the
related discretization schemes are assumed to be. Likewise, a large number of
similar looking patches points towards potential redundancies in the
dictionary. A clear sign, that the model does not fully exploit the potential of
the discrete code but instead has found convenient shortcuts to fulfill the
objective during training. Additionally, Figure \ref{fig:fail} and Figure
\ref{fig:fail2} in the appendix show typical failed examples and degenerated
modes experienced during various stochastic compression attempts. As pointed out
before, training VAEs is a difficult trade and minor deviations from optimal
hyperparameter choices can cause the learning to turn unstable. Concretely, in
the first few iterations, the reconstruction term in the VLB has little impact
on the overall loss. Approximate compliance of the posterior with the uniform
prior can then represent a local equilibrium configuration from which the
algorithm can have problems to escape. Successful validation results of selected
model runs are displayed for completeness in Figure \ref{fig:valid_runs} in the
appendix. Furthermore, Figure \ref{fig:roundabout} to Figure \ref{fig:winter}
give visual impressions of the qualitative reconstruction differences between
vocabulary sizes of $64$ and $1024$, respectively, alongside the normalized
histograms of the utilized latent codes. It becomes evident that dictionaries of
all sizes are able to stochastically decompose and reassemble diverse real-world
data, albeit with varying level of detail. Especially camera content located in
the far background can be reproduced significantly sharper with a greater
dictionary. Concerning the crispness of rD maps, larger vocabularies clearly
enable superior recovery of noise patterns and shape reproduction of complex
intensity clusters. This is of vital importance, for ignorance about every
single radar cell neglects all information within a
$\SI{0.1}{\m}\times\pm\SI{0.02}{\meter\per\second}$ interval. As it turns out,
accurately predicting additive noise, omnipresent in rD plots is extremely
difficult, noticeable among other things in clear contrast differences between
input and reconstructions. VAEs in general are known for inflicting a certain
amount of blur onto their output which is clearly recognizable particularly with
regard to dynamic objects or the precise modeling of intensity pattern in radar
input. The rationale for this is manifold: For one thing, the feature size
limits imposed on the bottleneck and the struggle of conforming to the prior
while having the L2 criterion evaluating the output means leads to a decline in
visual quality. Second, the model has seen multitudes of examples featuring the
many backgrounds and large areas of noise whereas vulnerable road users (VRU),
for example, have been presented to it only so many times and always in
different poses and settings. Also, individuals or moving objects and
reflections tend to occupy only small pixel regions of the images. Covering
those more precisely, the model would have to allocate an individual category
for each of these instances included in the dataset. Given the limited number of
tokens, the input can be split into, this makes it drastically harder for the
network to accurately restore the remaining image parts. It is thus more likely
that it chooses to represent dynamic elements and aggregated reflections as
compositions of several discrete constituents instead, relying on the decoder to
combine and assemble them correctly. To components, frequently present in images
like skylines or road segments, the model probably dedicates separate
categories, effectively using them as basic building blocks for the majority of
image generations. Videos showcasing the specific restoration capabilities for
exemplary scenes in a more vivid form can be found at
\href{https://cditzel.github.io/GenRadar/}{cditzel.github.io/GenRadar}. As
pointed out in chapter \ref{sec:method} the current approach depends on a
versatile and expressive dictionary to unambiguously compress continuous-valued
input into discrete sequences by preferably large margins in the decision
process. As seen before, this is rather challenging, especially concerning the
spectral plots with a lot of noise present across the entire depiction. Hence,
the goal has to be inhibiting the model from assigning a single token repeatedly
to radar patches which predominantly comprise noise. Enlarging the number of
categories the models can choose from in the regularized latent space should
make for enhanced disentanglement and better data representation of these
discrete image parts. In other words, a more distinguished vocabulary makes for
improved assignments of image patches to integers, thus aiding the
reconstruction altogether. To examine this claim in more detail, the total
amount of all possible information the models encode is investigated by
calculating their perplexity. After a change to binary base, this can be
expressed via entropy exponentiation
\begin{align}
  \text{perplexity}(q)  =  \prod_{\bm c}q_{\bm \phi}(\bm c\mid\bm x)^{-q_{\bm \phi}(\bm c\mid\bm x)} = 2^{\ent[q_{\bm \phi}(\bm
  c\mid\bm x)]}.
\end{align}
Figure \ref{fig:perplx} contrasts this variational entity of the camera with
that of the radar models across consecutive validation runs performed after
every training epoch. All measures are with regard to a single latent symbol
facilitating the association with an actual number of bits required to transmit
its state through the network. Figuratively speaking, perplexity measures the
amount of randomness in the model and quantifies how well the associated process
predicts samples. It calculates the weighted average number of choices each
latent variable is offered. Referring to the categorical encoder introduced in
section \ref{sec:cvae} this means, the higher the perplexity the more uniformly
distributed the underlying stochastic selection of latent constituents
is. Conversely, a lower number indicates that the model grows more confident at
predicting the category an image patch should belong to. In view of Figure
\ref{fig:perplx}, this aligns with the formerly stated conflict about the VLB in
equation \eqref{eq:beta}.
Making the bound approach the marginal likelihood demands a decrease in KL
divergence during training, which can only be achieved by the posterior
converging towards equilibrium. This raises the system's entropy, but would make
for a more homogeneous use of the dictionary at the expense of more bits
necessary to represent the compression. For a lowest possible data-restoration
error and likelihood maximization, less randomness of the feature choice is
preferable, but alienates the categorical posterior from the prior, visible by a
gain in KL-divergence. Though this generally reduces entropy and the number of
bits required to communicate the latent state it involves the risk of ending up
with only a small amount of effectively used codes. This does not necessarily
show in the reconstruction quality, as the latent feature vectors
$\bm e \in\R^D$ might still attain a broad knowledge during the course of training
but will certainly impede the autoregressive prediction, subject of the next
section. The inverse relations between the described quantities are clearly
recognizable by contrasting the first two rows of illustration
\ref{fig:perplx}. The noted arguments underline the importance of finding a
suitable combination of the divergence and temperature parameter which both
exert major influence on the arrangement of the latent spaces. For reference,
the plot also lists the effective perplexity for which discrete samples rather
than distributions were taken as a basis for evaluation. Deviations between both
measures of perplexity, particularly visible for the radar models, might bespeak
a less confident feature selection process. One problem associated with
perplexity in general is its dependence on the employed batch size. Therefore,
also the actual bin count of the vocabulary histogram over one complete
validation epoch is included in the graphs, giving a more reliable impression of
the latent space exploitation. Differences in divergence weights and
temperatures make a direct comparison between both modalities
difficult. Interestingly though, the radar models generally seem to settle for
more customized latent distributions and eventually, content with higher
structured latent spaces. Referring to the practical dictionary usage, however,
displays similar developments of the utilization than those of the camera models
after stabilization past epoch \num{25}. Further, the reconstructed pixel and
intensity distributions for the final model selection is shown in Figure
\ref{fig:recon_dist} in contrast to the dataset distributions previously given
in Figure \ref{fig:rad_histogram_valid_pure} and Figure
\ref{fig:cam_histogram_valid_pure} which are repeated here for convenience.

After having acquired stochastic compression models for both modalities, their
holistic processing procedure is now abandoned by breaking apart the categorical
autoencoders into the inference networks and the generators, including the
adjusted upstream vocabularies. In the next chapter, the former ones will serve
as tokenizers for the continuous-valued multi-modal input, whereas the latter
will come into play again when the modeled token sequences are to be decoded
into an intuitive and vivid image data space.

\subsection{Autoregressive Modeling of Measurement Constituents}
\label{sec:trafo}
Ever since their introduction in 2017, Transformers \cite{vaswani2017attention}
have seen growing interest and success in both application and research
alike. Originally invented with the purpose of advancing machine translation,
variations of the original architecture have made their way into various areas
like audio generation \cite{li2019neural} or bio-informatics to simulate protein
folding \cite{senior2020improved} underlining their remarkable
capabilities. More recently, they were also used for image classification
\cite{dosovitskiy2020image} and image generation on a per-pixel basis
\cite{chen2020generative} questioning the prevalent role convolutional
structures held in this field roughly since 2012 and the famous publication of
Alexnet \cite{krizhevsky2012imagenet}, albeit with thus far unheard-of use of
computational resources. In the following, transformers will be used to generate
consecutive integer sequences and conditionally predict camera image components
for given radar tokens.

\subsubsection{Multiheaded Self-Attention}
The unique trademark of any transformer architecture is the incorporated
attention mechanism, dynamically modeling relations between all discrete input
tokens, irrespective of their relative distance. In contrast to convolutional
kernels and their local receptive fields, attention layers lack any inductive
biases or preemptive data-based assumptions. Deprived of CNNs key features,
transformers instead are given the liberty to form cross-connections between
data points during training virtually unconstrained deciding which input to
attend to more or less strongly. This allows them to effectively capture global
structures and to account for possible correlations among the entire
one-dimensional input. Yet, complexity grows quadratically with sequence length
$N$ for reasons of pairwise inner products in the dot product formulation of
full attention
\begin{align}
  \bm A_h = \left(\bm Q_h\bm K^\top_h\right)/\sqrt{d_h} \quad \in \mathbb{R}^{N\times N}
  \label{eq:att}
\end{align}
in which $\bm Q_h,\bm K_h\in\R^{N\times d_h}$ represent query and key tensors. These
entities are the result of linearly contracting preceding sequence
representations with trainable projections $\bm
W^{q}_h$,$\bm W^{k}_h\in\R^{D\times d_h}$. This module thus represents a form of
soft-attention, since it relies on weighted sums of embedding vectors for its
calculation rather than determining hard scores between all inputs
directly. After scaling with the feature dimension length $d_h$ for an alleged
increase in numerical stability, attention matrix $\bm A\in\R^{N\times N}$ gives the
inter-token attention scores between all inputs, where each element signifies
the pairwise dot-product similarity of the respective key and query vectors. Yet
another learnable linear operator $\bm W_h^{v}\in\R^{D\times d_h}$ is applied to the
former sequence features yielding value matrix $\bm V_h\in\R^{N\times d_h}$ which is
multiplied with the attention map upon softmax normalization
\begin{align}
  \bm H_h = \text{softmax}\left(\bm A_h\right) \bm V_h \quad \in \mathbb{R}^{N\times d_h}.
  \label{eq:head}
\end{align}
The described model is reminiscent of information retrieval systems in which a
database is queried via its keys for associated values in a data-adaptive
importance-weighted manner. In modern transformers, this entire process is
replicated and executed separately $H$ times in parallel for this so-called
multihead attention supposedly increases model capacity and according to
\cite{vaswani2017attention} allows it to attend to information from different
representation subspaces at the same time. To aggregate all heads for subsequent
calculations they are concatenated along their feature dimension
\begin{align}
  \textbf{{MSA}}= \left[\bm H_1\mid\bm H_2|\dots|\bm H_H\right] \bm W~~ \in \mathbb{R}^{N\times D}
\end{align}
before multiplication with a final learnable linear matrix
$\bm W\in \R^{D\times Hd_h}$. Followed by pointwise fully-connected blocks these two
elements lay the foundation for nearly every modern transformer
architecture. Supplementing both with prefixed layer normalization (LN) for
gradient recentering \cite{ba2016layer} and introducing skip connections around,
arguably reduces training time and leads to improved generalization
capabilities. To induce a hierarchical structure similar to convolutional
networks, numerous of these layers are generally stacked sequentially $L$ times
for an increase in depth usually gives superior performance. For a more detailed
explanation of this seminal concept and its underlying algorithmic approach, see
\cite{alammar2018illustrated}. Lacking any understanding of spatial
relationships within its input sequence, another common characteristic of these
autoregressive models is the incorporation of locality information by the
addition of positional encoding vectors
$\bm \emb_{\text{pos}}\in \mathbb{R}^{N\times D}$. These typically take either the form
of discretized trigonometrical functions or rotation matrices
\cite{su2021roformer}. Lately it has been demonstrated though, that even those
vectors can be learned in a data-dependent manner, which further reduces manual
interference and thus has been followed here.

\subsubsection{Multi-Modal Token-Based Likelihood Estimation}
\label{sec:trafo_training}
\begin{figure}[h!]
  \centering
  \includegraphics[width=0.48\textwidth]{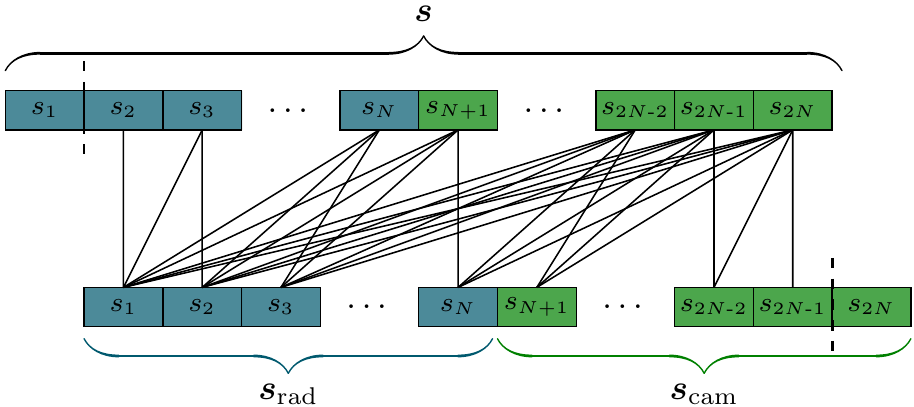}
  \caption{Cross-modal autoregressive sequence modeling with causal attention.}
  \label{fig:autoregression}
\end{figure}

Transformer models really shine at sequence modeling, a task in which they are
to predict the next part of a certain section, given only the preceding elements
to rely on. This so-called autoregressive training with discrete tokens is used
in the following, to conditionally forecast camera patches based on supplied
radar-frequency data within predefined uncertainty bounds. Concretely,
time-synchronized samples of both domains are encoded into their discrete
counterparts by means of the pretrained modal-specific encoders, as explained in
section \ref{sec:dvae}, with all of their weights frozen. Running in inference
mode, these now parameterize true categorical distributions, cf. equation
\eqref{eq:categorical}, which are sampled from for every latent variable
corresponding to $16\times 16$ partially overlapping patches of the original
$256\times 256$ images as described in section \ref{sec:training}. Both models are
now used as priors for the density estimation over sequences outlined in the
following. As camera and radar input dimensions are fixed a priori, so is the
length $N = h \times w$ of both sequences created line by line using raster order for
radar
\begin{align}
  \overset{\R^{H\times W}}{\bm x_{\text{rad}}}\xrightarrow{\text{encode}}
  \overset{\R^{h\times w\times K}}{\bm \pi_{\text{rad}}}\xrightarrow{\text{sample}}
  \overset{\mathbb{N}^{h\times w}}{\bm
  c_{\text{rad}}}\xrightarrow{\text{reorder}} \overset{\mathbb{N}^N}{\bm
  s_{\text{rad}}}
  \label{eq:radtoken}
\end{align}
and camera input simultaneously but independently
\begin{align}
  \overset{\R^{H\times W\times 3}}{\bm x_{\text{cam}}}\xrightarrow{\text{encode}}
  \overset{\R^{h\times w\times K}}{\bm \pi_{\text{cam}}}\xrightarrow{\text{sample}}
  \overset{\mathbb{N}^{h\times w}}{\bm
  c_{\text{cam}}}\xrightarrow{\text{reorder}} \overset{\mathbb{N}^{N}}{\bm
  s_{\text{cam}}}.
  \label{eq:camtoken}
\end{align}
The camera tokens are subsequently appended to the radar series forming a
single common multi-modal sequence
\begin{align}
  \bm s = \left[\bm s_{\text{rad}}|\bm s_{\text{cam}}\right]\quad \in\mathbb{N}^{2N}
  \label{eq:commonseq}
\end{align}
which serves as input for the autoregressive training process. Given its
constant length, no elaborate masking or complex token manipulation needs to be
devised. Rather, a simple one-off scheme suffices for the classification setting
at hand. After discarding the final element of the above sequence, radar tokens
predict their respective successors with the last radar token
$s_{\scriptscriptstyle N}$ autoregressively projecting the first camera token at
position $s_{\scriptscriptstyle N+1}$. Progressing in this manner, the last
camera token $s_{\scriptscriptstyle 2N}$ is eventually determined by its
immediate predecessor $s_{\scriptscriptstyle 2N-1}$. Consequently, the target
vector is formed by the original integer sequence given in equation
\eqref{eq:commonseq} after truncating its first element to facilitate
autoregressive prediction as illustrated in Figure \ref{fig:autoregression}.
The model is allowed to make decisions about a current token
$s_{\scriptscriptstyle i}$ based only on information derived from preceding
positions $s_{\scriptscriptstyle 1:i-1}$ which is realized by imposing causal
masks on the attention mechanism within every layer. Most often, these are
implemented by setting all but the lower triangular matrix in equation
\eqref{eq:att} to minus infinity prior to softmax normalization. For a certain
token under consideration this then yields negligible scores if contracted with
the value tensor entries of subsequent elements in the training sequences, and
effectively prohibits positions from looking ahead. The training objective in
this classification stage is to maximize the log-likelihood of the token
permutations given discrete samples of probabilistically compressed synchronous
multi-modal input
\begin{align}
  \argmax_{\bm \psi}\E_{\substack{\bm x\sim p(\bm x)\\ \bm s \sim q_{\bm \phi}(\bm s\mid \bm x)}}
  \left[\log p_{\bm \psi}(\bm s) \right].
  \label{eq:trafo_loss}
\end{align}
This is equivalent to minimizing the cross-entropy between data-conditional
priors in the form of both encoder models and estimated discrete distributions
over possible categories associated with each dual-domain sample
\begin{align}
  \lagr_{\bm \psi}(\bm x) = -\E_{\bm s\sim q_{\bm \phi}(\bm s \mid\bm x)} \left[\log
  p_{\bm \psi}(\bm s) \right] = \ent\left[q_{\bm \phi}(\bm s \mid\bm x), p_{\bm \psi}(\bm s)\right].
  \label{eq:ce}
\end{align}
Given above token series $\bm s$ of both modalities, $p_{\bm \psi}(\bm s)$ denotes
a transformer model emulating the inter-token dependencies
\begin{align}
  \bm \pi = \text{transformer}_{\bm \psi}(\bm
  s_{\scriptscriptstyle 1:2N-1}) \qquad \bm s \stackrel{\text{iid}}{\sim} q_{\bm \phi}(\bm s\mid \bm x).
\end{align}
This sequence model decomposes the joint distribution into factors
parameterizing $2N-1$ categoricals over possible vocabulary entries $K$ of each
domain according to
\begin{align}
  \begin{split}
    \log p_{\bm \psi}(\bm s_{\scriptscriptstyle 2:2N}) &= \log \prod^{2N}_{i=2}p_{\bm \psi}(s_i\mid \bm s_{\scriptscriptstyle 1:i-1}) \\
    &= \sum^{2N}_{i=2}\log\text{Cat}(s_i;\bm \pi_i(\bm s_{\scriptscriptstyle 1:i-1})).
  \end{split}
\end{align}
Above equation describes a directed graphical model so that the predicted
distribution for any token position is influenced only by the previous section
of the input sequence. Masking procedures are applied to the raw logits
$\bm \pi$ to make the first section
$\bm \pi_{\scriptscriptstyle 2:N}\in \R^{N-1\times2K}$ only predict radar constituents
$\bm \emb_{\text{rad}}\in \R^{K\times D}$ whereas the second part
$\bm \pi_{\scriptscriptstyle N+1:2N}\in \R^{N\times 2K}$ including the mass function
$\bm \pi_{\scriptscriptstyle N+1}\in \R^{2K}$ associated with the last provided
radar entry $s_{\scriptscriptstyle N}$ is restricted to yield probabilities
exclusively over the camera dictionary
$\bm \emb_{\text{cam}}\in \R^{K\times D}$. During training, minimizing the
cross-entropy in equation \eqref{eq:ce} thus effectively stimulates the
transformer to shift probability mass towards the one-off input token
configuration for a given example sequence, improving the prediction accuracy
during the course of several runs across the dataset. No sampling takes place
along the backpropagation paths, ending just before both inference models which
are drawn from merely to provide the network input and ground truth. The
objective therefore can be optimized directly via end-to-end training by
replacing the expectations with Monte Carlo sampling and adjusting the
transformer parameters $\bm \psi$ with batchwise gradient estimates.
\subsubsection{Implementation of the Multi-Modal Sequence Model}
In view of the concrete objective of this work, certain modifications and
adjustments were necessary to the general transformer framework described in the
former section.
\begin{algorithm}[h!]
  \caption{Autoregressive next-token prediction}
  \label{alg:alg1}
  \begin{algorithmic}[1]
    \renewcommand{\algorithmicrequire}{\textbf{Input:}}
    \renewcommand{\algorithmicensure}{\textbf{Output:}}
    \REQUIRE Discrete sequences $\bm s_{\text{rad}}\leftarrow\bm x_{\text{rad}}~ \text{and}~ \bm s_{\text{cam}}\leftarrow\bm x_{\text{rad}}$
    \ENSURE Multinomials over successive token sequence\\

    \textit{Init}: $\bm W_\psi,\bm \emb_\psi,\bm \emb_{\text{pos}} \sim \N(0,0.02),\bm b_\psi =0, \bm \gamma = 1, \bm \beta=0 $\\
    \STATE $\bm s \gets \text{concat}\left(\bm s_{\text{rad}} , \bm
      s_{\text{cam}}\right),~\bm s \in\mathbb{N}^{2N-1}$
    \STATE $\bm z_0 \gets \left[s_1 \bm \emb_1;\dots;s_{2N-1} \bm \emb_{2N-1}\right]+ \bm \emb_{\text{pos}} ,~ \bm z_0\in\R^{2N-1\times D}$
    \FOR {$\ell = 1$ to $L$}
    \STATE $\bm z_{\ell} \gets \text{LN}^1_{\ell_{\bm\beta,\bm\gamma}}(\bm z_{\ell-1})$
    \FOR {$h = 1$ to $H$}
    \STATE $\bm Q_{\ell h} \gets \bm z_\ell \bm W^q_{\ell h}+\bm b^q_{\ell h}$
    \STATE $\bm K_{\ell h} \gets \bm z_\ell \bm W^k_{\ell h}+\bm b^k_{\ell h}$
    \STATE $\bm V_{\ell h} \gets \bm z_\ell \bm W^v_{\ell h}+\bm b^v_{\ell h}$
    \STATE $\bm A_{\ell h}\gets\text{masking}(\bm Q_{\ell h} \bm K^\top_{\ell h}/\sqrt{d_h})$
    \STATE $\bm H_{\ell h}\gets\text{softmax}(\bm A_{\ell h})\bm V_{\ell h}$
    \ENDFOR
    \STATE $\bm z_{\ell} \gets \bm z_{\ell-1} +\text{concat}\left(\bm H_{\ell1},\bm
      H_{\ell2},\dots,\bm H_{\ell H}\right) \bm W_\ell$
    \STATE $\bm z_{\ell} \gets \bm z_{\ell} + \text{gelu}\left(\text{LN}^2_{\ell_{\bm\beta,\bm\gamma}}(\bm
      z_{\ell})\bm W^1_\ell + \bm b^1_\ell\right)\bm W^2_\ell + \bm b^2_\ell$
    \ENDFOR
    \STATE $\bm \pi_{\bm \psi} \gets \text{LN}_{\bm\beta,\bm\gamma}(\bm
    z_{\ell})\bm W + \bm b, \quad \bm W\in\R^{D\times 2K},b\in\R^{2K}$
    \RETURN $\bm \pi_{\bm \psi} \quad \in\R^{2N-1 \times 2K}$
  \end{algorithmic}
\end{algorithm}
The algorithm starts by embedding each sequence token separately via a learnable
lookup table $\bm \emb_\psi\in \R^{2N-1\times D}$ prior to element-wise addition of
learnable positional information. The multi-head attention calculations $h$ in
every layer $\ell$ are succeeded by two adaptable linear operators
$\bm W^1 \in \R^{D\times 2D}$, $\bm W^2 \in \R^{2D\times D}$ with interleaved Gaussian Error
Linear Units (GELU) \cite{hendrycks2016gaussian}. Finally, a classification head
is attached, consisting of another linear projection $\bm W \in \R^{D\times2K}$
followed by a single layer norm, which yields unnormalized probabilities for the
intended prediction task. The forward signal flow of a single multi-modal input
sequence during the data-fitting process is summarized in Algorithm
\ref{alg:alg1}. For improved conditioning of the system matrices, weights of
lookup tables and positional embeddings as well as for all linear operators were
initialized with values drawn from normal distributions with biases enabled but
initially set to zero. All normalization layers had their learnable scale
parameters assigned to one without any initial shift. Additionally, dilution, as
proposed in \cite{srivastava2014dropout}, was employed quite aggressively at
various positions contributing to the overall model regularization. Aside from
its concrete application after step \num{2} and after the softmax operation in
step \num{10} of the algorithm, dropout was also used following the attention
and feed-forward blocks in step \num{12} and \num{13} just before adding the
skip connections. In combination with input sampling, this measure is to prevent
the model from just memorizing token configurations of the training dataset
which would entail poor performance on the validation set. Randomly switching
off activations during training effectively prunes the network and forces it to
continuously find novel features in combination with a varying number of
different neurons. The probability of disabling nodes was chosen to
$p=\num{0.25}$ and slightly but constantly increased with layer index. The
latent feature dimension $D$ was kept constant throughout the entire network and
at an integer multiple $H$ of the latent dimension of each head $d_h$ so that
their concatenation yields the original feature size again. This is a deliberate
decision though, simplifying numerical treatment rather than required by the
algorithm itself. Since the relative positions of intensity clusters in rD maps
hold immediate meaning and establish direct correspondences to integral physical
quantities like range and velocity (cf. section \ref{sec:data}), translation and
rotation invariance as offered by CNNs are comparatively unfavorable
properties. It is expected, however, that the relative ordering of frequency
plot constituents given in expression \eqref{eq:radtoken} and data-adaptive
positional encodings in combination with the attention mechanism will allow the
model to attain a basic comprehension about relations between data points as
well as their spatial configuration. Yet, conditioning the prediction of camera
information on consecutive sections of radio-frequency spectra is challenging
due to the inherently low signal-to-noise ratio across vast areas of the
frequency plots. It is thus of crucial importance being able to rely on a
versatile and expressive discretization scheme of these signal representations,
as detailed in section \ref{sec:training}. Only then will it be possible, to
differentiate between the slightest of frequency saliencies and discerning even
the subtlest of reflections. This, in turn, can only be achieved by providing
the compression models of the first stage with enough flexibility for having
them assign distinctive vocabulary entries to similar yet different input
patches, an issue which was examined extensively in the former
sections. Generally, it is hoped that even minor signal parts contained in the
radar spectrum might provide useful evidence and, by finding correspondence in
camera features, help in improving the distribution predictions of subsequent
patches. The same datasets the results of stage one were achieved with are
reused in this phase with identical splits into train and validation sets,
respectively. Again, no artificial data augmentation was applied for reasons
similar to those presented in chapter \ref{sec:training}. The capacities of both
discrete autoencoders and the comparably small dataset used for their parameter
estimation imply that sampling from categorical distributions rather than taking
the modes of the encoder logits, contrasted against in section
\ref{sec:vae_res}, might add additional regularization and contribute to tackle
overfitting. This choice is likely to impose slight amounts of noise onto the
targets during likelihood estimation in equation \eqref{eq:ce} but could prove
beneficial, if understood as a form of label smoothing. According to
\cite{mueller2019when} this technique can prevent models from becoming
over-confident too early in training. In other words, constructing the
multi-modal sequence through sampling provides for soft targets and slightly
fuzzy model input in the employed cross-entropy formulation. Also, it
pseudo-increases the dataset size by providing added variability of the data
distributions. Absolute prerequisite for this strategy are well-adapted
compression models which, if sampled from, exhibit sufficient precision so as
not to defy the purpose of supplying a ground truth in the first
place. Predicting tokens along the entire sequence instead of classifying only
the camera-related subpart promotes the models generalization capabilities. In
fact, including the error made by forecasting radar tokens in the overall loss
formulation keeps the network from focusing too eagerly on the camera token
prediction. This auxiliary loss contribution also encourages the exact
replication of the preceding RF section and thus has direct influence on the
radar conditioning. Its classification error is scaled down by a factor of
$\alpha = \num{1/8}$ though to put more emphasis on the camera-related prediction
accuracy. To reduce the risk of overfitting further, during training $10\%$ of
the input camera tokens are replaced by random integers drawn from the
corresponding sample space. This likewise causes the model not to rely on
learned internal correlations too strongly, but entices it to acquire more
general and robust features instead. All trainings were performed on GeForce RTX
2080 TI units with \SI{\sim12}{\giga\byte} of RAM for compression models with
dictionary sizes of $K=64$, $K=256$ and $K=1024$ for both modalities to retain
the chance for comparison. A batch size of \num{16} was chosen and gradients
were accumulated over \num{4} iterations for improved objective estimates with an
effective batch size of \num{64}. Decoupled weight decay
\cite{loshchilov2018decoupled} of $\lambda = \num{1e-2}$ was used for the Adam solver
to adapt the transformer parameters $\bm \psi$ for the classification task while
simultaneously constraining their magnitude. The learning rate was set to
\num{3e-5} and halved after every \num{10} validation epochs in which no loss
reduction could be observed.

\subsubsection{Results and Discussion of Multi-Modal Predictions}
\begin{figure}[h!]
  \centering
  \includegraphics[width=0.48\textwidth,trim=0 0 0 0]{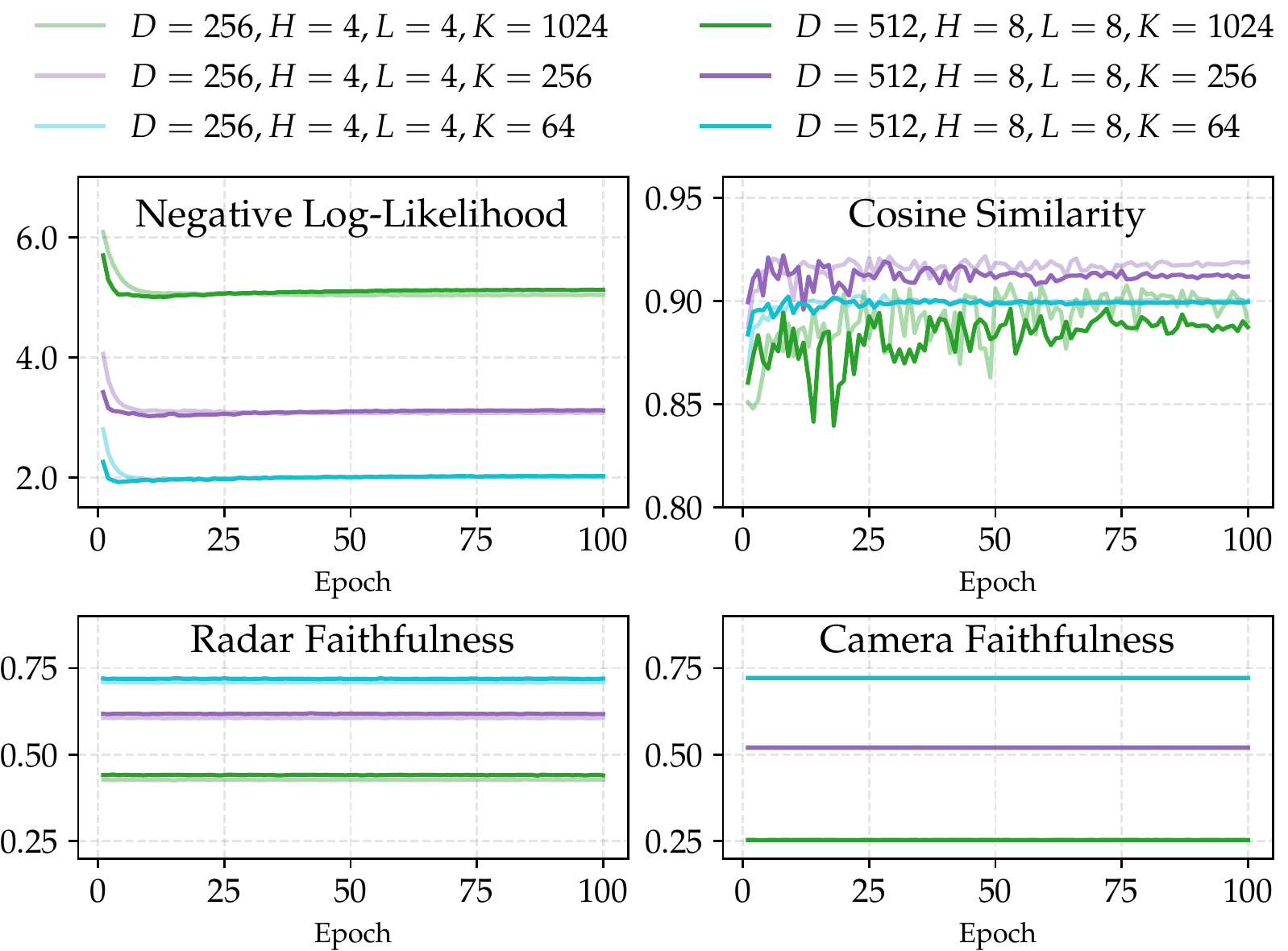}
  \caption{Performance of the autoregressive sequence modeling in validation.}
  \label{fig:pred_curves}
\end{figure}

\begin{figure*}[h!]
  \centering \subfloat[$16\times 16$ pixel discretization of the RF
  plot]{\includegraphics[width=0.65\columnwidth]{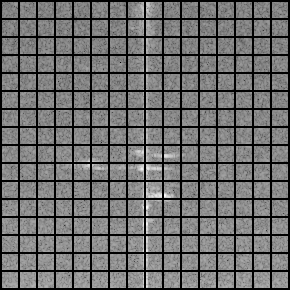}} \hfil
  \subfloat[$16\times 16$ pixel discretization of the RGB
  image]{\includegraphics[width=0.65\columnwidth]{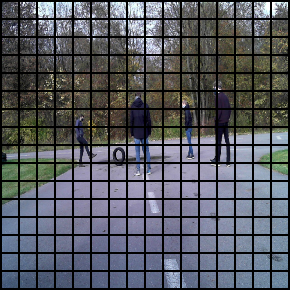}}
  \caption{Approximate discretization regions each token represents upon
    stochastic decomposition. In reality, subsampling in combination with
    zero-padding of the boundaries entails slight overlapping between the
    individual patches. The larger the downsampling (a factor of $16$ was chosen
    in this work), the smaller the memory requirements within the transformer
    model. Yet, for stronger compression ratios, the discretization grid would
    become less refined and individual integers had to represent larger amounts
    of continuous sensor data, generally compromising information preservation.}
  \label{fig:att_disc}
\end{figure*}

\begin{figure*}[h!]
  \centering \subfloat[$16\times 16$ pixel discretization of the RF
  plot]{\includegraphics[width=0.65\columnwidth]{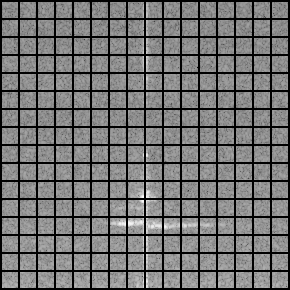}} \hfil
  \subfloat[$16\times 16$ pixel discretization of the RGB
  images]{\includegraphics[width=0.65\columnwidth]{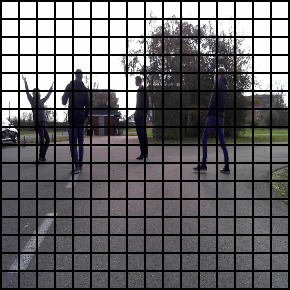}}
  \caption{Approximate discretization regions each token represents upon
    stochastic decomposition. In reality, subsampling in combination with
    zero-padding of the boundaries entails slight overlapping between the
    individual patches. The larger the downsampling (a factor of $16$ was chosen
    in this work), the smaller the memory requirements within the transformer
    model. Yet, for stronger compression ratios, the discretization grid would
    become less refined and individual integers had to represent larger amounts
    of continuous sensor data, generally compromising information preservation.}
  \label{fig:att_disc_deep}
\end{figure*}

Table \ref{tab:trafo_ablation} shows the results of a corresponding study
ablating several important parameters of the transformer network structure
introduced before. The lowest negative log-likelihood loss (NLL) is reported as
central reference metric for the data-fitting progress alongside the epoch on
the validation set in which it was recorded. Obviously, models forced to choose
among $K=1024$ possibilities for the prediction of the next token, display
higher classification errors than those relying on a dictionary size of $K=64$
per modality. Then again, a larger selection range may be beneficial for
assembling more realistic images, allowing the model to include finer details
and differentiate content more clearly as was demonstrated in section
\ref{sec:vae_res}. Additionally, the FID score (cf. equation \eqref{eq:fid}) was
calculated by using the modes of the predicted camera-related distributions over
the respective vocabulary to select reconstruction features. These continuous
latents were then decoded into image space to fit multivariate Gaussians to the
validation dataset as explained in section \ref{sec:vae_res}.
\begin{table}[ht]
  \caption{Ablation of classification error and FID over transformer parameters.}
  \renewrobustcmd{\bfseries}{\fontseries{b}\selectfont}
  \sisetup{detect-weight,mode=text,group-minimum-digits = 4}
  \centering
  \begin{tabular}{l c c c c c c c}
    \toprule
    $K$ & $D$ & $H$  & $L$   & NLL $\downarrow$&  Cos-Sim $\uparrow$& FID $\downarrow$& Epoch\\
    \midrule
    \multirow{2}{*}{64}
        &256 & 4 & 4 & 1.94&0.899&161.66&11\\
        &512 & 8 & 8 & 1.92&0.902&157.71&4\\
    \midrule
    \multirow{2}{*}{256}
        &256 & 4 & 4 & 3.05&\bfseries 0.923&126.38&37\\
        &512 & 8 & 8 & 3.02&0.922& 116.76&10\\
    \midrule
    \multirow{2}{*}{1024}
        &256 & 4& 4 & 5.02&0.900&125.10&51\\
        &512 & 8 & 8 & 5.00&0.894&\bfseries 116.04&11\\
    \bottomrule
  \end{tabular}
  \label{tab:trafo_ablation}
\end{table}
Figure \ref{fig:pred_curves} illustrates the models evolvement across epochs on
the validation set for the above parameter choices. Inflection points in the
classification loss are a sign of overfitting, indicating the point at which a
model has reached its maximum generalization potential for a given
configuration. Further, the plot shows the average cosine similarity between
sampled camera token input and modes of the predicted PMFs. This quantity serves
as a measure of a models confidence and expresses its ability to adjust its
$N=256$ camera-related distributions during training. The extent to which the
categorical camera latents actually deviate from the mode of their posterior
probability mass function (cf. equation \eqref{eq:hardsampled}) can be
quantified by examining the element-wise compliance.
This is reported as faithfulness in the given Figure for both domains and
averaged per epoch over the entire validation set. Even the models with modest
dictionary sizes of $K=64$ sample the largest-probability category only about
\num{3} of \num{4} times, with minor differences between modalities. This does
not necessarily harm the overall density estimation and camera sequence
prediction goal since other categories might be almost equally suitable
candidates, exhibiting probabilities similar to the modes of the
distributions. One possible reason is the acquisition of several nearly
identical dictionary features during the compression training in stage one and
the advanced interplay between learning vocabulary and adapting generator.  In
conclusion, the performance of the autoregressive predictor with highest cosine
similarity in table \ref{tab:trafo_ablation} using $K=256$ categories as well as
the model with $K=1024$ per modality and lowest FID score appear most promising
and will be used in the final evaluation part. Figure \ref{fig:attn_maps} and
Figure \ref{fig:attn_maps_deep} in the appendix highlight the inter-sequence
attention span for one exemplary sample if passed through both trained
transformers. The illustration allows to qualitatively examine in which head of
which layer camera tokens paid particular interest to radar information. Showing
only the lower-left submatrix gives clear insights into the formed cross-modal
connections within the network. The original input and their approximate
discretization boundaries are indicated alongside the corresponding attention
maps in Figure \ref{fig:att_disc} and Figure \ref{fig:att_disc_deep} for visual
reference.

\subsection{Conditional Synthesis of Camera Symbols}
\label{sec:synthesis}
\begin{figure}[h!]
  \centering
  \includegraphics[width=0.5\textwidth,height=0.81\textheight,keepaspectratio]{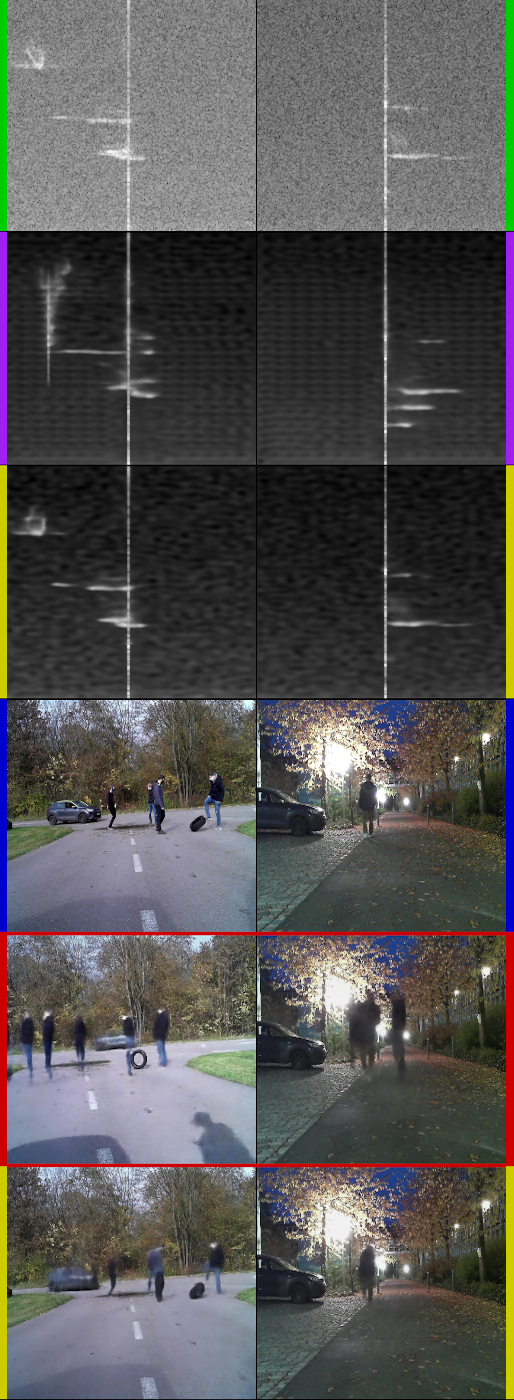}
  \caption{Range-Doppler maps \textcolor{darkgreen}{(green)} contain information
    unaffected by lighting conditions, shadow casting or occlusion
    phenomena. The probabilistic generations of the camera view
    \textcolor{red}{(red)} can therefore rely on a robust and stable perceptual
    prior to not miss out on vital scene elements. For visual reference, the
    ground truth camera image \textcolor{blue}{(blue)} as well as the stochastic
    reconstructions of both discretized domains \textcolor{darkyellow}{(yellow)}
    are given. The autoregressivley predicted radar reconstruction
    \textcolor{darkpurple}{(purple)} allows a visual inspection of the causal
    conditioning within the transformer network. The synthesized views convey
    clear impressions of the environment, even if the number of objects or their
    exact position and orientation is slightly off at times.}
  \label{fig:mid_example}
\end{figure}

\begin{figure*}[h]
  \centering
  \includegraphics[width=0.63\textheight]{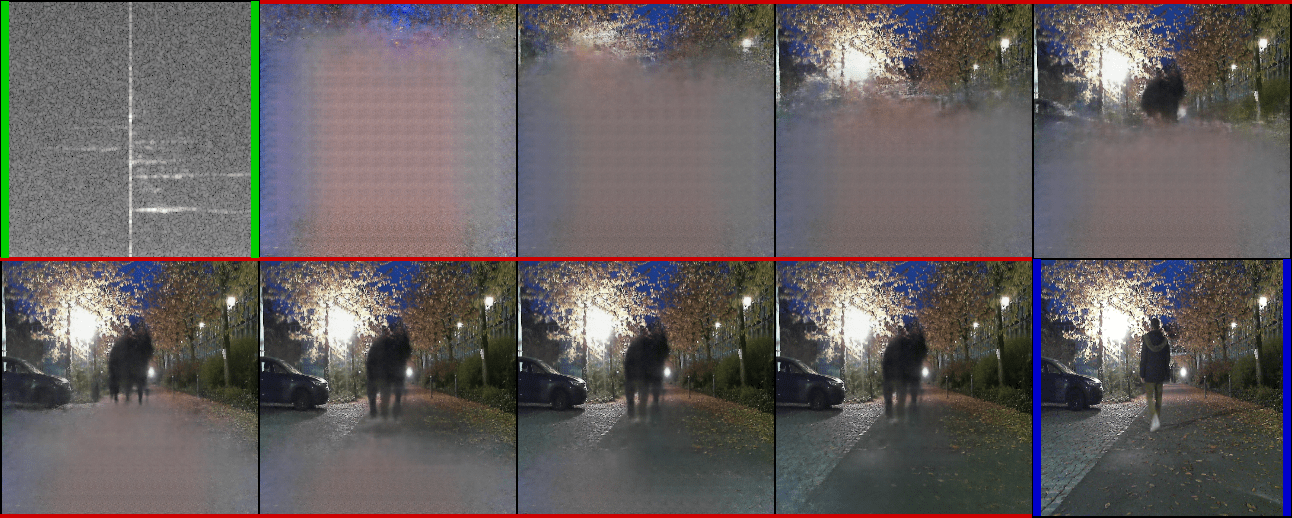}
  \caption{\textcolor{red}{Prediction process}, \textcolor{green}{RF
      conditioning} and \textcolor{blue}{camera GT}: The parked car reflects
    strongly on the vertical rD line, and the model includes it in its generated
    view. Multiple additional scatterings at various distances causes the
    network to become insecure about the precise number of dynamic reflectors in
    the scene. Both compression models use $K=256$.}
  \label{fig:stepwise1}
\end{figure*}

\begin{figure*}[h]
  \centering
  \includegraphics[width=0.63\textheight]{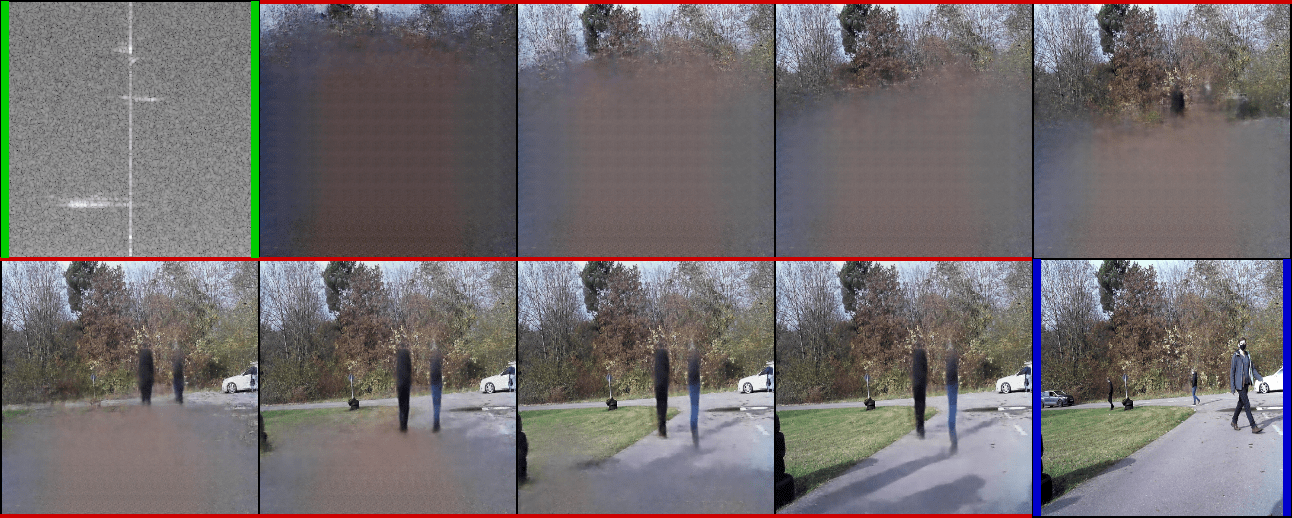}
  \caption{\textcolor{red}{Prediction process}, \textcolor{green}{RF
      conditioning} and \textcolor{blue}{camera GT}: The model synthesizes two
    abstract VRU representations in the foreground but misses the two persons in
    the background. The white parked car shows as a strong reflection on the
    vertical center line of the rD plot and is included in the
    generated view. The black car is beyond the maximum range of the radar so
    that the algorithms fails to establish a cross-modal correspondence,
    precluding it from the final image. Both compression models use $K=256$.}
  \label{fig:stepwise2}
\end{figure*}

\begin{figure*}[h]
  \centering
  \includegraphics[width=0.63\textheight]{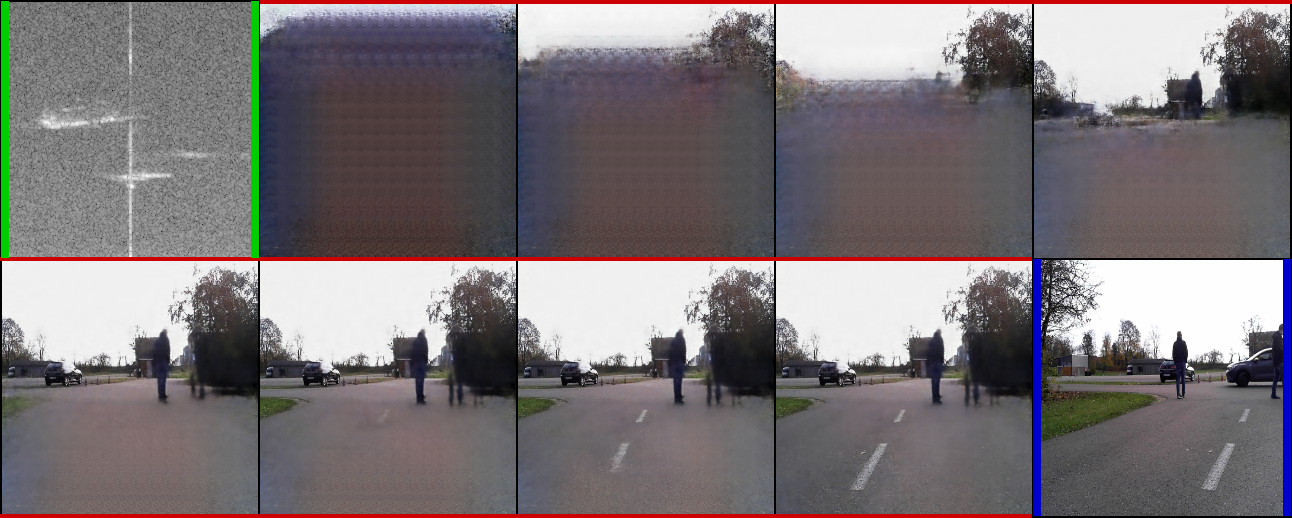}
  \caption{\textcolor{red}{Prediction process}, \textcolor{green}{RF
      conditioning} and \textcolor{blue}{camera GT}: Dynamic objects like the
    two VRU and the moving vehicle can be discriminated via their relative
    radial velocities in the rD map. With no angle information included in the
    signal though, the model is unsure about their lateral position and general
    orientation. It generates trees and about one and a half persons at rather
    random positions. Both compression models use $K=256$.}
  \label{fig:stepwise3}
\end{figure*}

\begin{figure*}[h]
  \centering
  \includegraphics[width=0.63\textheight]{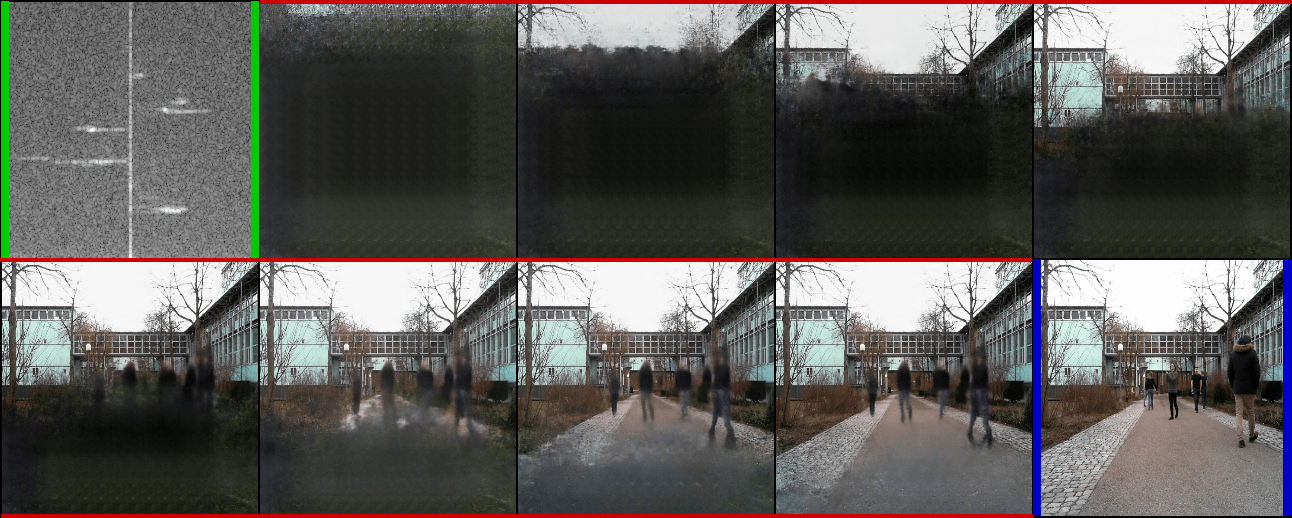}
  \caption{\textcolor{red}{Prediction process}, \textcolor{green}{RF
      conditioning} and \textcolor{blue}{camera GT}: This highly dynamic scene
    is well resolved via unique Doppler frequencies and allows the model to
    almost accurately predict the number of VRU in the scene through abstract
    creations. Their distance to the sensor is featured almost exactly and
    allows for an immediate assessment of the environment. Both compression
    models use $K=1024$.}
  \label{fig:stepwise4}
\end{figure*}

\begin{figure*}[h]
  \centering
  \includegraphics[width=0.63\textheight]{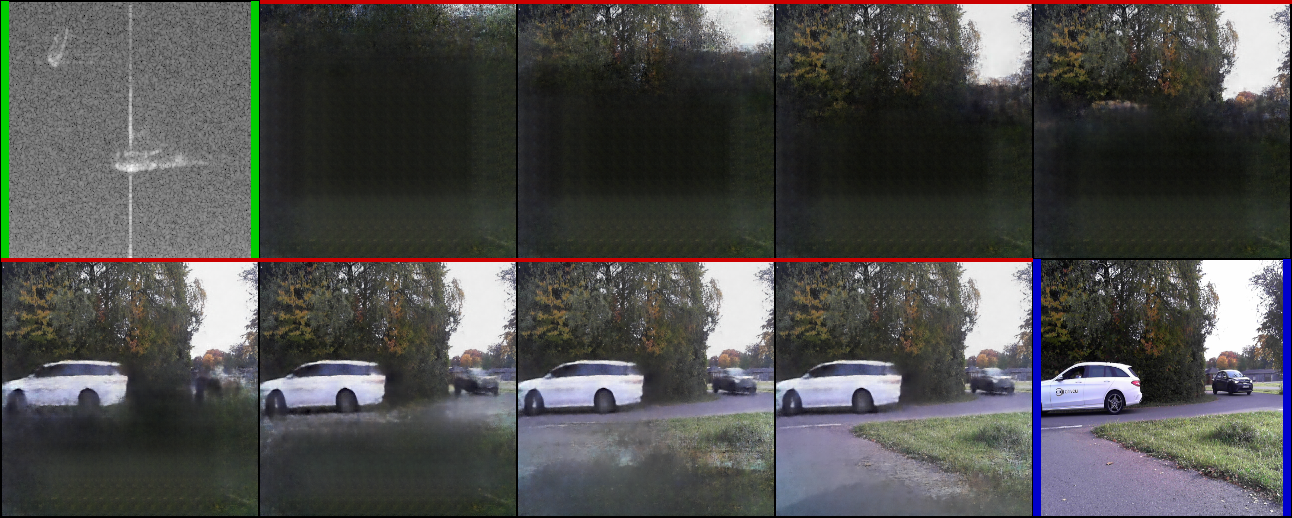}
  \caption{\textcolor{red}{Prediction process}, \textcolor{green}{RF
      conditioning} and \textcolor{blue}{camera GT}: Both vehicles are clearly
    recognizable through their rD signatures by revealing multiple reflection
    centers and are therefore included in the created camera view with large
    probability, albeit with deviant angular precision. Both compression models
    use $K=1024$.}
  \label{fig:stepwise5}
\end{figure*}

\begin{figure*}[h]
  \centering
  \includegraphics[width=0.63\textheight]{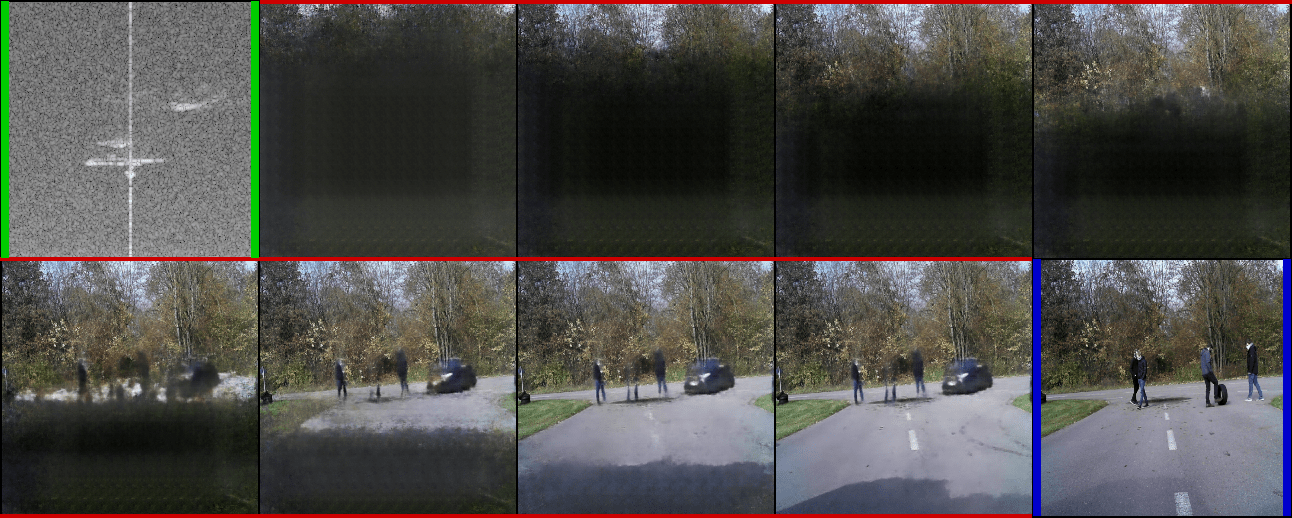}
  \caption{\textcolor{red}{Prediction process}, \textcolor{green}{RF
      conditioning} and \textcolor{blue}{camera GT}: Being out of the cameras
    horizontal field of view, the retracting vehicle is originally not captured
    by the camera sensor (lower-right image). Aside from multiple VRU-caused
    echoes, the networks attention is drawn to the car's reflection in the radar
    modality (right half of the rD map). It therefore correctly includes a
    corresponding car representation in the camera view. Both compression
    models use $K=1024$.}
  \label{fig:stepwise6}
\end{figure*}

Having successfully adapted the transformer weights and biases to the training
sequences, the ultimate proof of concept yet remains. Supplied with nothing but
radio frequencies, is the model able to construct a high-fidelity view of the
sensed environment and capture the essential details and important aspects of
the surroundings? If so, this would open up unprecedented possibilities for the
control of autonomous systems in disadvantageous circumstances and significantly
enhance the safety of related applications. To investigate the stochastic
composition of radar-conditioned camera images, rD maps of the validation
dataset are tokenized as given by equation \eqref{eq:radtoken} and serve as
input to the autoregressive predictor without any camera information to rely
on. As per defined procedure described in section \ref{sec:trafo_training} and
according to the incorporated causal attention, the networks very last output
yields a distribution over all possible camera constituents. The associated PMF
is conditioned exclusively on microwave-sensed information and discrete sampling
results in the first camera token, corresponding to the content of the upper
left image region. This autoregressive decision process is influenced by all
radar latents fully attended to in all transformer layers. After appending the
found camera index to the initial RF part, the sequence is input into the
transformer again so that the next camera token is depending not only on radar
features but also the formerly predicted camera component and its
high-dimensional representation within the model. This iterative process is
repeated, obtaining one symbol at a time, until all $256$ camera indices have
been determined. Drawing on an ever-growing amount of prior information, makes
the model's decisions increasingly confident but also strongly dependent on the
quality of previous predictions. In fact, having by chance, sampled a suboptimal
initial camera token can possibly cause the entire subsequent synthesis to
deteriorate.
\begin{figure}[h!]
  \centering
  \includegraphics[width=0.48\textwidth,trim=0 0 0 0]{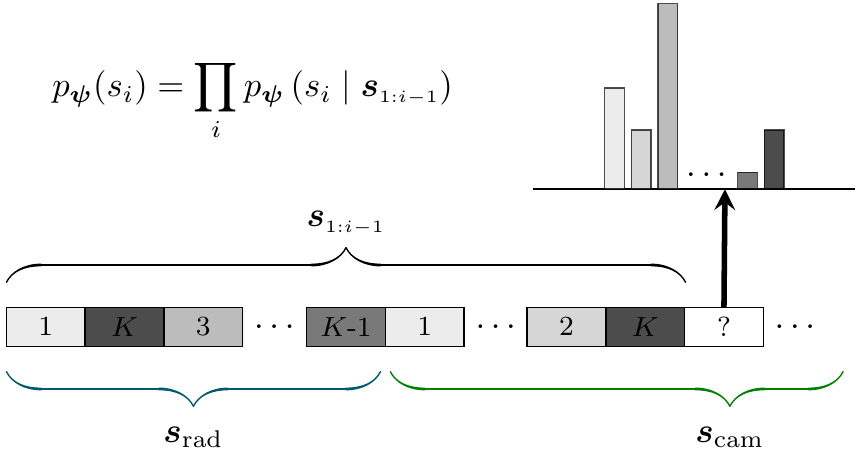}
  \caption{Autoregressive prediction of camera content based on discretized
    radar information and successively generated camera components.}
  \label{fig:prediction}
\end{figure}
This is an important consequence of choosing a unified probabilistic treatment
for this multi-modal fusion approach and stands in stark contrast to the
previously outlined training procedure. During parameter fitting, the density
estimation for every sequence element was only based on ground truth samples
rather than successively predicted symbols. Now, half of the sequence is
appended dynamically during runtime, adding another element of uncertainty to
the final scene construction aside from the stochastic radar conditioning
through encoder-induced categoricals. For better intuition, the described
generation procedure is depicted in Figure \ref{fig:prediction} for a single
sample stream and some camera tokens already predicted. Upon sequence
completion, the detached camera section selects the respective camera-specific
dictionary entries before those are reshaped and eventually decoded by the
camera generator into the desired view. Given its stochastic nature, the results
accomplished by the presented method are better investigated by visual
inspection rather than by quantitative measures. Figure \ref{fig:mid_example}
exemplarily shows two randomly inferred probabilistic views of the environment
based on prior RF information. For reference, the actual camera image is
displayed as well as the straight-forward reconstructions via categorical
samples of both domains as introduced in section \ref{sec:vae_res} for the
discretization task. The replicated rD map provides insight into the radar
conditioning the model can build upon for predicting the view. The camera
reconstruction should serve as an upper bound for the visual quality, achievable
by the algorithm. Finally, the illustration also displays the frequency plot
recovery based on the modes of the autoregressive forecasting of the radar
subsequence. Even though, these have no influence on the actual synthesis as
only the original discretized RF sequence is used for conditioning, this image
allows to obtain a notion of the models inner workings resulting in the final
context prediction. The model generally succeeds in reproducing the global
structure of the surroundings and for the most part manages to compile a
realistic rendering of its central components. Occasionally, regions show severe
artifacts or distortions of objects within the scene. Backgrounds are usually
reflected accurately and sharp, presumably due to its limited variation within
the dataset. Also, the first camera symbol, if sampled carefully based on
synchronous RF input, seems to be a strong informant with major influence on the
following visual assembly, determining the overall composition
quality. Additional results substantiating the validity of the approach can be
found in Figure \ref{fig:trafo1} to Figure \ref{fig:trafo5} in the appendix
alongside the attempt to justify the specific conclusions made by the models in
both visual and written form. Additionally, Figure \ref{fig:stepwise1} to Figure
\ref{fig:stepwise6} highlight the actual prediction process and explain possible
links between RF conditioning and generated outcome. It is important to note
that the network is not provided any temporal context which would significantly
simplify the forecasting task and make the prediction more accurate.
As no resampling is applied to correct for vastly misclassified first camera
tokens, the importance of drawing high-quality samples cannot be
overstated. Introducing a temperature parameter into the softmax normalization
of the raw transformer logits akin to equation \eqref{eq:decline} and truncating
the tails of the PMF by top-k selection, enables so-called \textit{nucleus
  sampling} \cite{holtzman2019curious} of the camera tokens. More precisely,
shrinking the sample space to only a few categories $\hat{K}\ll K$ comprising the
bulk of the probability mass increases both the sample quality and reliability
by preventing low-probability outcomes. Then, for larger temperatures $\tau > 1$
the truncated densities become more uniform, which promotes sample
diversity. Smaller values $\tau < 1$ on the other hand, further enhance large
probabilities and thus strengthen sample coherence. This claim can be
investigated by consulting Figure \ref{fig:roundabout_joined} to Figure
\ref{fig:winter_joined} in the appendix. Additional illustrative material and
videos showcasing video sequences of various reconstructed outdoor scenes can be
found at
\href{https://cditzel.github.io/GenRadar/}{cditzel.github.io/GenRadar}. Clearly,
the models are capable of synthesizing intuitive camera views albeit with
varying degree of realism and credibility to them. Often times it is difficult
to explain why one particular image has a more natural look and feel to it than
another. At other times, the defects are more evident or the model is just
completely off with its predictions. Unsurprisingly, typical failure modes are
horizontally-flipped generations with respect to the camera ground truth. As
mentioned in section \ref{sec:radar}, the rD maps lack any azimuthal information
which would better help the network to distinguish left from right. The
important aspect though, as mentioned in the beginning of this paper, is the
recovery of essential information and the visual identification of central
elements for the decision-making of autonomous systems. The designed method
regularly succeeds in reflecting on the integral objects present in a scene and
is able to plausibly reconstruct most crucial entities located in the sensors
vicinity. For safety-critical systems, the accurate representation of human
anatomy or the exact topological order of potentially dangerous obstacles is
often less relevant than those objects mere detection and robust
localization. At times, the models compose completely new and artificial
environments not included in the training data but which it obviously deemed
most corresponding to the supplied rD frequency information. This gives a hint
towards the possible capabilities and the true potential of these generative
models in general. Some of those rather abstract defective attempts are
illustrated in Figure \ref{fig:fail_gen} in the appendix.

\section{Conclusion and Limitations}
Autonomous systems need to be able to rely on a fail-safe and instructive
environment perception, independent of their current surrounding
conditions. Camera images are prone to all sorts of weather-induced failings and
therefore have to be supplemented with complementary robust measuring principles
like radar to ensure a stable inflow of information. Yet, range-Doppler
frequencies usually lack the intuitive representation cameras provide. It is
thus worth attempting to combine the advantages of sensors and aim for an
enhanced visual impression of their proximity. As a possible solution, this
paper demonstrated the design and applicability of a self-learning model,
capable of generating insightful camera views conditioned on millimeter-wave
sensing. Circumventing the need for explicit data-annotations, the system
combines the strengths of convolutional and transformer architectures to exploit
a maximum of multi-modal information and to make predictions about the sensors
vicinity. First, imposing a stochastic bottleneck onto the convolutional input
restoration task led to the effective and discrete compression of the
high-dimensional measurements in both domains. Second, the resulting
memory-efficient symbol representation was assessed by an autoregressive
sequence model, establishing cross-modal dependencies through its attention
mechanism. Multiple qualitative and quantitative results proved, that the system
eventually learns the multi-sensory composition of captured surroundings. This
ultimately allows for the visual reconstruction of environments by predicting
radar-conditioned distributions over camera components. Validity and
expressiveness of the results as well as the models generalization abilities
relate directly to the amount and diversity of the available data. Consequently,
in order to create even more generic and realistic predictions, not only the
datasets size but also the network's complexity would need to be enlarged by
orders of magnitude. Additionally, at the expense of even larger data volumes,
radar-based angle estimations should be employed to exclude the systemic lateral
position error from the predictions. Nonetheless, has this feasibility study
revealed the future potential and far-reaching possibilities of self-supervised
systems if applied to low-level multi-modal sensor data. Particular value is
attributed to the complementary nature of the employed sensor types, providing
unique and characteristic features for the correspondence learning task in the
second stage. Although the effectiveness of convolutional operations for natural
images is indisputable, the same might not hold true for range-Doppler maps. It
fact, it is arguable if the employed frequency plots are even the most suitable
representation for the purpose of neural network processing. Perhaps the data
could be siphoned even closer to the sensor, intercepting the IF-signal
immediately after the mixing process. Also, the probabilistic discretization of
the rD maps marks the confluence point of the entire RF spectrum. Performed
carelessly, this compression discards potentially important information,
entailing major problems for the following cross-modal correspondence learning
and the coherent combination of sensor constituents. It would therefore be
interesting to search for alternative ways of integrating the radar information
into the multi-modal fusion process.

\section{Acknowledgments}
The author would like to mention the EleutherAI community and members of the
EleutherAI discord channels for fruitful and interesting discussions along the way
of composing this paper. Additional thanks to Phil Wang (lucidrains) for his
tireless efforts of making attention-based algorithms accessible to the humble
deep learning research community.

\printbibliography


\begin{figure*}[h]
  \centering
  \includegraphics[width=0.8\textwidth]{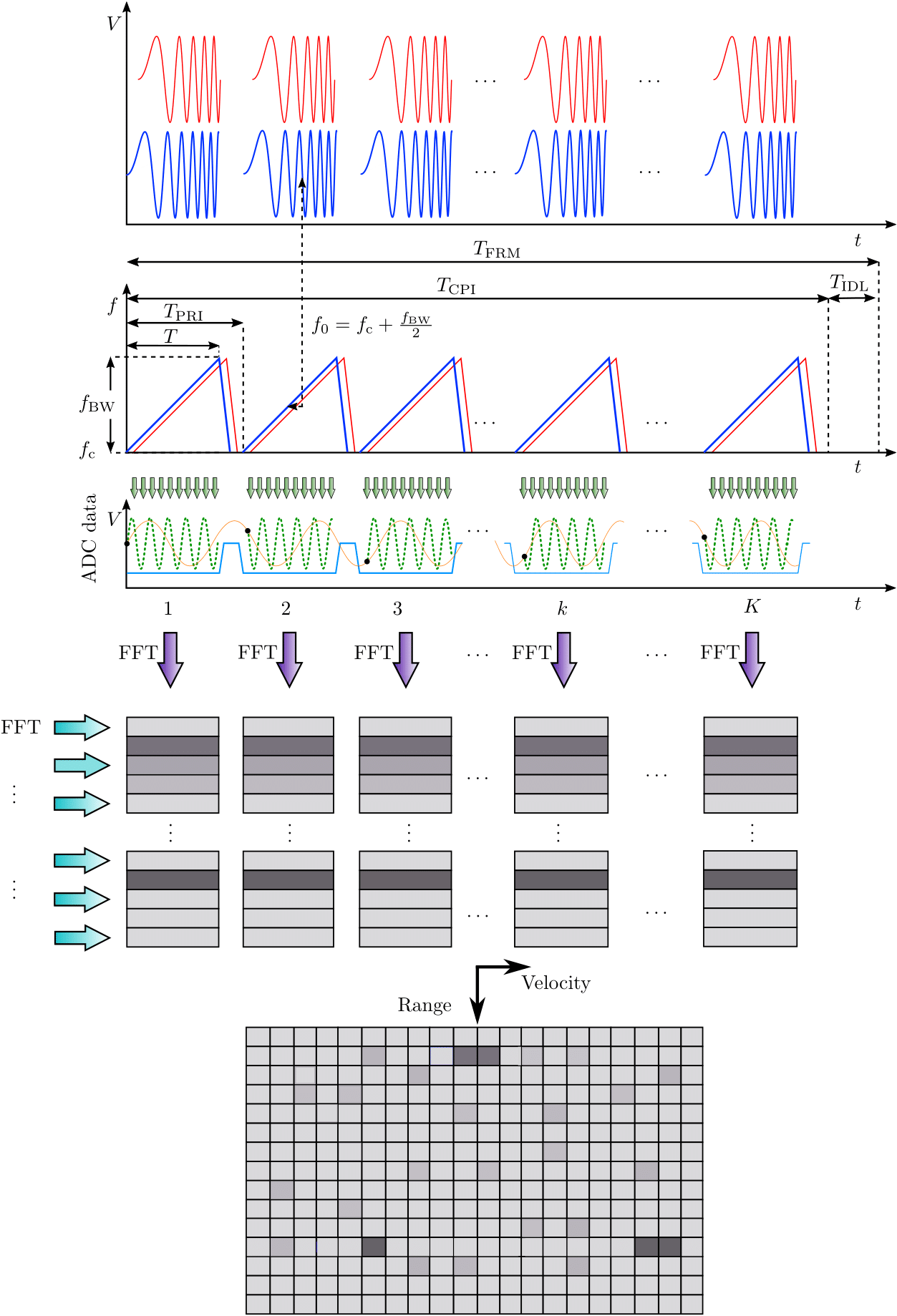}
  \caption{Two-dimensional Fourier decomposition of a differential frequency
    signal over multiple chirps. Ramp-synchronous sampling across frequency
    excursions allows for the identification of phase shifts between chirps. The
    derived Doppler information is uses to discriminate objects at equal
    distance to the sensor through the slightest of differences in relative
    velocity. Radial range and relative velocity constitute powerful
    discriminative features to complement the strengths of camera sensors.}
  \label{fig:2dfft}
\end{figure*}
\begin{figure*}
  \centering \subfloat[Progression on the validation set over epochs for camera
  models]{\includegraphics[width=0.95\columnwidth]{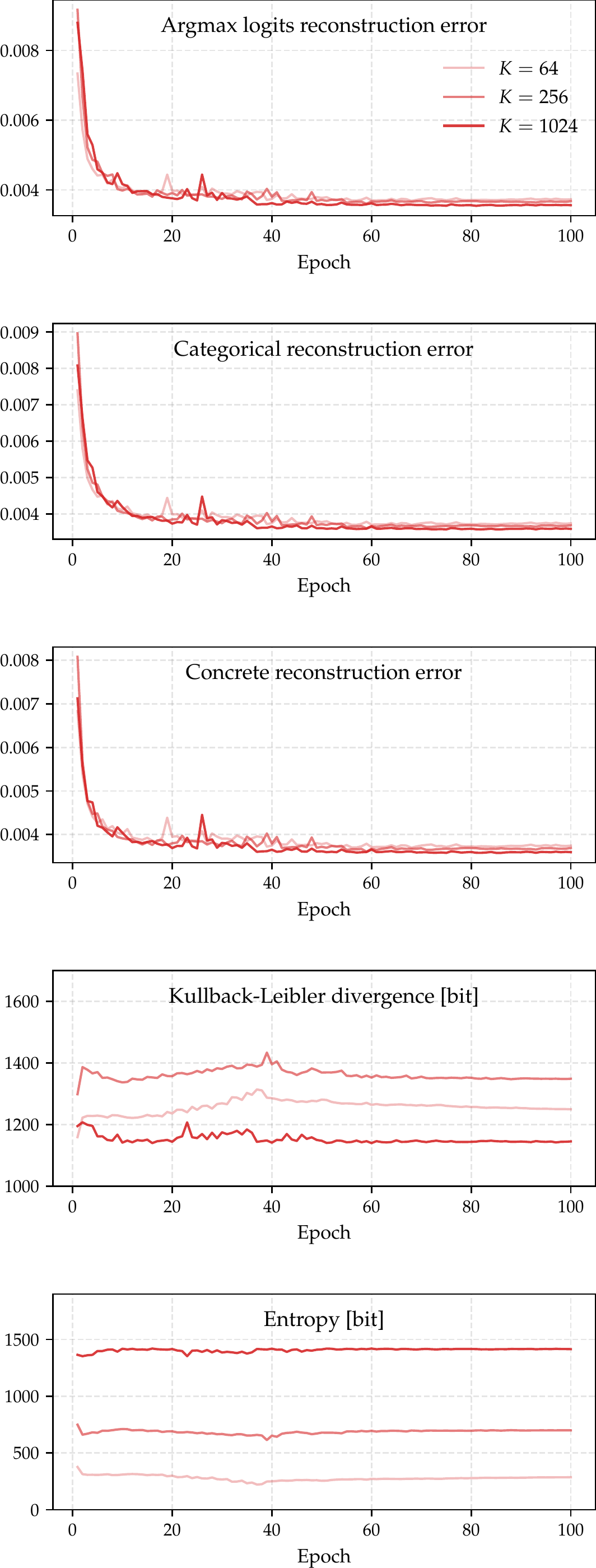}} \hfil
  \subfloat[Progression on the validation set over epochs for radar
  models]{\includegraphics[width=0.95\columnwidth]{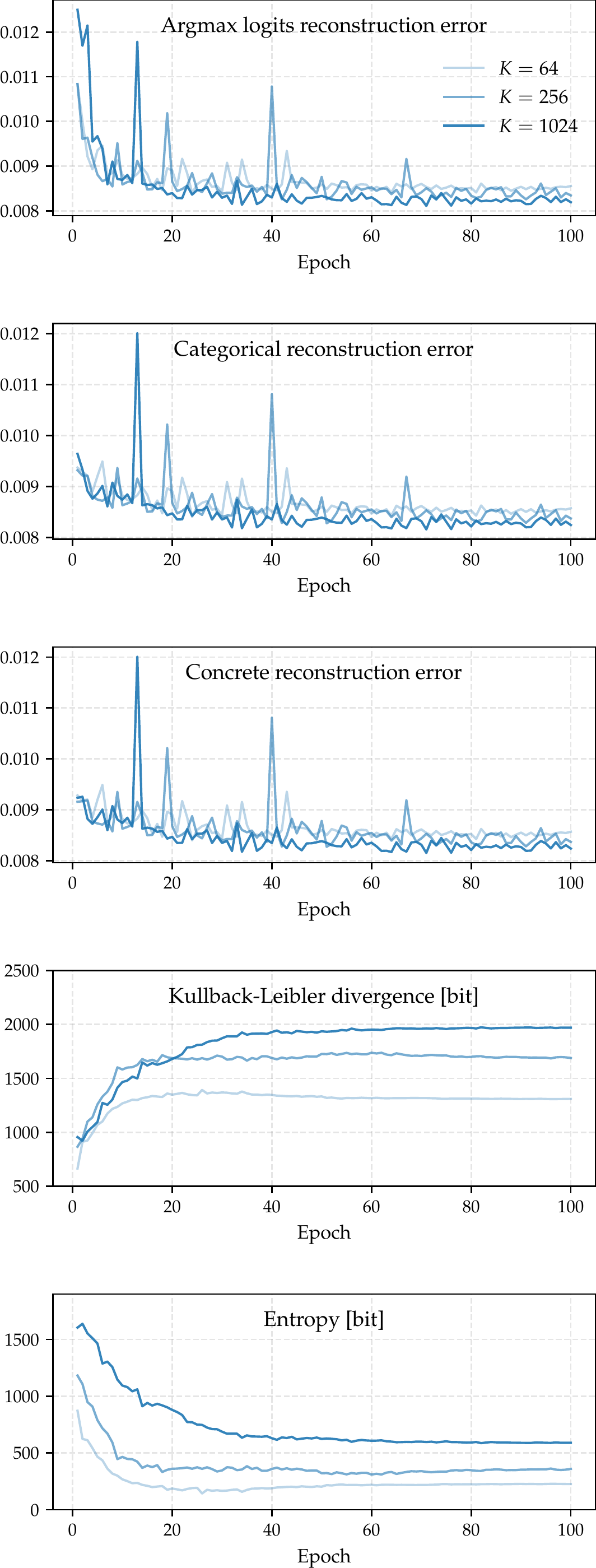}}
  \caption{Performance development of the stochastic discretization models with
    varying dictionary sizes on the validation dataset after every training
    epoch. The first three plots show similar but slightly different decreases
    in the reconstruction error, depending on the respective sampling method as
    explained in section \ref{sec:training}. The KL-term stabilizes during the
    training progress and eventually settles for an equilibrium which trades
    uniformity in vocabulary use for more customized latent spaces. The
    contradiction between maximum entropy configuration excited by the uniform
    prior and fulfillment of the reconstruction objective through an increase in
    divergence is clearly recognizable.}
  \label{fig:valid_runs}
\end{figure*}

\begin{figure*}[h]
  \centering
  \includegraphics[width=0.98\textwidth]{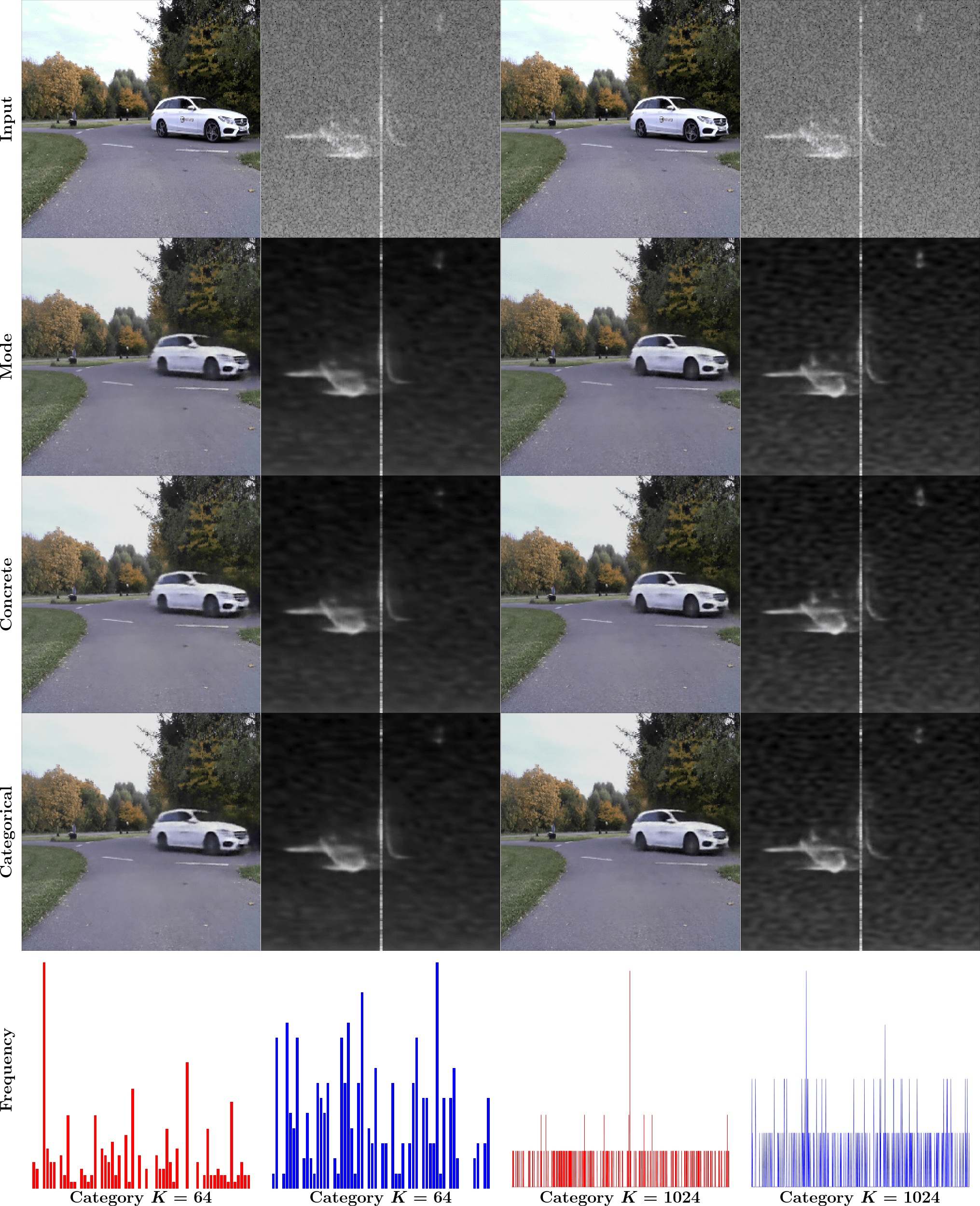}
  \caption{Camera and radar input (tow row) and probabilistic reconstructions
    for dictionary sizes of $K=64$ (left) and $K=1024$ (right). The second row
    displays results achieved by mode/argmax selection of the latent
    categories. The third row highlights the restoration of using concrete
    samples in the stochastic bottleneck and the fourth row shows results
    obtained by categorically sampling the domain constituents. The bottom row
    shows the frequency of these chosen categorical entries for the given
    input. With a larger vocabulary size, the model is able to retain more
    details by assigning separate categories to infrequent image content. The
    reconstruction quality based on categorical latent samples is comparable to
    the restoration achieved by the other two sampling methods, which visually
    proves the efficacy of the stochastic discretization approach.}
  \label{fig:roundabout}
\end{figure*}

\begin{figure*}[h]
  \centering
  \includegraphics[width=\textwidth]{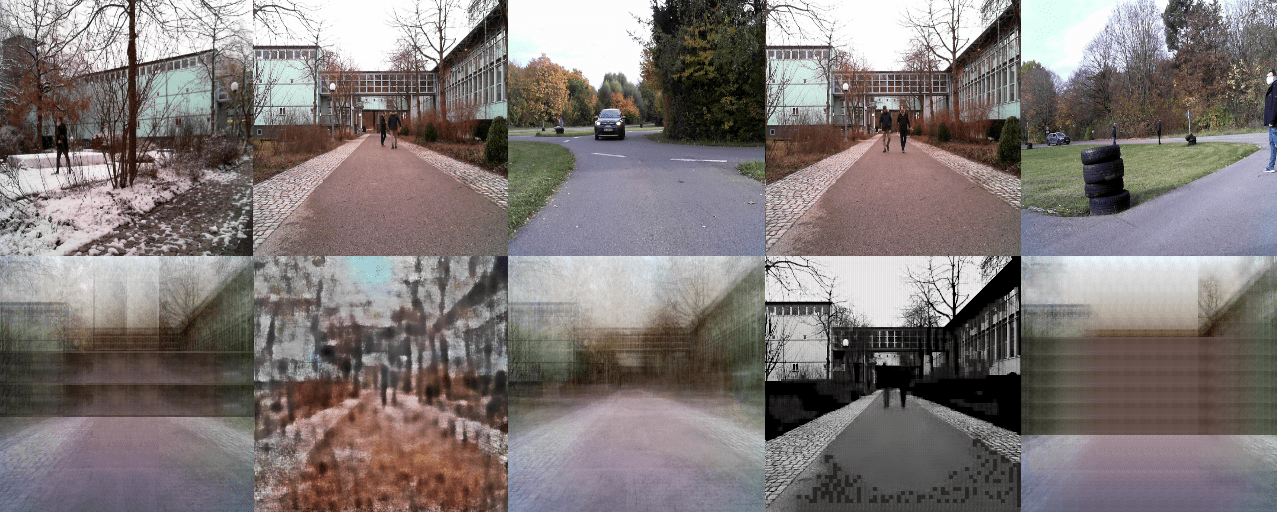}
  \caption{Selection of fail cases observed on the validation set during the
    probabilistic reconstruction training in stage one. The reasons for
    erroneous restorations are manifold, ranging from unfavorable initial weight
    constellations to imbalanced weightings between reconstruction and entropy
    loss terms in the VLB. The images clearly show the tiling subdivision of the
    input due to their regular discretization in squared patches. \textbf{Left,
      center and right image:} The model wrongly assembles the image by means of
    dictionary entries tuned to different camera samples. \textbf{Second to the
      left image:} The reconstruction exhibits severe artifacts and
    over-colorizes the output. \textbf{Second to the right image:} The model
    fails to infer color channel and instead shows checkerboard patterns, vastly
    neglecting the coverage of high-frequency details and fine-grained
    structures.}
  \label{fig:fail}
\end{figure*}

\begin{figure*}[h]
  \centering
  \includegraphics[width=\textwidth]{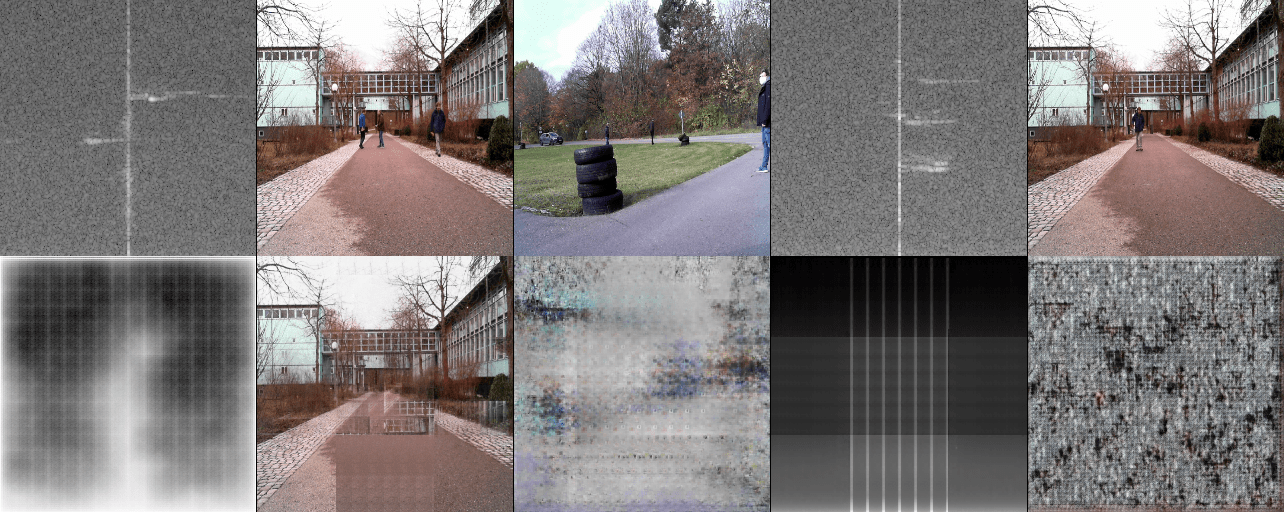}
  \caption{Typical failure modes of both modalities observed during the training
    process in stage one. The models fail to restore both local and global
    features and neglect the vast majority of image contents. \textbf{Camera
      images:} The camera data is recovered beyond recognition for overly
    exaggerated entropy weights or too rapid temperature declines. \textbf{Left
      image:} The numerical procedure die not properly converge for an initial
    Gumbel-Softmax temperature larger than one. \textbf{Second to the right
      image:} The regularization term in the loss function was given too much
    influence forcing the model to uniformly use almost all dictionary entries
    which hinders data restoration.}
  \label{fig:fail2}
\end{figure*}

\begin{figure*}[h!]
  \centering \subfloat[Depiction of selecting all $256$ latent camera categories
  separately and decoding the associated features back into measurement
  space.]{\includegraphics[width=0.9\columnwidth]{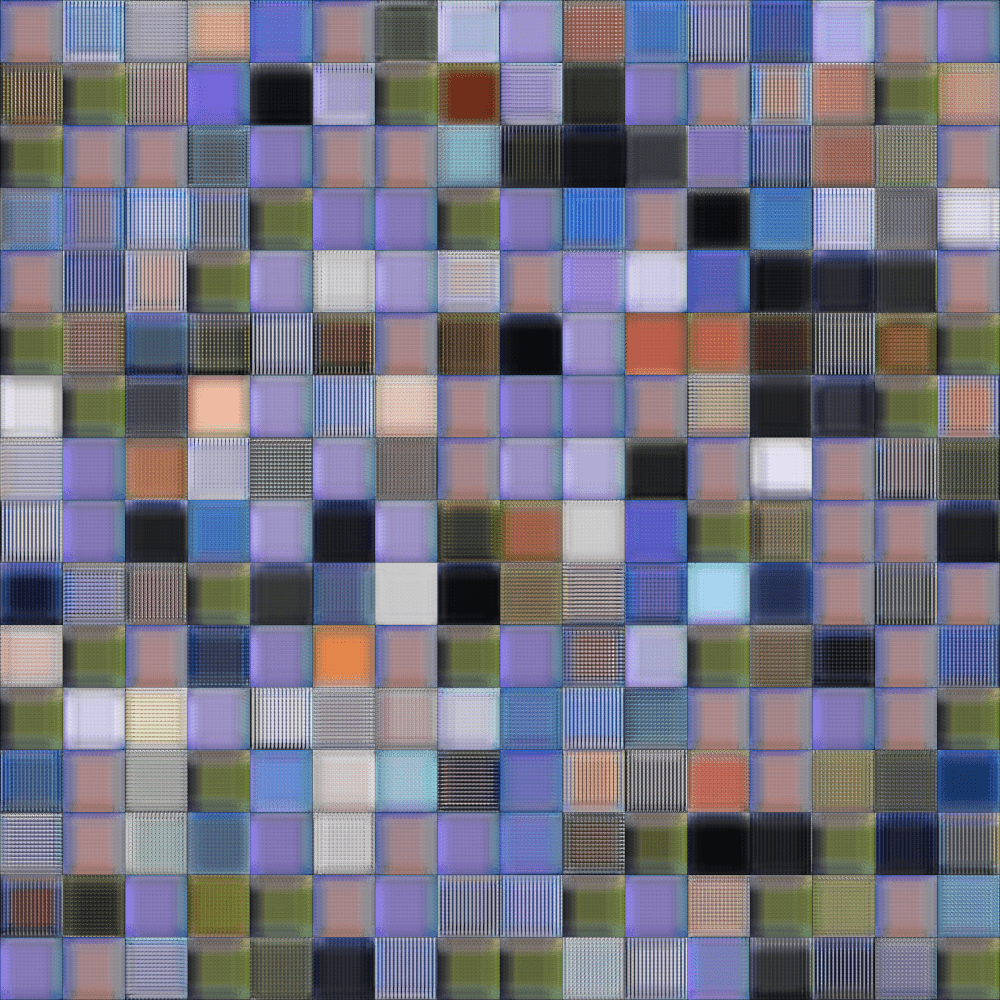}} \hfil
  \subfloat[Depiction of selecting all $256$ latent radar categories separately
  and decoding the associated embeddings back into measurement space.
  ]{\includegraphics[width=0.9\columnwidth]{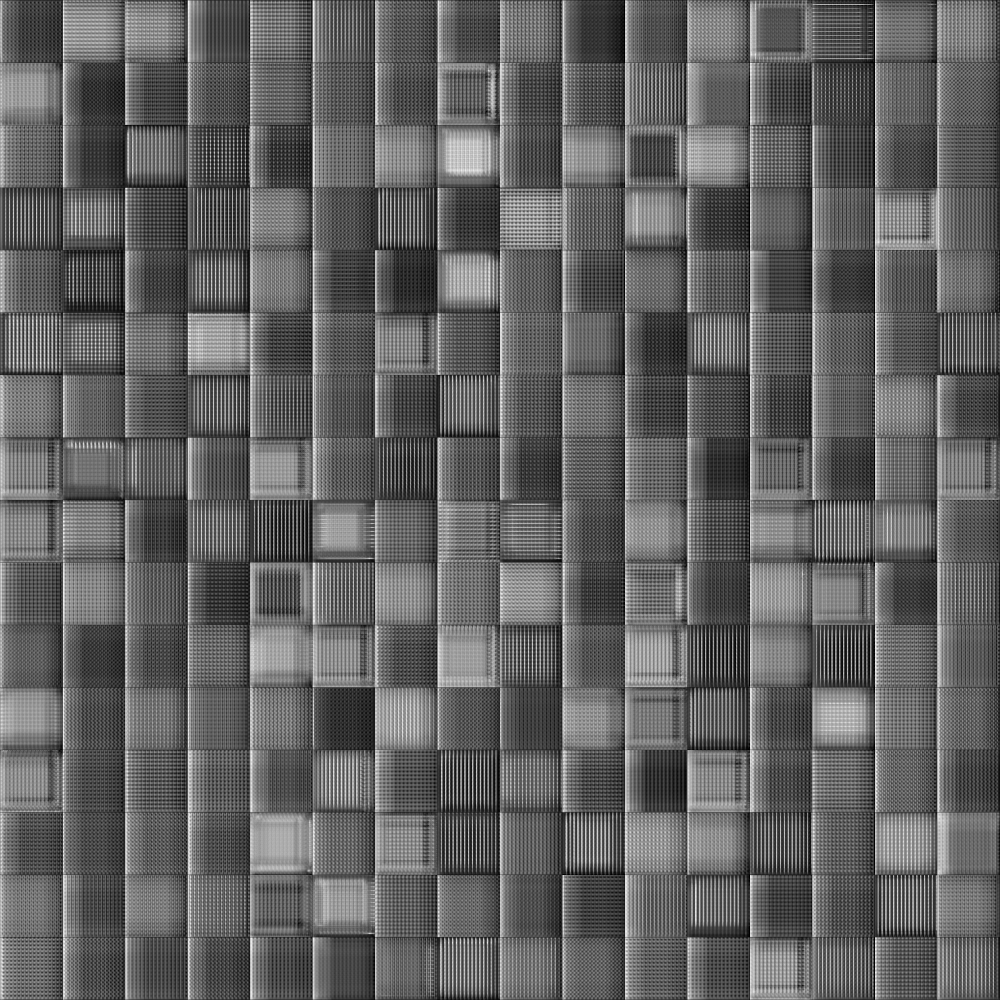}}
  \caption{Successively choosing identical indices for all latent variables and
    decoding the corresponding feature vector of the dictionary with $K=256$
    back into image space provides insights into the model's diversity, adopted
    during training. The individual patches themselves are rarely homogeneous
    due to spatially overlapping upsampling and specific boundary treatments
    within the convolutional decoders. Visual redundancies between patches could
    hint towards potential redundancies in the acquired vocabulary and might
    justify dimensionality reductions of the model.}
  \label{fig:mod_patches}
\end{figure*}

\begin{figure*}[h!]
  \centering
  \includegraphics[height=0.52\textheight]{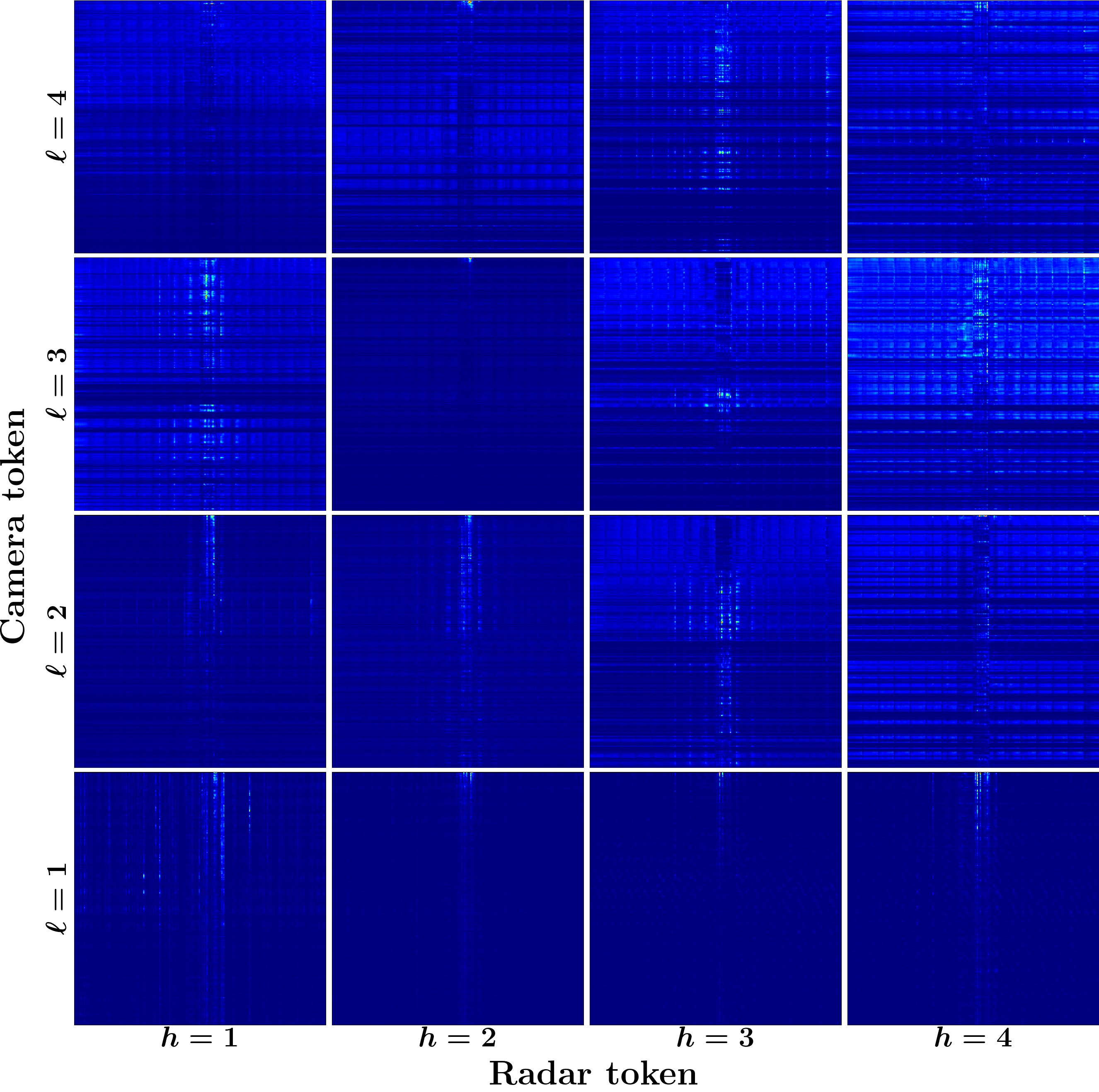}
  \caption{Attention span for the transformer with $4$ layers and $4$ heads and
    the discretized multi-modal input shown above, compressed by models with
    dictionary sizes of $K=256$. Only the lower-left submatrix (cross-modal
    attention part) is shown for improved visualization. Local maxima in every
    plot denote camera tokens attending to radar information. Often, camera
    symbols reacting most actively to the radio-frequency conditioning are
    located in the upper half of the subimages. This is to be expected, as the
    first few camera constituents are predicted almost exclusively, depending on
    the radar spectrum, particularly in the first few layers. For subsequent
    layers as the information is transmitted deeper into the network, the
    attention span is significantly broadened. Best viewed in color and zoom on
    a computer.}
  \label{fig:attn_maps}
\end{figure*}

\begin{figure*}[h!]
  \centering
  \includegraphics[height=0.52\textheight]{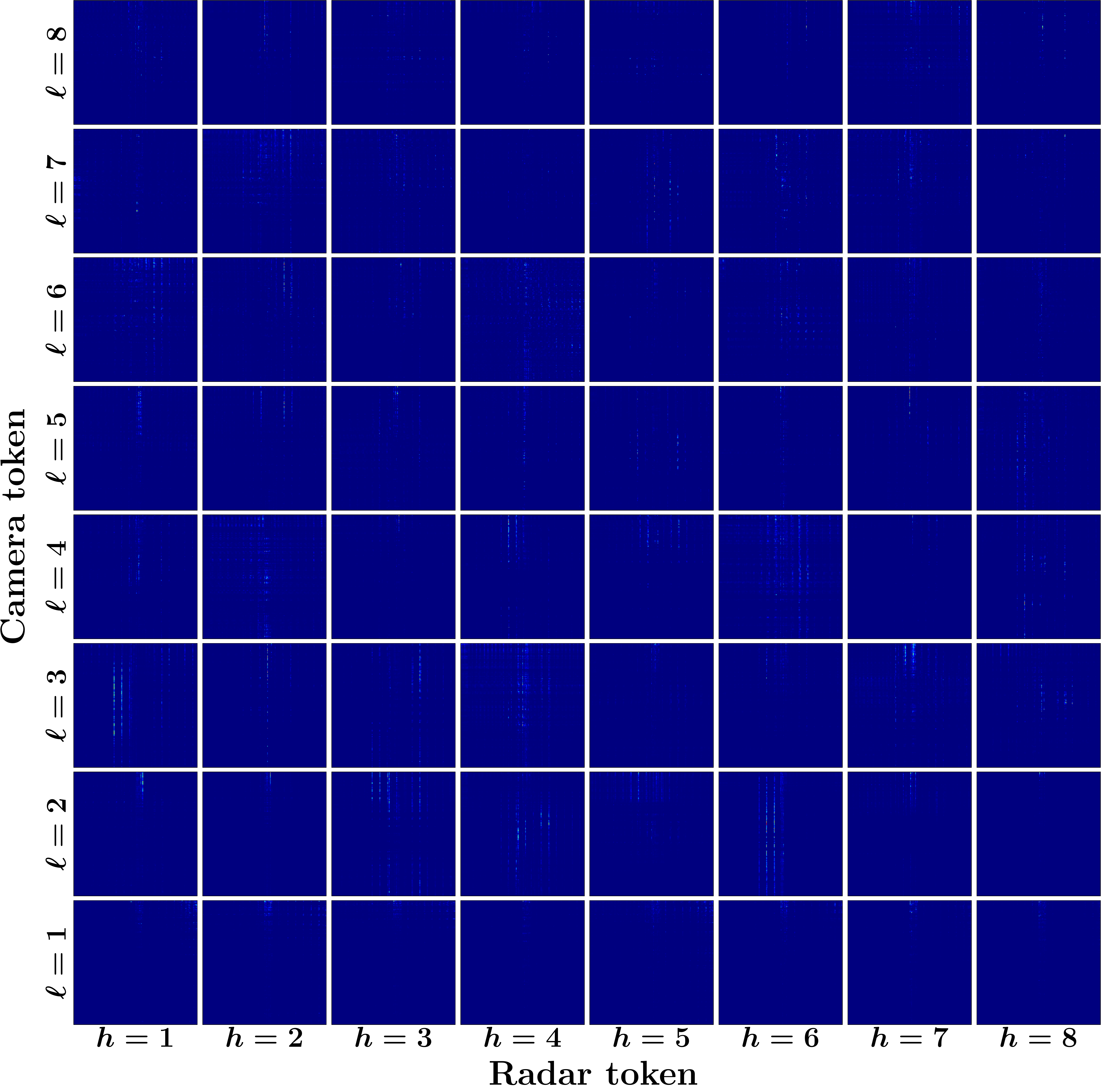}
  \caption{Attention span for the transformer with $8$ layers and $8$ heads and
    the discretized multi-modal input shown above, compressed by models with
    dictionary sizes of $K=1024$. Only the lower-left submatrix (cross-modal
    attention part) is shown for improved visualization. Local maxima in every
    plot denote camera tokens attending to radar information. Often, camera
    symbols reacting most actively to the radio-frequency conditioning are
    located in the upper half of the subimages. This is to be expected, as the
    first few camera constituents are predicted almost exclusively, depending on
    the radar spectrum, particularly in the first few layers. For subsequent
    layers as the information is transmitted deeper into the network, the
    attention span is significantly broadened. Best viewed in color and zoom on
    a computer}
  \label{fig:attn_maps_deep}

\end{figure*}

\begin{figure*}[h]
  \centering
  \includegraphics[width=\textwidth]{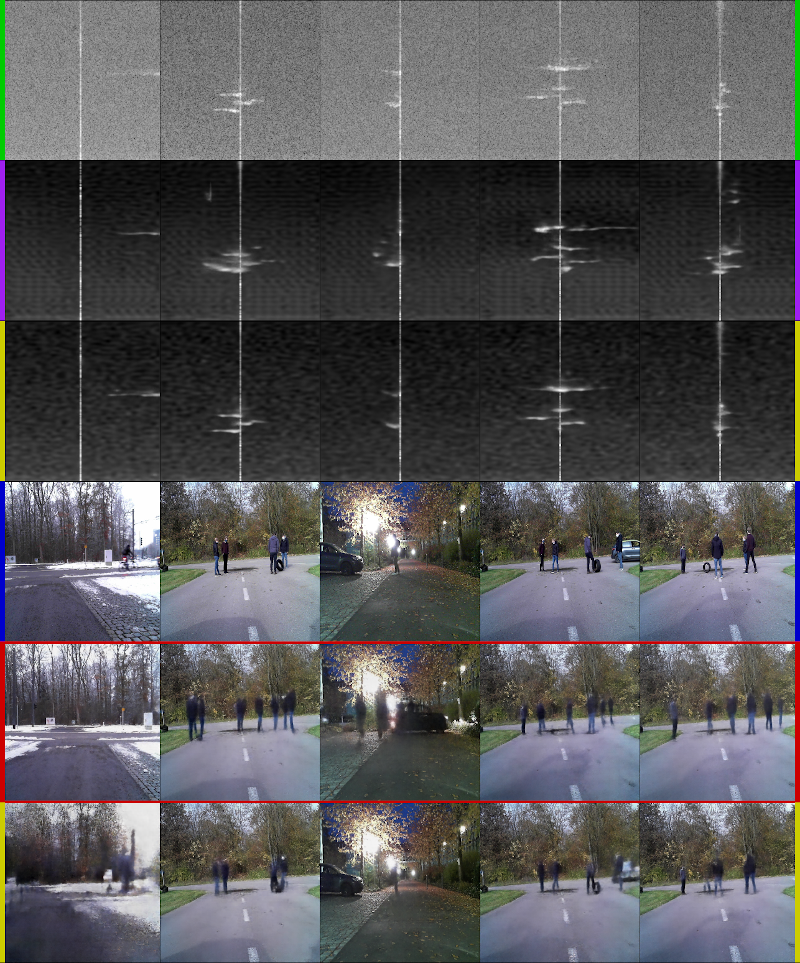}
  \caption{Five random samples taken from the validation dataset and the
    probabilistic camera view reconstruction with top-1/mode sampling of models with
    $K=256$ based on the depicted RF information \textcolor{green}{(green)}. The
    same color coding introduced in Figure \ref{fig:mid_example} applies. The
    transformer model is capable to include most of the essential scene elements
    in the reconstructions \textcolor{red}{(red)} albeit with some creative
    freedom concerning their position and number. A general problem is the
    orientation of entities like cars and pedestrians. Only the camera sensor
    provides azimuthal information so that predictions with respect to the
    lateral positioning of objects are naturally less accurate than estimating
    their radial distance.}
  \label{fig:trafo1}
\end{figure*}
\begin{figure*}[h]
  \centering
  \includegraphics[width=\textwidth]{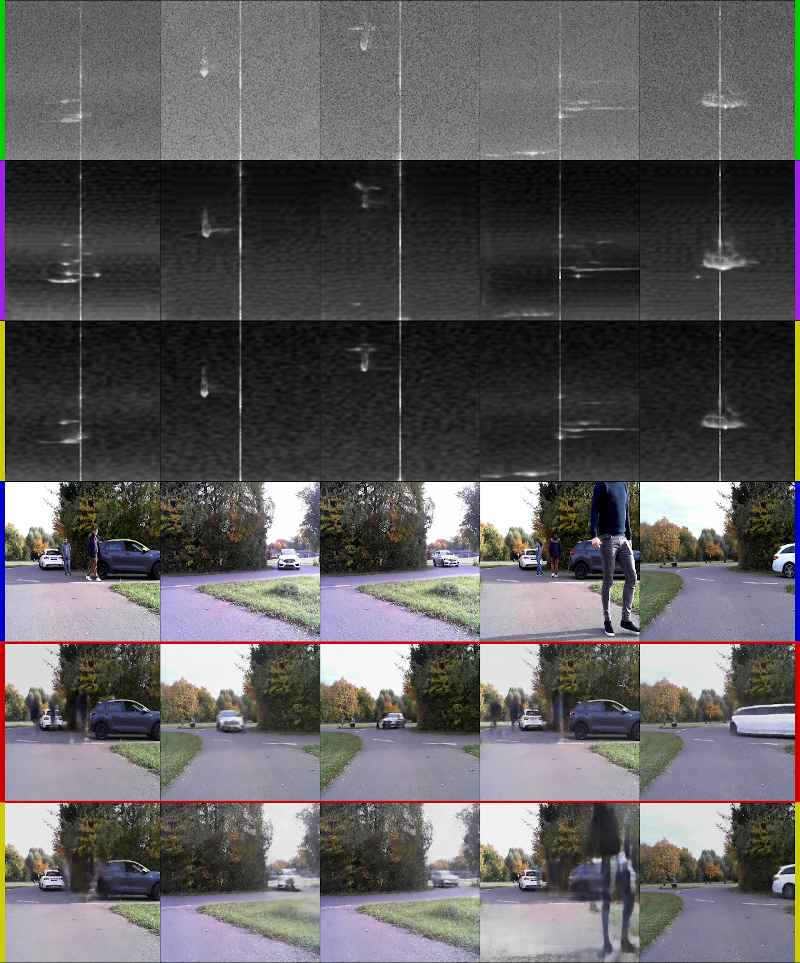}
  \caption{Surprisingly, at times, the top-1 predicted environment perception
    \textcolor{red}{(red)} for the models with $K=256$ surpasses the stochastic
    reconstruction of the discretized camera image
    \textcolor{darkyellow}{(yellow)} in terms of visual quality and contour
    sharpness, as for the dark car in the left example. Again, the model
    confuses left and right and therefore mixes up the direction of the car
    driving in the roundabout. Sometimes, it only manages to include rough
    sketches of what seems to be pedestrians or synthesizes artificial car
    shapes. The same color coding introduced in Figure \ref{fig:mid_example}
    applies.}
  \label{fig:trafo2}
\end{figure*}
\begin{figure*}[h]
  \centering
  \includegraphics[width=\textwidth]{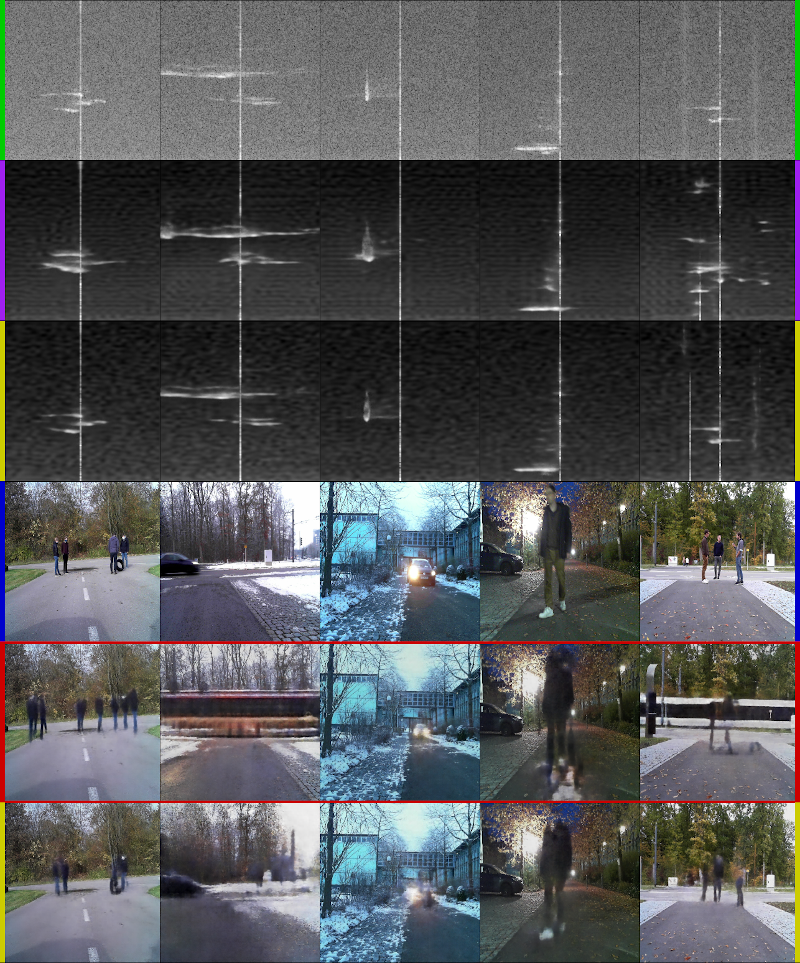}
  \caption{For highly dynamic scenarios, the model with $K=256$ can become
    distracted so that the radar-based conditioning is impaired and spread out
    across most Doppler cells \textcolor{darkpurple}{(purple)}. As a
    consequence, the network assumes the existence of fast extended objects and,
    for top-1 sampling, integrates suitable elements into the scene which it saw
    during cross-modal training (second and fifth example). The center
    illustration shows a typical case in which the prediction qualitatively
    outranks the visual upper bound \textcolor{darkyellow}{(yellow)}, i.e. the
    immediate reconstruction of the discretized input, although the cars
    distance is slightly off. The same color coding introduced in Figure
    \ref{fig:mid_example} applies.}
  \label{trafo3}
\end{figure*}

\begin{figure*}[h]
  \centering
  \includegraphics[width=\textwidth]{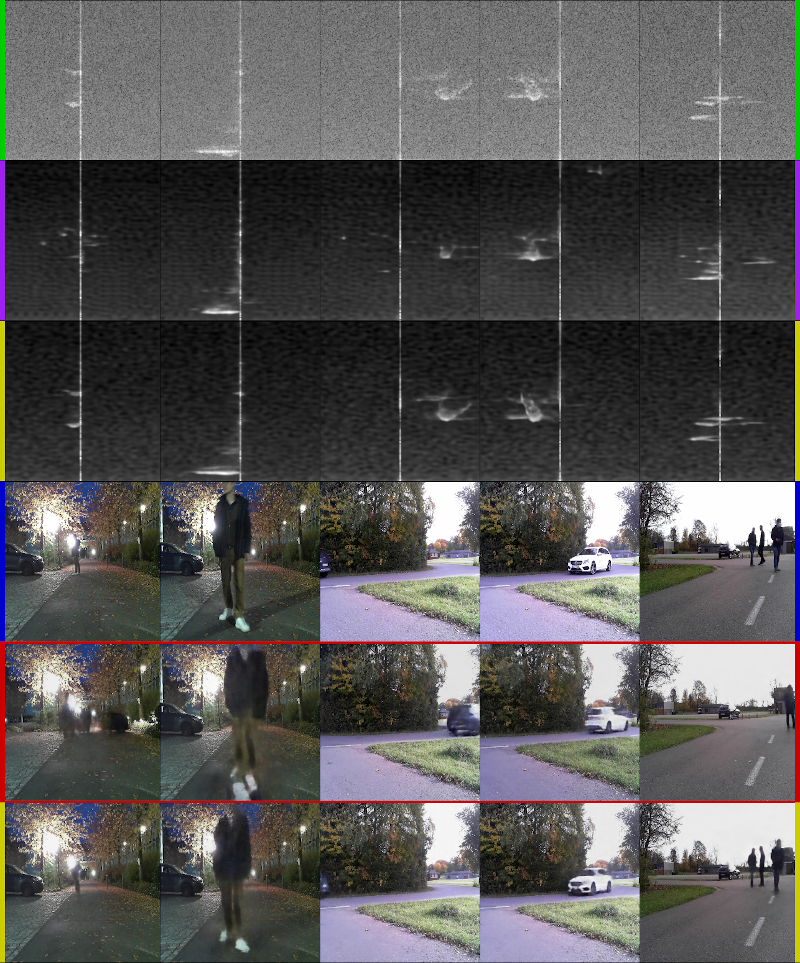}
  \caption{The transformer model relying on compression models with $K=1024$ and
    top-1 sampling are likewise able to infer the general topological ordering
    of diverse outdoor scenes but with enhanced level of detail
    \textcolor{red}{(red)}. However, they too suffer from lateral position
    insecurity and generate objects with azimuthal offset or reversed
    orientation into the environment. The inferred entities are sometimes just
    indistinct shapes in the darkness but still convey vital information about
    the potential existence of dangerous objects in front of the sensor. The
    same color coding introduced in Figure \ref{fig:mid_example} applies.}
  \label{fig:trafo4}
\end{figure*}
\begin{figure*}[h]
  \centering
  \includegraphics[width=\textwidth]{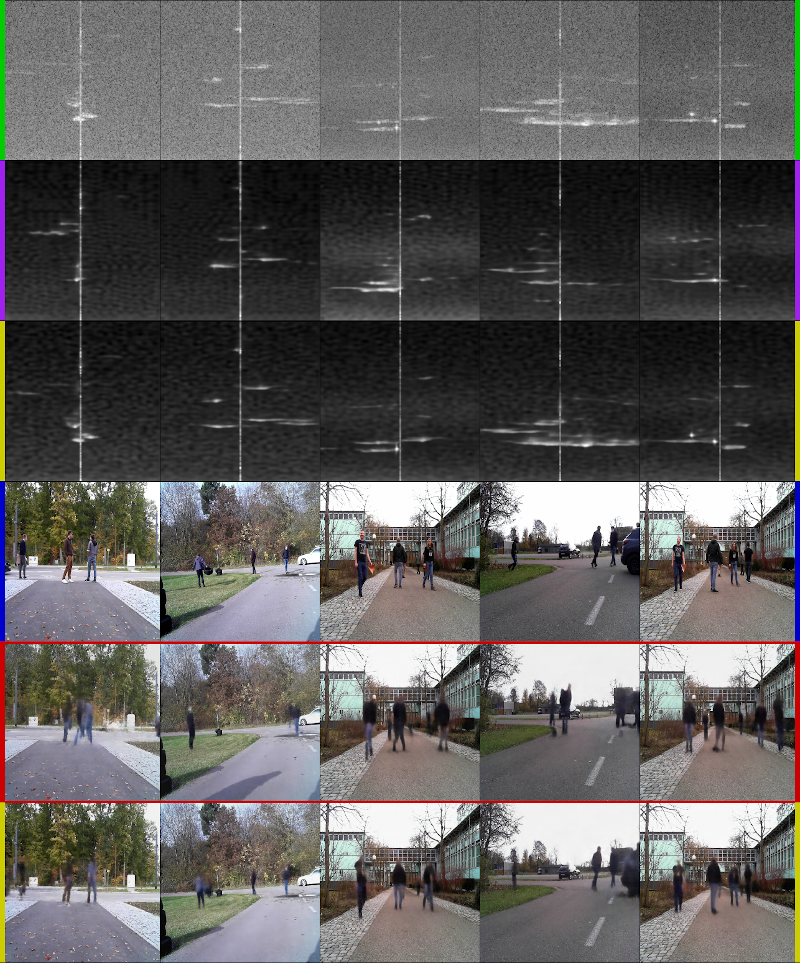}
  \caption{The model with $K=1024$ and top-1 sampling is able to infer the
    relative distance between obstacles and sensor with high precision but
    occasionally misses out on individual subjects entirely. The inclusion of
    even partially occluded pedestrians in the far background (third and fifth
    example) underlines the significance of using a complementary multi-modal
    sensor setup and supports the arguments stated in the introduction about
    retaining as much information as possible. The same color coding introduced
    in Figure \ref{fig:mid_example} applies.}
  \label{fig:trafo5}
\end{figure*}



\begin{figure*}[h]
  \centering
  \includegraphics[width=\textwidth]{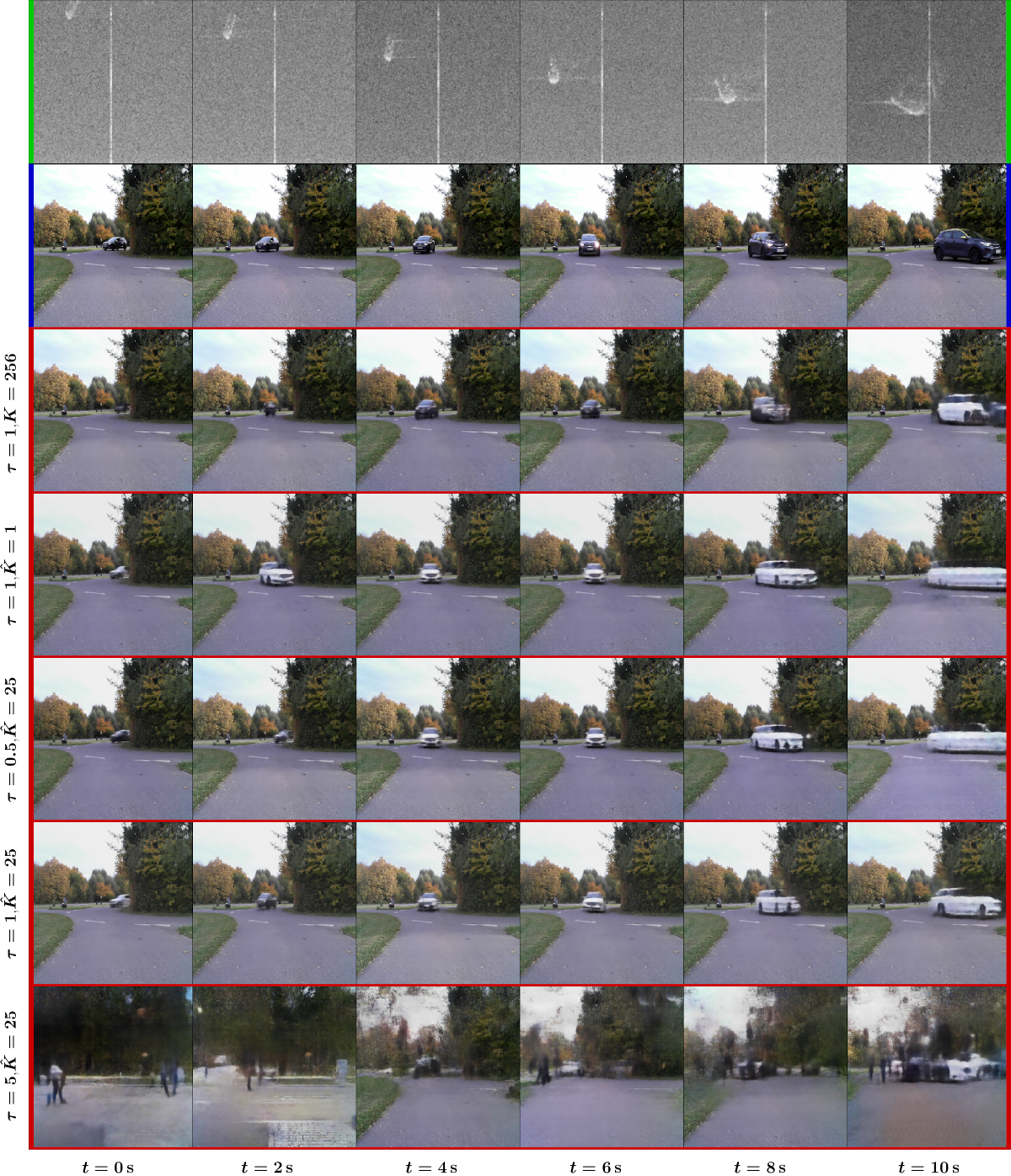}
  \caption{Nucleus and top-k inference with a limited number of categories
    $\hat{K}=25$ to sample camera constituents from. The temporal context
    underlines the differences in synthesis quality when the sample temperature
    varies. Unconstrained category selection over the entire camera sample space
    $K=256$ as well as top-1 sampling with $\hat{K}=1$ serve as basic visual
    references. \textbf{Most strikingly is the change in inferred car color and
      car positions for different temperatures highlighting this parameters
      influence. For too large temperatures, the model still predicts some
      coarse tendencies correctly but generally fails to achieve a valid scene
      reproduction.} The same color coding introduced in Figure
    \ref{fig:mid_example} applies.}
  \label{fig:roundabout_joined}
\end{figure*}

\begin{figure*}[h]
  \centering
  \includegraphics[width=\textwidth]{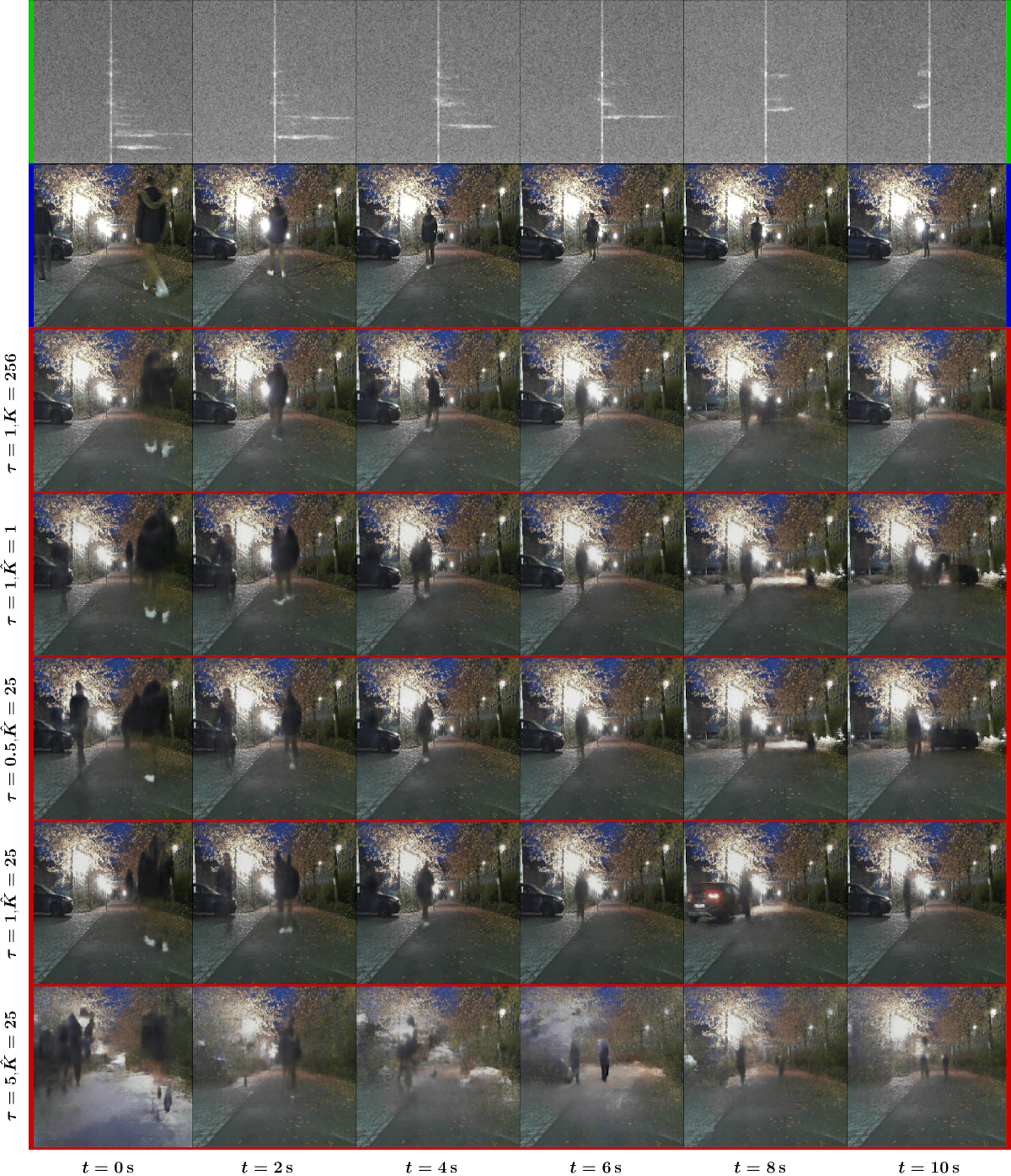}
  \caption{Nucleus and top-k inference with a limited number of categories
    $\hat{K}=25$ to sample camera constituents from. The temporal context
    underlines the differences in synthesis quality when the sample temperature
    varies. Unconstrained category selection over the entire camera sample space
    $K=256$ as well as top-1 sampling with $\hat{K}=1$ serve as basic visual
    references. \textbf{This challenging scenario at dawn with glaring lights is
      sufficiently recovered, but the model occasionally senses more VRU than
      actually present, places the car in the center of the road or erroneously
      turns on its rear lights. These errors are understandable having their
      origin in the absence of cross-modal correspondence and rarely compromise the
      validity of the perception in terms of risk assessment.} The same color
    coding introduced in Figure \ref{fig:mid_example} applies.}
  \label{fig:old_max_joined}
\end{figure*}

\begin{figure*}[h]
  \centering
  \includegraphics[width=\textwidth]{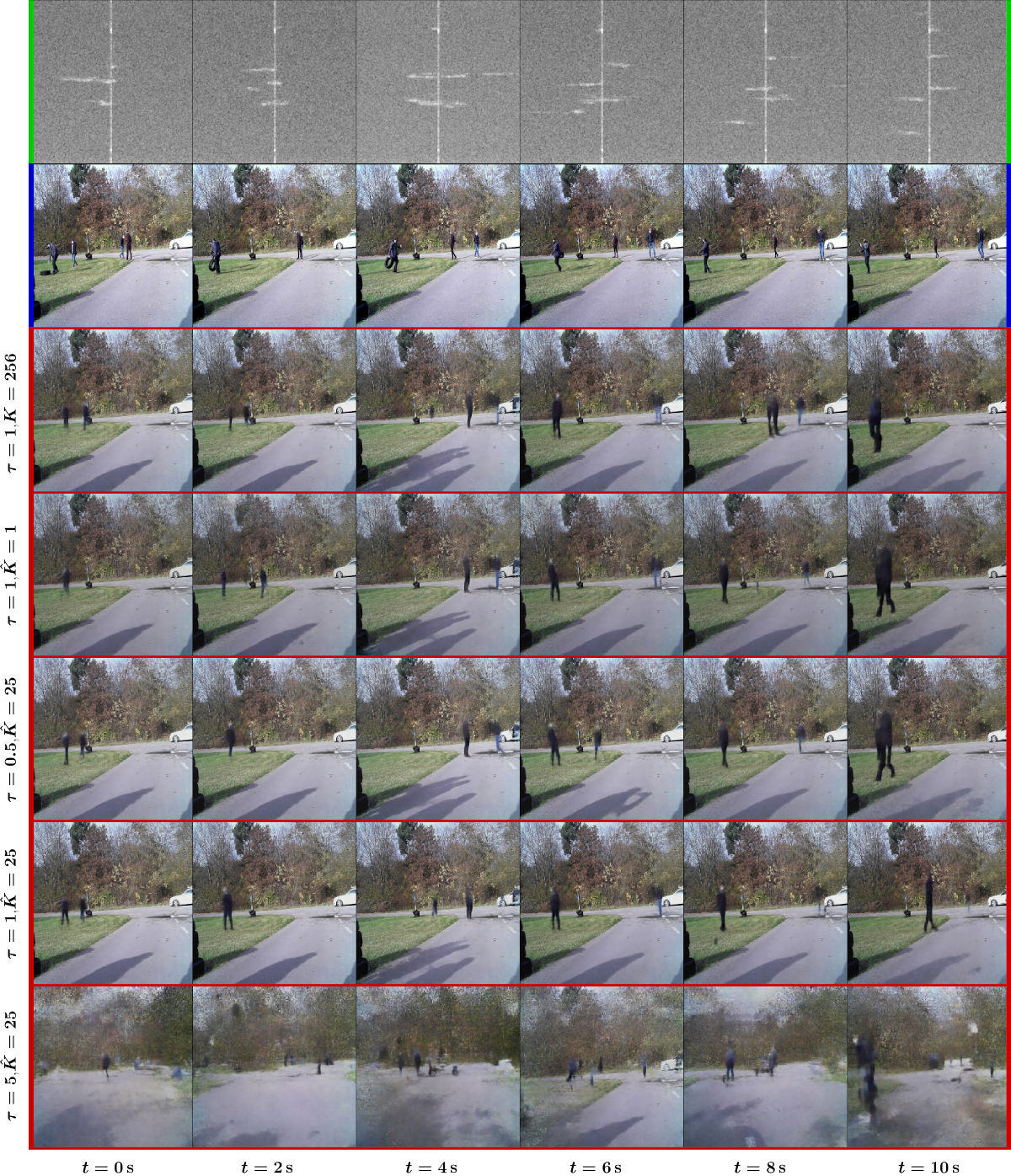}
  \caption{Nucleus and top-k inference with a limited number of categories
    $\hat{K}=25$ to sample camera constituents from. The temporal context
    underlines the differences in synthesis quality when the sample temperature
    varies. Unconstrained category selection over the entire camera sample space
    $K=256$ as well as top-1 sampling with $\hat{K}=1$ serve as basic visual
    references. \textbf{Multiple similar reflections by static objects like
      poles, tires and parked cars as well as micro-Doppler information of
      pedestrians make this scene challenging. The model is able to recapitulate
      only some of its complex details, but regularly confuses the number and
      position of the VRU. Missing out on individuals is the worst-case
      scenario for autonomous systems whereas generating VRU in the proximity
      of the sensor when there are actually none is less problematic}. The
    same color coding introduced in Figure \ref{fig:mid_example} applies.}
  \label{fig:old_wheels_joined}
\end{figure*}

\begin{figure*}[h]
  \centering
  \includegraphics[width=\textwidth]{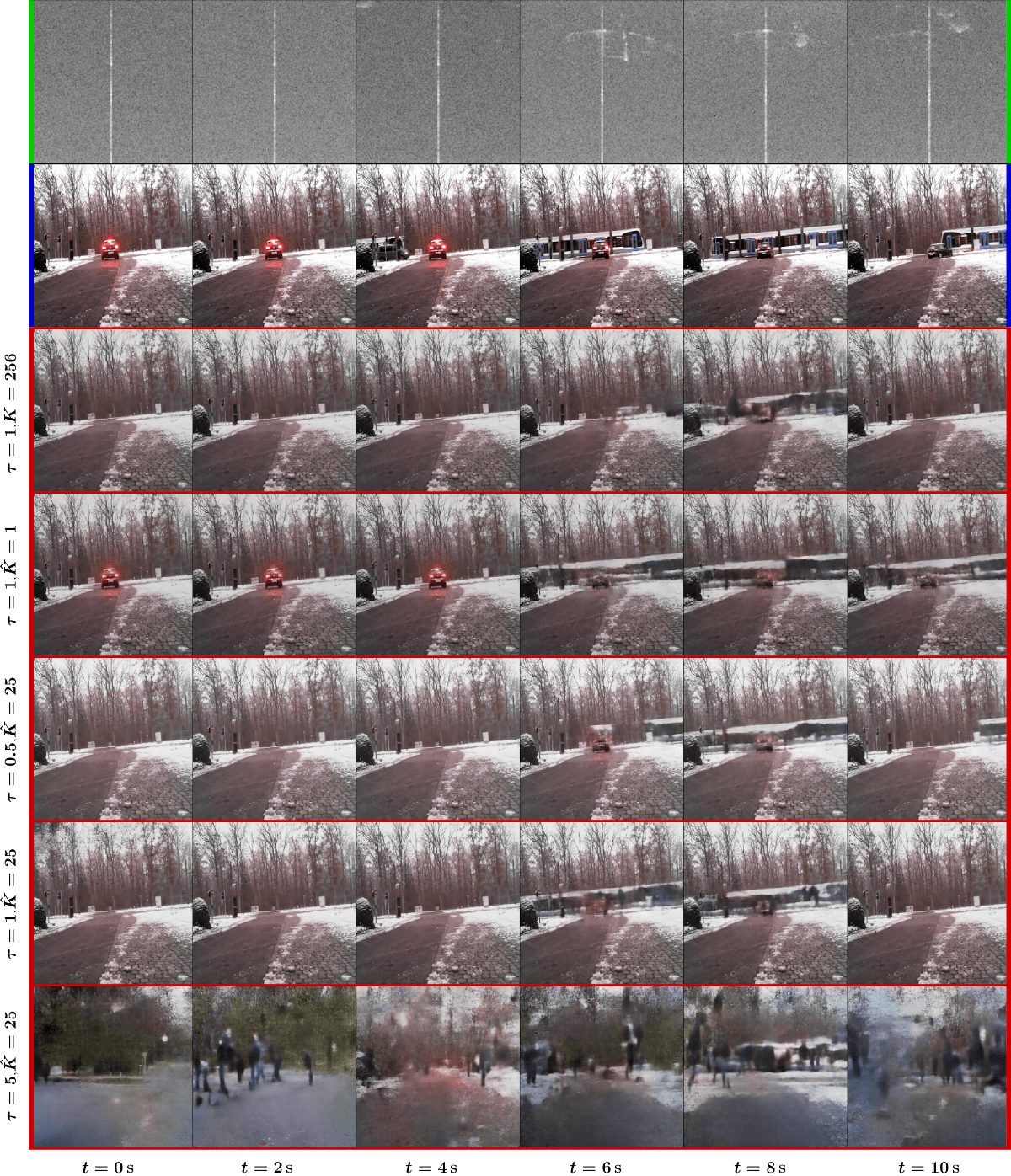}
  \caption{Nucleus and top-k inference with a limited number of categories
    $\hat{K}=25$ to sample camera constituents from. The temporal context
    underlines the differences in synthesis quality when the sample temperature
    varies. Unconstrained category selection over the entire camera sample space
    $K=256$ as well as top-1 sampling with $\hat{K}=1$ serve as basic visual
    references. \textbf{The unique Doppler signature of the passing tram is used
      to establish cross-modal correspondence so that the model correctly
      depicts train-resembling instances in the synthesized scene. The standing
      car only shows in the range section of the frequency plot without any
      relative velocity to the sensor and is therefore easier to miss by the
      models attention.} The same color coding introduced in Figure
    \ref{fig:mid_example} applies.}
  \label{fig:winter_joined}
\end{figure*}

\begin{figure*}[h]
  \centering
  \includegraphics[width=\textwidth]{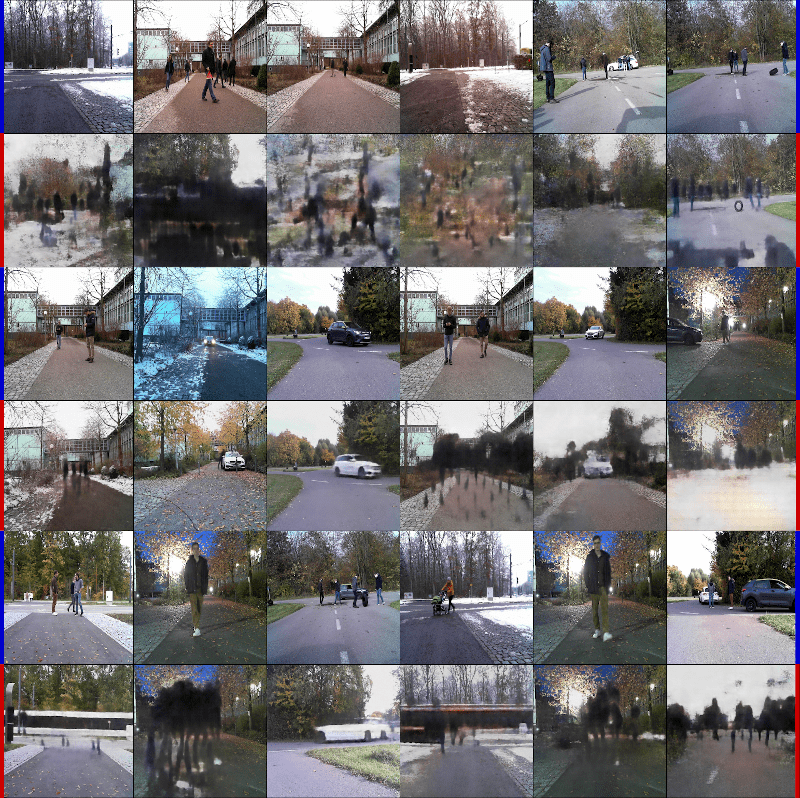}
  \caption{Typical fail cases that occur during the probabilistic inference
    phase \textcolor{red}{(red)} below the actual camera ground truth
    \textcolor{blue}{(blue)}. The reasons for the various misconceptions and
    complex errors are obscure and not always immediately comprehensible. Still,
    it is instructive to take a closer look at the many deceptive examples to
    get a better understanding of the models assumptions and trains of thought.}
  \label{fig:fail_gen}
\end{figure*}


\end{document}